\documentclass[preprint,3p,times]{elsarticle}

\usepackage{ecrc}

\volume{00}

\firstpage{1}

\journalname{Information Sciences}

\runauth{}

\jnltitlelogo{Information Sciences}

\usepackage{amssymb}
\usepackage{amsthm}

\usepackage{amsmath}

\usepackage[caption=false,font=normalsize]{subfig}

\usepackage{algorithmic}
\usepackage[linesnumbered,vlined,ruled]{algorithm2e}

\usepackage{array}

\usepackage{multirow}
\usepackage{booktabs}
\usepackage{tabu}
\usepackage{colortbl}
\usepackage[table*]{xcolor}

\usepackage{sidecap}
\usepackage{cases}

\usepackage{hyperref} 

 \biboptions{sort&compress}

\begin{document}

\begin{frontmatter}



\dochead{}

\title{A collaborative decomposition-based evolutionary algorithm integrating normal and penalty-based boundary intersection for many-objective optimization}


\author[GZHU]{Yu Wu}
\author[GZHU]{Jianle Wei}
\author[SCUT]{Weiqin Ying\corref{cor1}}
\cortext[cor1]{Corresponding author.}
\ead{yingweiqin@scut.edu.cn}
\author[SCUT]{Yanqi Lan}
\author[GZHU]{Zhen Cui}
\author[SCUT]{Zhenyu Wang}
\address[GZHU]{School of Computer Science and Cyber Engineering, Guangzhou University, Guangzhou 510006, PR China}
\address[SCUT]{School of Software Engineering, South China University of Technology, Guangzhou 510006, PR China}

\begin{abstract}
Decomposition-based evolutionary algorithms have become fairly popular for many-objective optimization
owing to their excellent selection pressure in recent years. 
However, the existing decomposition methods still are quite sensitive to the various shapes of frontiers of many-objective optimization problems (MaOPs). 
On the one hand, the cone decomposition methods such as the penalty-based boundary intersection (PBI) are incapable of acquiring uniform frontiers due to the radial spatial distribution of their reference lines for MaOPs with very convex frontiers. On the other hand, the parallel reference lines of the parallel decomposition methods including the normal boundary intersection (NBI) might result in poor diversity because of under-sampling near the boundaries for MaOPs with concave frontiers.
In this paper, a collaborative decomposition method is first proposed to integrate the advantages of parallel decomposition and cone decomposition to overcome their respective disadvantages.
This method inherits the NBI-style Tchebycheff function as a convergence measure to heighten the convergence and uniformity of distribution of the PBI method. Moreover, this method also adaptively tunes the extent of rotating an NBI reference line towards a PBI reference line for every subproblem to enhance the diversity of distribution of the NBI method.
Furthermore, a collaborative decomposition-based evolutionary algorithm (CoDEA) is presented for many-objective optimization.
A collaborative decomposition-based environmental selection mechanism is primarily designed in CoDEA to rank all the individuals associated with the same PBI reference line in the boundary layer and pick out the best ranks.
The proposed algorithm is compared with several popular many-objective evolutionary algorithms on 85 benchmark test instances. The experimental results show that CoDEA achieves high competitiveness benefiting from the collaborative decomposition maintaining a good balance among the convergence, uniformity, and diversity of distribution.
\end{abstract}


\end{frontmatter}


\section{Introduction}
\label{section:introduction}

Multi-objective optimization problems (MOPs) exist widely in engineering practice, scientific research, and other fields, such as portfolio optimization \cite{Zhou2019MDM,Coello2013Survey}, shop scheduling \cite{shop_schedual1,shop_schedual2,cui2020hybrid}, and neural architecture search \cite{NAS1,NAS2}. MOPs with four or more objectives are typically referred to as many-objective optimization problems (MaOPs). How to solve these problems has become a hotspot of scientific research  \cite{2007Pareto,wang2018hybrid}. Compared with the traditional mathematical programming methods, evolutionary algorithms (EAs) are widely applied to solve MaOPs \cite{li2019mobile,li2018improved,li2019performance,cui2020hybrid2}, because they do not need to carry out complex mathematical reasoning and they can obtain a set of approximately optimal solutions in one single run.

The existing many-objective evolutionary algorithms (MaOEAs) can be roughly divided into three categories: Pareto-dominance-based MaOEAs such as the nondominated sorting genetic algorithm-II (NSGA-II)  \cite{NSGA2,NSGA2MTO}, hyper-volume-based MaOEAs including the S-metric selection evolutionary multiobjective optimization algorithm (SMS-EMOA)  \cite{Jiang2015A,Coelho2015Bi,Yuan2017Population,Beumea2007SMS}, and decomposition-based MaOEAs represented by the multi-objective evolutionary algorithm based on decomposition (MOEA/D) \cite{Zhang2007MOEA,6423879,4633340,4982949}. Among them, decomposition-based MaOEAs usually have obvious advantages in terms of running efficiency, distribution uniformity, and selection pressure \cite{7160727,Liu2015Bi,2014A,wang2020improving,cui2019hybrid}.

The population distribution of  decomposition-based MaOEAs is mainly determined by the distribution of reference vectors and the characteristics of aggregation function  \cite{Zhang2007MOEA,7744399,6595549,zjyjysysj}. After decades of development, a unified method for sampling reference vectors had been established \cite{8477730}. Therefore, a large number of studies on the population distribution of decomposition-based MaOEAs are aimed at the improvements of aggregation functions  \cite{7386636,YING202097}. The decomposition methods for decomposition-based MaOEAs can be roughly divided into the cone decomposition methods  \cite{9185860} and parallel decomposition methods  \cite{Li2010NBI,DasAndDenns}. The cone decomposition methods represented by the penalty-based boundary intersection (PBI) \cite{Zhang2007MOEA} can not guarantee the uniform distribution of obtained frontiers. The reason is that the radial spatial distribution of their reference lines generally results in the lack of the ability in dealing with strongly convex Pareto frontiers (PFs). The parallel decomposition methods such as the normal boundary intersection (NBI) \cite{Li2010NBI} are very suitable for the problems with strongly convex PFs. But it is only applicable to the bi-objective optimization problems at the present. When the number of objectives is more than two, the diversity of frontiers generated through the parallel decomposition methods remarkably degrades due to an under-sampling phenomenon
near the boundaries for MaOPs with concave frontiers. As a result, it is generally difficult to apply the parallel decomposition methods to MaOPs.

In addition, when the number of objectives exceeds five, the decomposition-based MaOEAs usually need both a boundary layer of reference lines and an inner one to avoid the hollow phenomenon of population in the inner regions of PFs and acquire a more complete distribution of the population \cite{8477730}. Most of the existing decomposition MaOEAs generally adopt the same aggregation function to optimize both the inner and boundary subproblems \cite{7386636,9185860,8931629}. However, it often leads to a phenomenon of population gathering in the center of PFs at a certain degree and it restricts the improvement of the uniformity of population distribution. 

To solve these issues, this paper puts forward a collaborative decomposition (CoD) method by
integrating both advantages of cone decomposition and parallel decomposition. This method combines the component of the PBI method responsible for preserving the diversity of population with the component of the NBI method responsible for maintaining the convergence and uniformity of the population.
Not only it inherits the advantage of the extensive distribution of the population in the cone decomposition methods, but also it gains an advantage of the uniform population distribution of the parallel decomposition methods.
Furthermore, an MaOEA based on the collaborative decomposition method, referred to as CoDEA, is designed for many-objective optimization.
It mainly adopts a collaborative decomposition-based environmental selection mechanism to rank all the individuals associated with every boundary subproblem and pick out the best ranks.
Meanwhile, this mechanism pushs the individuals associated with the subproblems in the inner layer as far away as possible from the center of the normalized hyperplane. Hence it relieves the phenomenon of population gathering and maintains the uniformity of population distribution along the PFs.

The remainder of this paper is organized as follows. Section \ref{sec:Preliminaries} introduces the preliminaries of this paper. Section \ref{sec:cod} describes the motivations and details of the collaborative decomposition method. 
Section \ref{sec:CoD-DEA} is devoted to the framework of the proposed CoDEA and the procedure of CoD-based environmental selection mechanism. The experimental results and discussions are provided in Section \ref{sec:EXPERIMENTAL}. Finally, Section \ref{sec:CONCLUSION} draws the conclusions of this paper.

\section{Preliminaries}
\label{sec:Preliminaries}

In this section, the basic definitions of MOPs and the classical decomposition-based MOEA are first introduced briefly. At the end of this section, some work related to decomposition-based MaOEA is reviewed.

\subsection{Classical decomposition-based MOEA}
Generally, an MOP can be mathematically defined as follows  \cite{moea}:
\begin{eqnarray}
&minimize&F(x) = (f_1(x),f_2(x),\cdots,f_m(x))^T \label{eqn:MOP}\\
&subject\ to& x\in \Omega \subseteq \mathbb R^n\label{eqn:decision space}
\end{eqnarray}
where $m$ is the dimension of  objectives. In (\ref{eqn:decision space}), $\Omega = \prod_{i=1}^n[l_i,u_i] \subseteq \mathbb R^n $ is called the decision space. Here, $n$ is the dimension of decision variables; $u_i$ and $l_i$ are the upper and lower bounds of the $i$-th decision variable, respectively. 

In the classical decomposition-based MOEA, i.e., MOEA/D  \cite{Zhang2007MOEA}, a complete MOP is decomposed into a series of subproblems according to a set of weight vectors uniformly distributed in the objective space, and each subproblem is associated with a specific weight vector to form a unique aggregation function. Each individual in the population is assigned to one of the subproblems for optimization with the help of aggregation function.

\begin{figure}[htbp]
	\centering
	\includegraphics[width =0.6\linewidth]{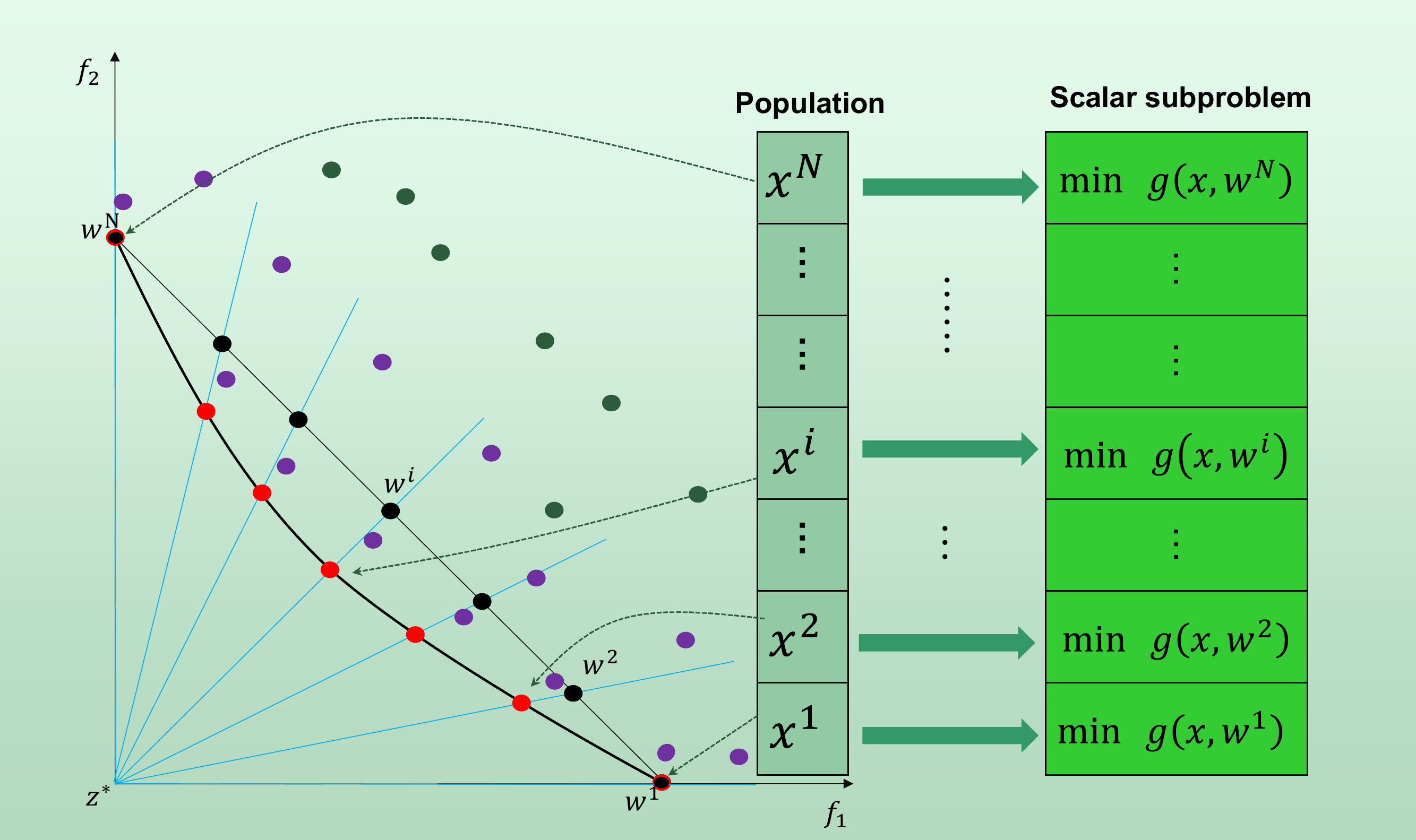}
	\caption{An illustration of the classical MOEA/D.}
	\label{fig:MOEAD}
\end{figure}

As shown in Fig. \ref{fig:MOEAD}, the green dots represent the initialized population, and the newly initialized population is generally far away from the real PF (the black curve). The purple dots represent the population in the evolutionary process, while the red dots represent the Pareto optimal individuals that can be obtained in theory. Here, the reference points are represented by blue rays labeled $w^i (i = 1,2,\cdots,N)$, which are constructed by evenly setting reference points (the black dots) on the hyperplane ($\overline{w_1w_N}$) (The reference vectors and reference points will be denoted by $w^i (i = 1,2,\cdots,N)$ in the following text). In the evolution process, the solutions $x^i (i = 1,2,\cdots,N)$ associated with subproblems are gradually approaching PF with the help of the preset aggregation functions $g(x,w^i) (i = 1,2,\cdots,N)$.

\subsection{Decomposition-based MaOEAs}

The core idea of decomposition-based MaOEAs is decomposition  \cite{Zhang2007MOEA}, and the most important is how to optimize a series of subproblems after decomposition  \cite{8998284,7564425}.


MOEA/D proposed by Zhang and Li studied three aggregation functions, namely, weight-sum method (WS), Tchebycheff method (TCH), and PBI method respectively. The WS method is better than the TCH method in search performance, but the WS method is not suitable for finding Pareto optimal solutions on MOPs with non-convex PF. Therefore, Ishibuchi H.  \cite{AdaptationSF} proposed an idea of automatic selection between the WS method of each generation of individuals and the TCH method. The feature of this idea is the TCH method is only used when a solution is in the nonconvex regions of the Pareto frontier. In MOEA/D, the PBI method has better performance on all three targets than the WS method and the TCH method, but the setting of the parameter $\theta$ of the PBI method depends on the shape of the PF  \cite{7257248,7744053}. Yang S et al.   \cite{2016Improving} studies the influence of $\theta$ on the PBI method and proposed two new punishment schemes, namely adaptive penalty schemes (APS) and subproblem-based penalty schemes (SPS). APS adaptively assigns the same penalty value to all subproblems by considering different requirements in every search stage, while SPS specifies different penalty values for different subproblems. Jin believes that the population should pay attention to convergence in the early stage of evolution and more diversity in the later stage of evolution. Therefore, in RVEA  \cite{7386636}, Jin proposed an adaptive angle penalty distance (APD) method which changes with the evolution process. Based on the existing methods, the above studies put forward some improvement schemes and achieved certain results. However, it also limits them to the inherent defects of the existing methods, which cannot be made up by structural adjustment.

The PBI method is a typical method based on cone decomposition, that is, its reference vectors are distributed in a radial cone shape in the target space, and this is a spatially uneven distribution.
Li and Zhang put forward an NBI-style Tchebycheff approach based on parallel decomposition in the MOEA/D framework  \cite{Li2010NBI}. The reference vectors constructed by this method are parallel to each other in objective space. However, the bottleneck of the NBI method is that it is impossible to explore and develop the region beyond the preset parallel vector coverage when the objective dimension is greater than two  \cite{DasAndDenns}. Recently, Carlos A. et al based on the PBI method introduced the mirror point, combined the idea of parallel decomposition with the idea of cone decomposition, and proposed a new algorithm MP-DEA  \cite{8931629}. In MP-DEA,  a mirror point relation (MPR) method is used to rank the individuals in a niche. This method replaces the Euclidean distance between the individual and the ideal point associated with it by the Euclidean distance between the solution and the mirror point, and the uniformity is improved to a certain extent. However, the aggregation function is still based on the idea of penalizing the Euclidean distance between an individual and a point, so it is essentially a variant of the PBI method.


\section{Collaborative decomposition}
\label{sec:cod}

\begin{figure}[htbp]
	\centering
	\includegraphics[width =0.35\linewidth]{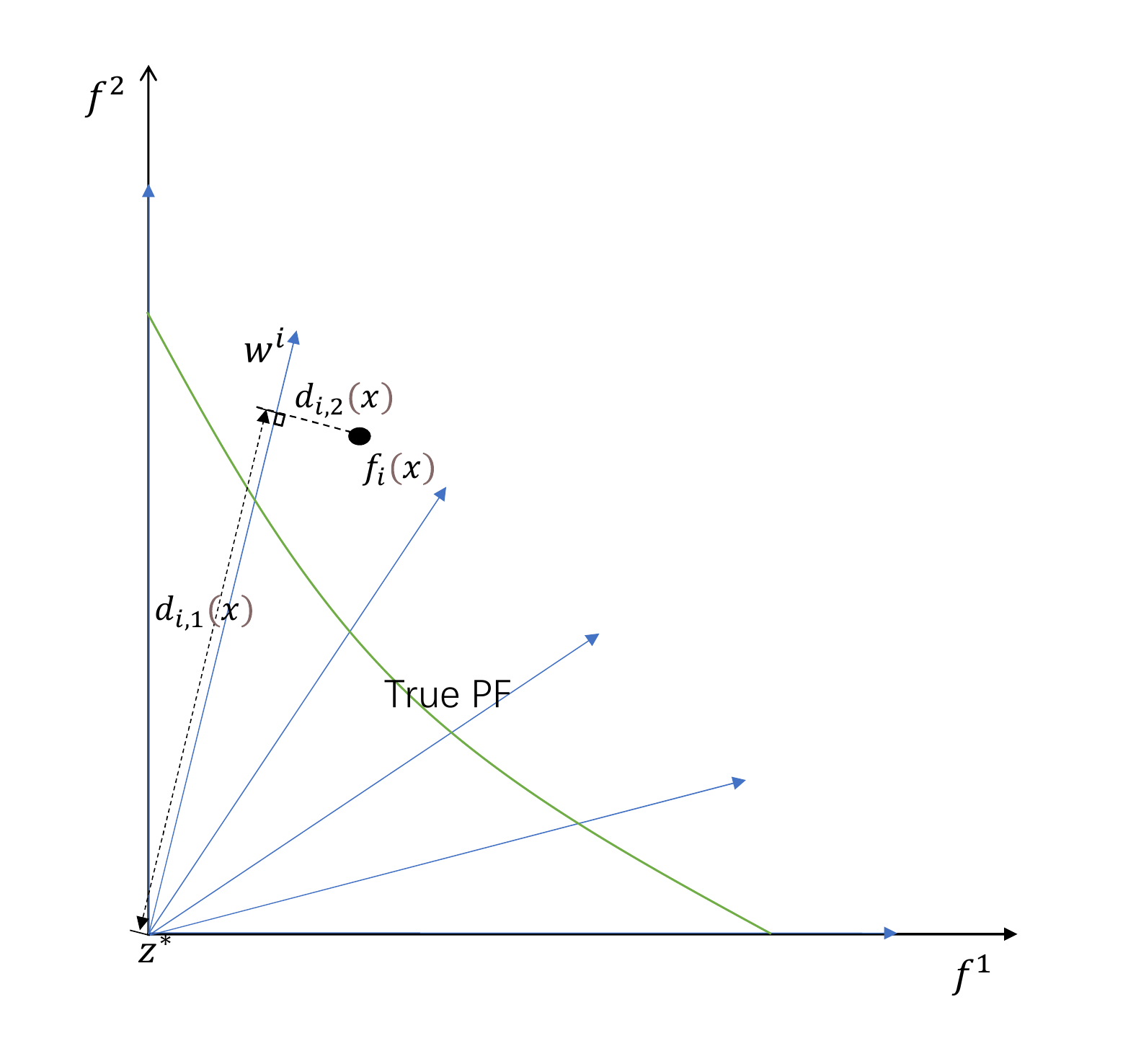}
	\caption{An illustration of the PBI method.}
	\label{fig:PBI}
\end{figure}

The proposed CoD method combines the ideas of conical decomposition and parallel decomposition. Therefore, this section first introduces the disadvantages of the PBI and NBI methods and the motivation of the CoD method briefly. Afterwards,
the details of the CoD method is presented.
\subsection{Motivation}

The PBI method has been widely studied and applied since it can usually obtain the frontiers with the better distributions than the WS and TCH methods.
Generally, given any individual $\textbf{x}$ associated with the $i$-th reference vector $\textbf{w}^i$, the PBI aggregation function of the $i$-th subproblem can be formulated as:
\begin{eqnarray}
&minimize& g^{PBI}(\textbf{x}|\textbf{w}^i) = d_{1}(\textbf{f}(\textbf{x}),\textbf{w}^i) + \theta \times d_{2}(\textbf{f}(\textbf{x}),\textbf{w}^i). \label{eq:PBI}
\end{eqnarray}
In the above formula, $d_{1}(\textbf{f}(\textbf{x}), \textbf{w}^i)$ denotes the length of the projection under the reference vector $\textbf{w}^i$ of the objective vector $\textbf{f}(\textbf{x})$, and $d_{2}(\textbf{f}(\textbf{x}),\textbf{w}^i)$ indicates the vertical distance from $\textbf{f}(\textbf{x})$ to $\textbf{w}^i$, as shown in Fig. \ref{fig:PBI}. In addition, the parameter $\theta$ 
denotes a constant, which is used to balance the convergence and diversity. And it is usually necessary to customize the parameter $\theta$ according to the characteristics of the problems.

\begin{figure}[htbp]
	\centering
		\includegraphics[width =0.35\linewidth]{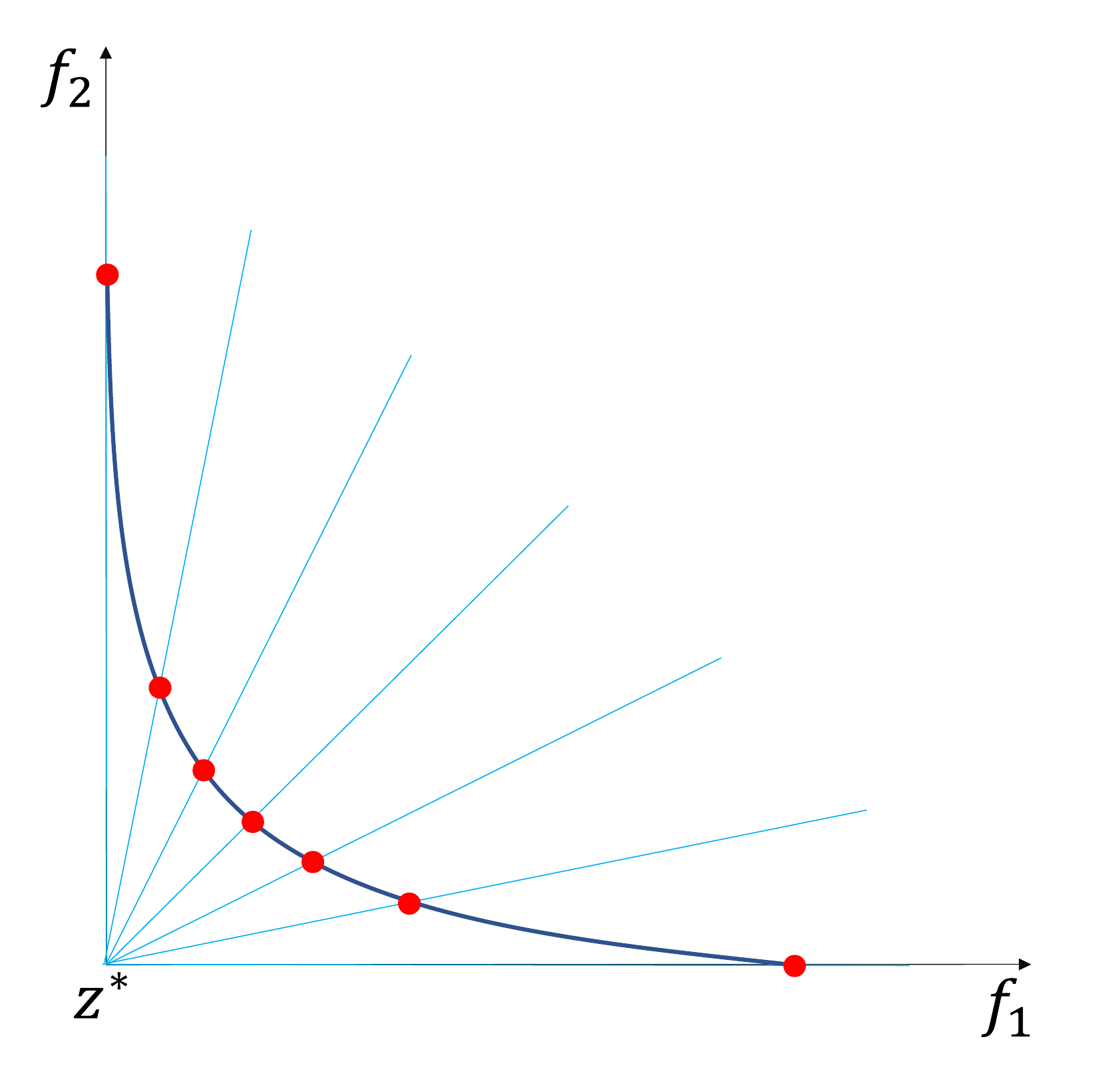}
	\caption{An illustration of intersections of a strongly convex PF with the reference lines of the PBI method.}
	\label{fig:PBI_ntersections}
\end{figure}

In general, the cone decomposition methods represented by the PBI method perform well on the MaOPs whose PFs are relatively close to the unit hyperplane formed by the reference points. However, given an MOP with a strong convex PF, it is difficult for the PBI method to obtain a set of uniformly distributed non-dominated solutions because its reference vectors are radial.
As shown in Figs. \ref{fig:PBI_ntersections}, when the PF of an MOP is strongly convex, the portions of PF near the objective axis sharply change. It means that a small conical region near the objective axis covers a very long portion of the PF, while the conical region with the same size at the center of the objective space includes only a relatively short portion.
As a result, for any MOP with a strongly convex PF, the intersections of the PF with the reference lines of the PBI method are very dense near the boundary region of PF and sparse in the middle region.


\begin{figure}[htbp]
	\centering
	\includegraphics[width =0.35\linewidth]{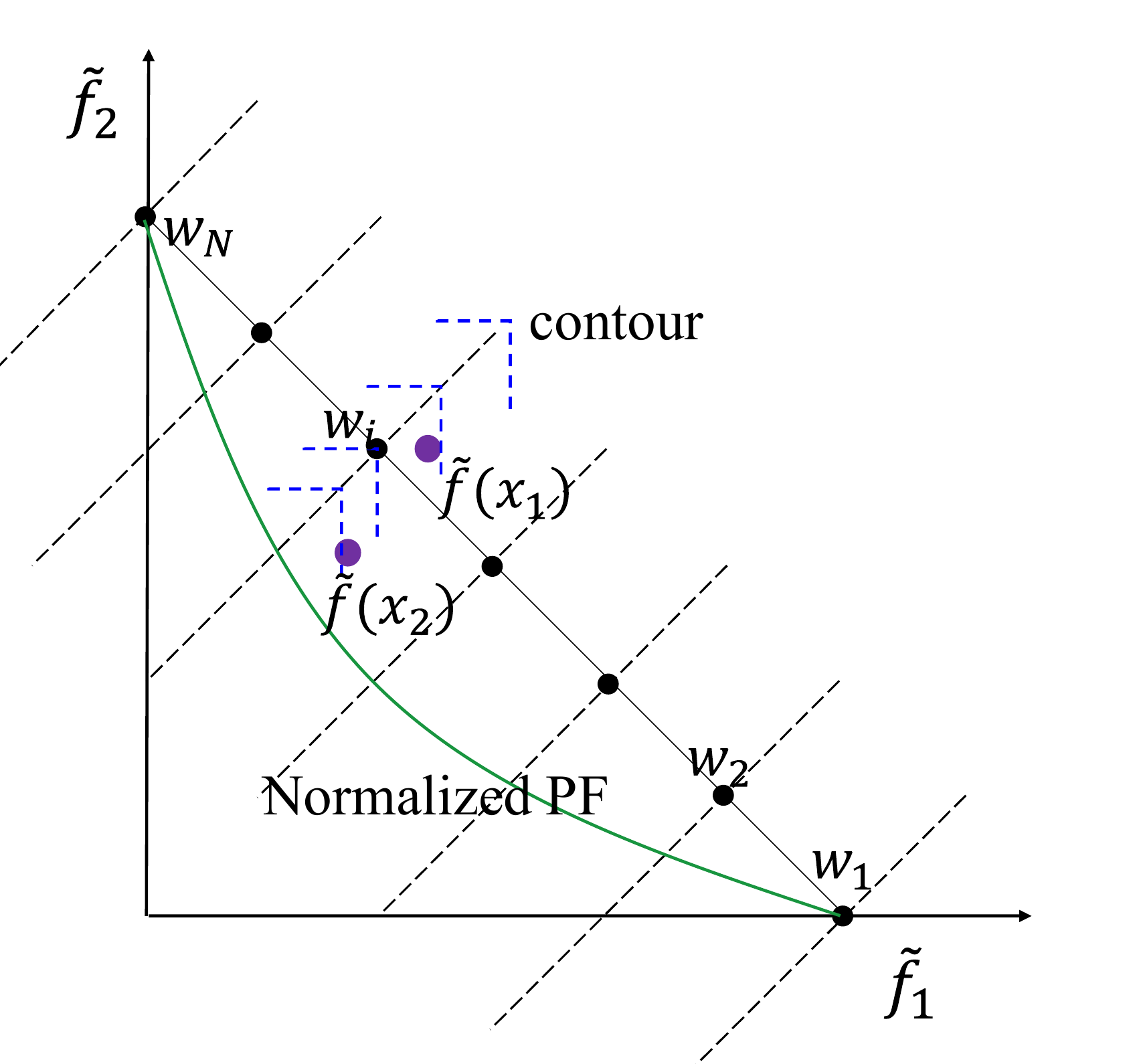}
	\caption{An illustration of the NBI method.}
	\label{fig:NBI}
\end{figure}

Essentially, the NBI method constructs a group of parallel reference lines through the reference vectors $\textbf{w}^i$, $i = 1, 2,\cdots,N$, as shown in Fig. \ref{fig:NBI}.
The aggregation function \cite{Li2010NBI} of the $i$-th NBI-style Tchebycheff subproblem in the NBI method can be formulated as follows:
\begin{eqnarray}
&minimize& g^{NBI}(\textbf{x}|\textbf{w}^i) = max^{m}_{j=1}\{f_{j}(\textbf{x}|\textbf{w}^i)-w_j^i\}. \label{eq:NBI}
\end{eqnarray}
In the above formula, the $i$-th subproblem aims to minimize the function $g^{NBI}(\textbf{x}|\textbf{w}^i)$ for any solution $\textbf{x}$ associated with the $i$-th subproblem. And the value of $g^{NBI}(\textbf{x}|\textbf{w}^i)$ indicates the maximum component of the difference between the objective vector $f_{j}(\textbf{x}|\textbf{w}^i)$ in the normalized objective space and the corresponding reference point $\textbf{w}^i$.
The blue dotted lines in Fig. \ref{fig:NBI} denote the contours of $g^{NBI}(\textbf{x}|\textbf{w}^i)$ and the solution in a contour closer to the ideal point has the higher priority. In the case of Fig. \ref{fig:NBI}, solution $\textbf{x}^2$ has a better NBI aggregation value than solution $\textbf{x}^1$.

\begin{figure}[htbp]
	\centering
	\includegraphics[width =0.35\linewidth]{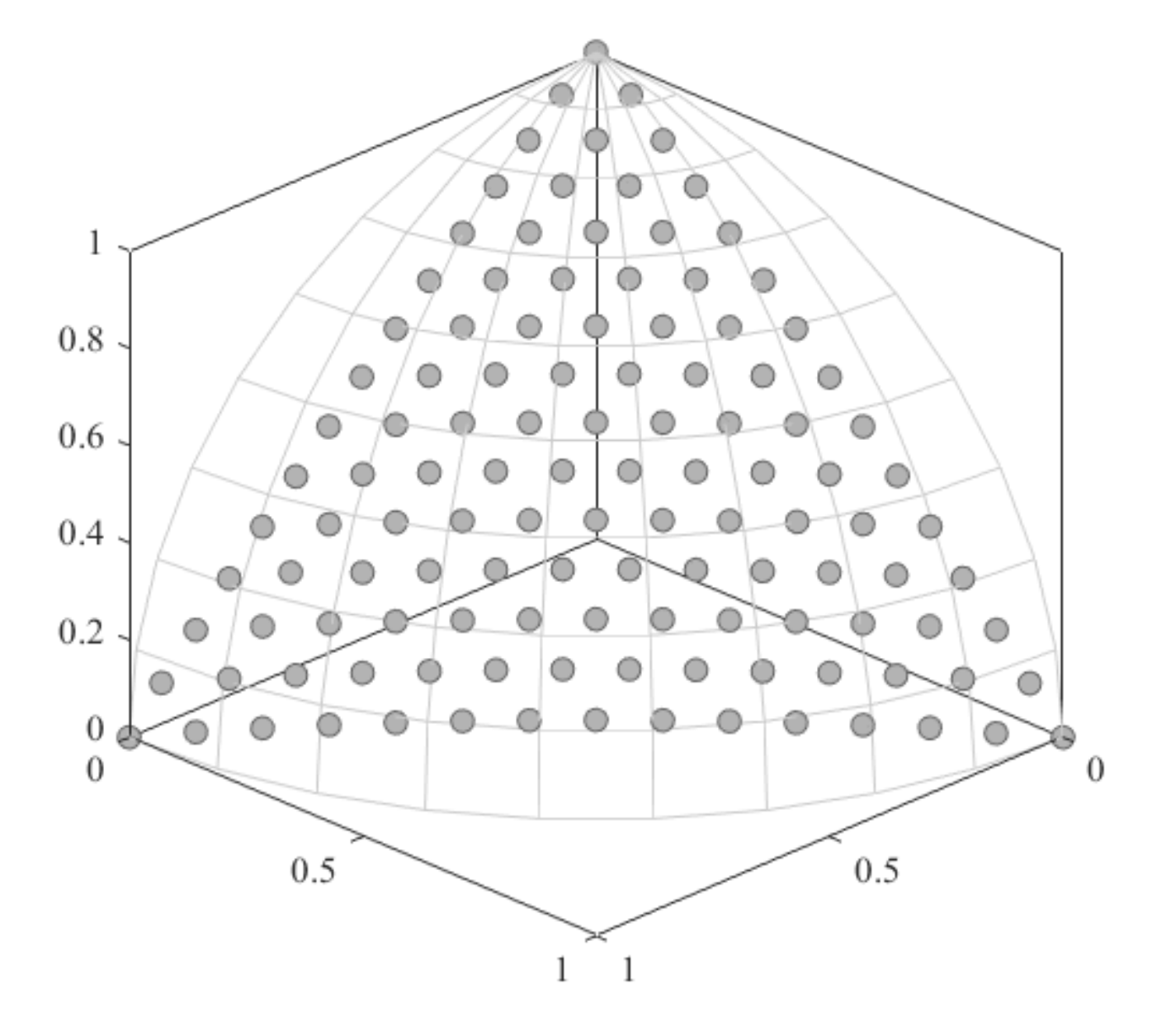}
	\caption{An illustration of an incomplete frontier sampled by the NBI method on the tri-objective DTLZ2 test problem.}
	\label{fig:NBI_1}
\end{figure}

In general, the NBI method is able to capture much more uniform frontiers even for MaOPs with strongly convex frontiers than the PBI method owing to the parallel reference lines.
However, the NBI method might result in poor diversity because of under-sampling
near the boundaries for MOPs with both more than two objectives and concave frontiers.
For example, the mesh surface in Fig. \ref{fig:NBI_1} indicates the concave frontier of the DTLZ2 test problem \cite{2002Scalable} with three objectives. The parallel normal reference lines perpendicular to the unit hyperplane have no intersection with the boundary regions of concave frontiers.
Therefore, the NBI method has a very poor ability of capturing the boundary regions of concave frontiers, and it can only obtain an incomplete approximate frontier, as shown by the grey dots in Fig. \ref{fig:NBI_1}. This results in that it is very difficult to extend the advantage of uniformity of the NBI method to MaOPs.


In short, the PBI method is good at keeping the population diversity, but not at maintaining the population uniformity. Coincidentally, the NBI method is just the opposite of the PBI method. The motivation of the collaborative decomposition method proposed in this paper is to integrate both advantages of the PBI and NBI methods. In this paper, the collaborative decomposition method combines the component of the NBI method preserving the uniformity with the component of the PBI method maintaining the diversity. It takes advantage of their strengths to make up for their respective disadvantages and improve the performances of decomposition-based MaOEAs.

\subsection{Collaborative decomposition method}

\begin{figure}[htbp]
	\centering
	\includegraphics[width =0.35\linewidth]{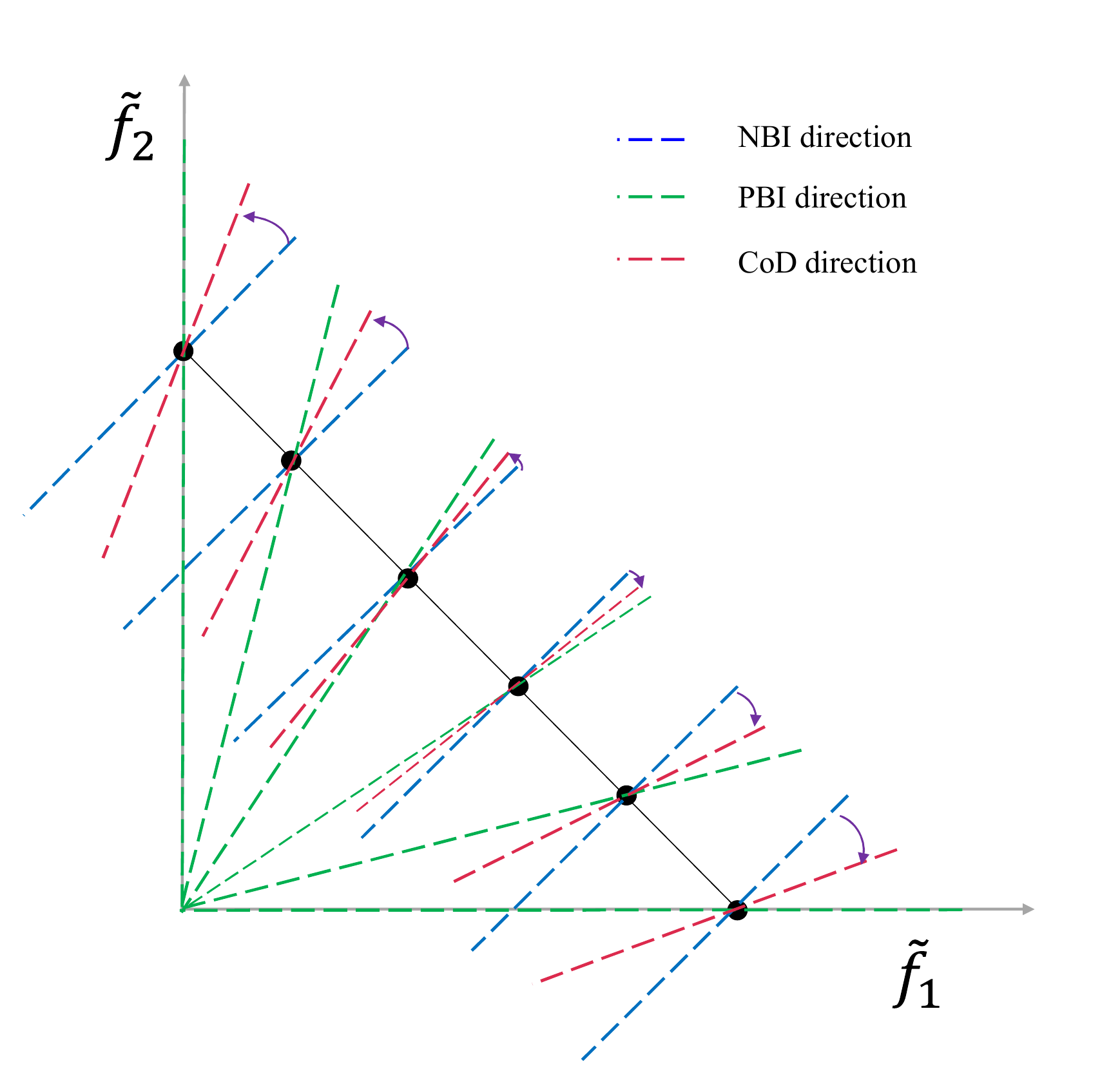}
	\caption{The CoD reference lines.}
	\label{fig:CoDD_1}
\end{figure}

As mentioned earlier, the CoD method is to combine the component of the NBI method and the component of the PBI method. 
From the viewpoint of implementation, the CoD method exerts the influence of the PBI method on the basis of the NBI method. 
For the convenience of intuitive observation, a reference line of the CoD method can be viewed as an NBI reference line rotated towards the respective PBI reference line to a certain extent. As shown in Fig. \ref{fig:CoDD_1}, the NBI reference lines are rotated towards the PBI reference lines by different angles to obtain the reference lines of the CoD method.
And the distribution of population is guided by the reference lines of the CoD method.

Mathematically, the aggregation function of the $i$-th subproblem in the CoD method is designed as follows:
\begin{eqnarray}
&minimize& g^{CoD}(\textbf{x}|\textbf{w}^i) = g^{NBI}(\textbf{x}|\textbf{w}^i)+r_i\times k_m \times d_{2}(\textbf{f}(\textbf{x}),\textbf{w}^i).  \label{eq:CoD_1}
\end{eqnarray}
In the above equation, $r_i$ represents a reference-point-based rotation factor at the $i$-th reference-point $\textbf{w}^i$, and $k_m$ denotes a objective-number-based rotation factor for the number $m$ of objectives. According to Eq. (\ref{eq:CoD_1}),  
the CoD aggregation function is composed of the NBI aggregration function in Eq. (\ref{eq:PBI}) and the vertical distance of the PBI aggregration function in Eq. (\ref{eq:NBI}).
It
can be regarded as an implicit implementation of the idea of rotating the NBI reference lines towards the PBI reference lines to a certain extent. 
The reference-point-based rotation factor $r_i$ and objective-number-based rotation factor $k_m$ adaptively tune the extents of rotating the NBI reference lines towards the PBI reference lines according to different reference points and different numbers of objectives, respectively. Both rotation factors together control the total extent of rotation.



\begin{figure}[htbp]
	\centering
	\includegraphics[width =0.35\linewidth]{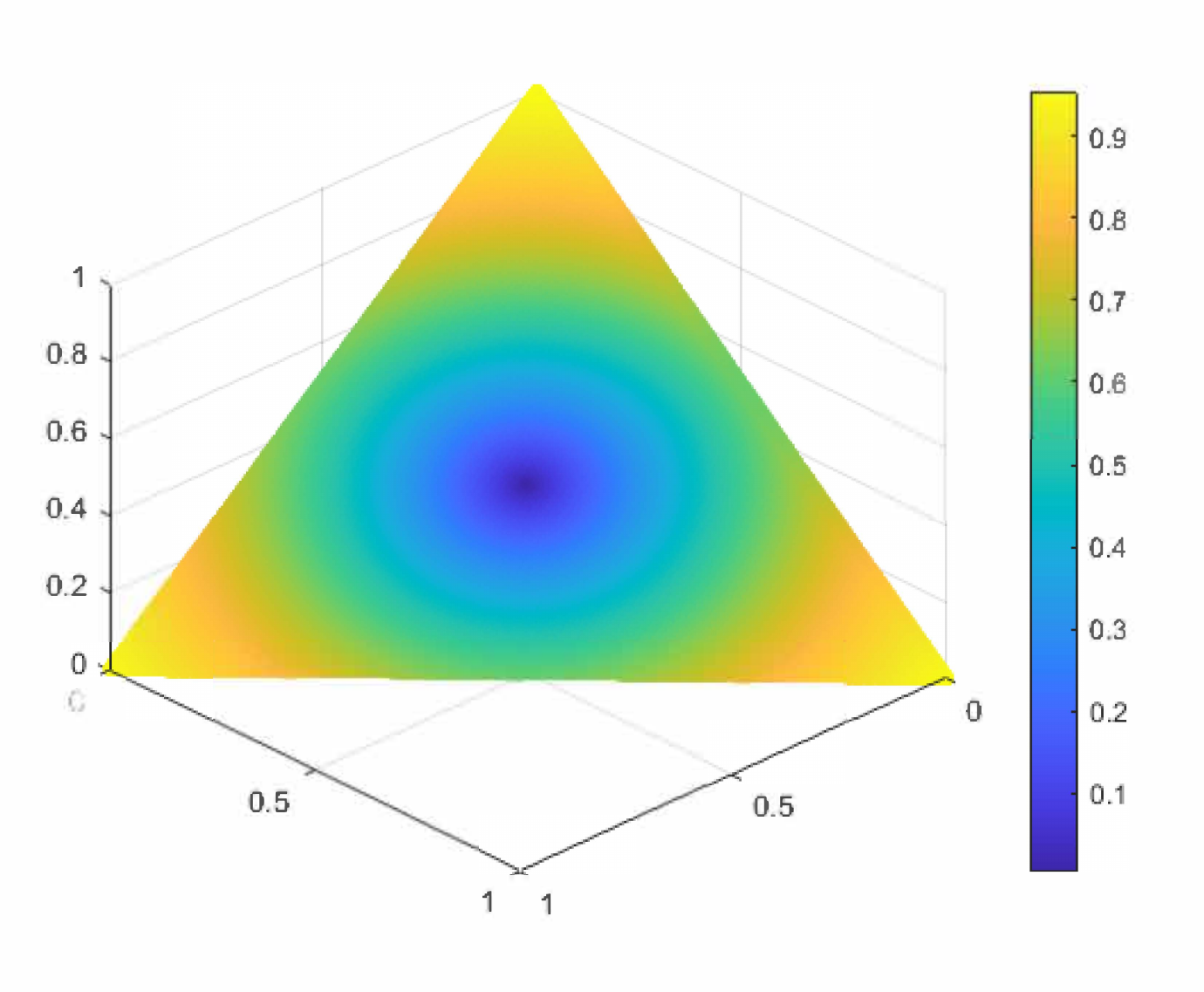}
	\caption{The distribution of the angles between the NBI and PBI reference lines in the tri-objective space.}
	\label{fig:angle}
\end{figure}

It is worth noting that the angle between the NBI and PBI reference lines gradually increases from the center to the edges and vertices of the unit hypervolume.
Figure \ref{fig:angle} plots the distribution of the angle between the NBI and PBI reference lines in the tri-objective space. 
In addition, the NBI method has the more serious phenomenon of under-sampling near the boundaries of the unit hyperplane than at the center.
Therefore, the extent of the rotation  should change at different reference points. 
Based on the advantages and disadvantages of the NBI method and the PBI method analyzed in Section \ref{sec:cod}, the CoD method divides $r_i$ into two modules,  $\alpha_i$ and $\beta_i$, according to the following equation: 
\begin{eqnarray}
r_i = \frac{\alpha_i+\beta_i}{2}.  \label{eqn:DC}
\end{eqnarray}


 On one hand, to inherit the diversity of the PBI method at the edge, we need to ensure that the closer to the edge the reference vector is, the larger its rotation angle is. Here, we define $\alpha_i$ as follows:
\begin{eqnarray}
\alpha_i = 1-min_{j=1}^m\{w_j^i\} \times m. \label{eqn:alpha}
\end{eqnarray}
\begin{figure}[htbp]
	\centering
	\subfloat[]{
		\includegraphics[width = 0.35\linewidth]{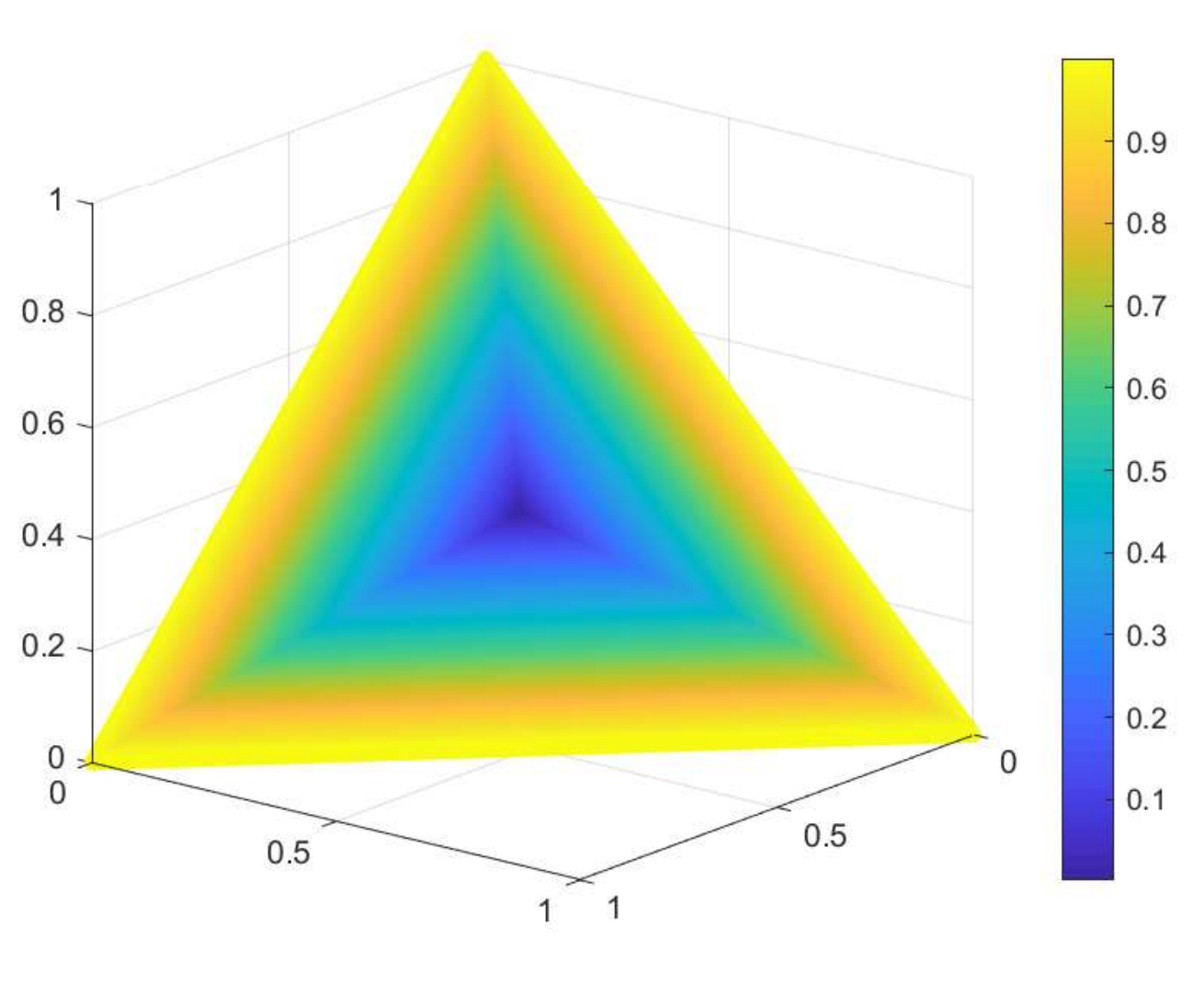}
		\label{fig:alpha}
	}
	\subfloat[]{
		\includegraphics[width = 0.35\linewidth]{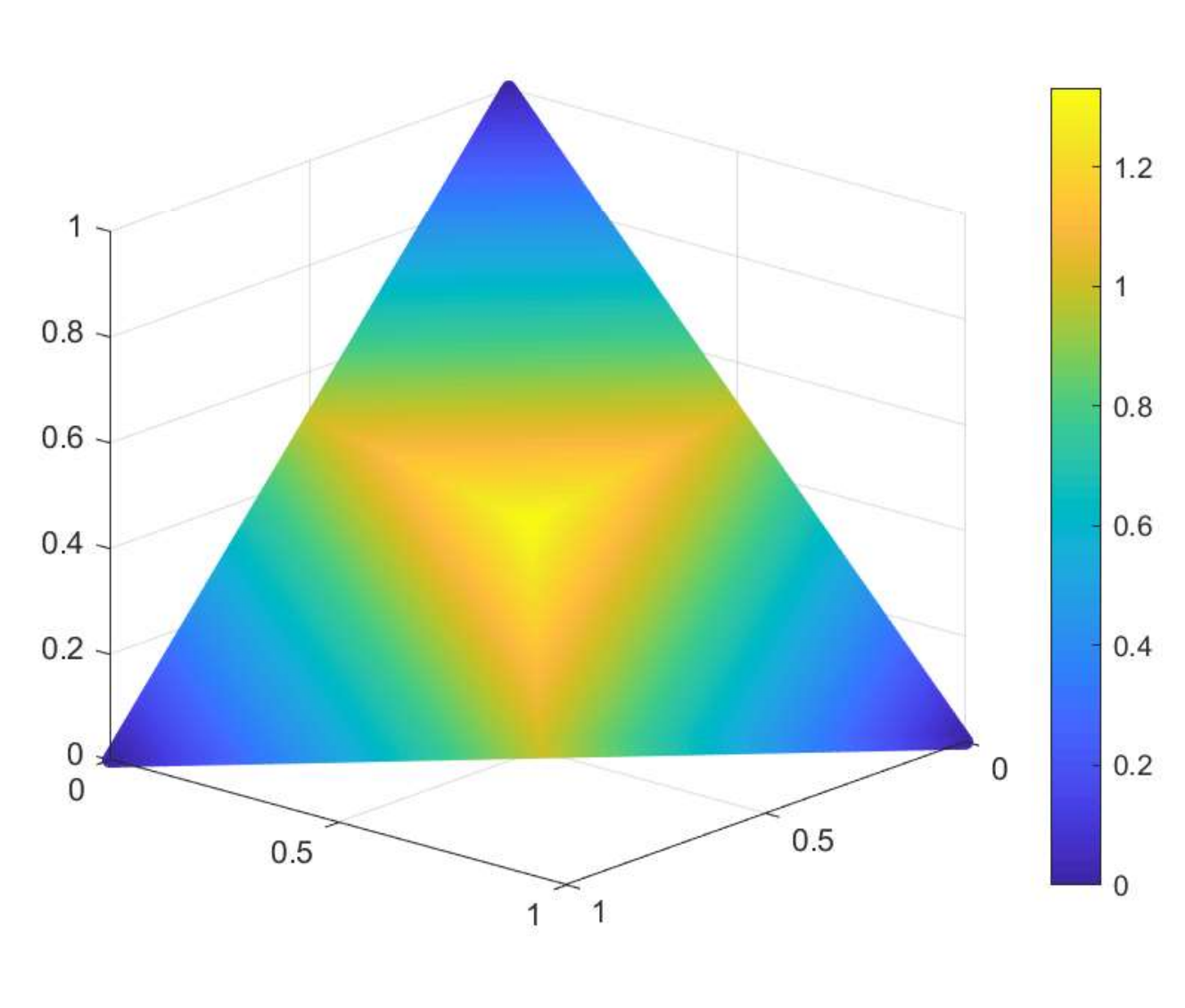}
		\label{fig:beta}
	}
	\caption{Illustrations of the value distributions of $\alpha$ (a) and $\beta$ (b) on the linear hyperplane}
	\label{fig:alphabeta}
\end{figure}
Given a reference point $\textbf{w}^i$, its minimum term reflects the distance between the reference point and the edge of the linear hyperplane, and the smaller the value is, the closer it is. The value $\alpha_i$ of each reference point is calculated by Equation \ref{eqn:alpha}, and their distribution on the linear hyperplane is shown in Fig. \ref{fig:alphabeta}(a).

\begin{figure}[htbp]
	\centering
	\includegraphics[width =0.35\linewidth]{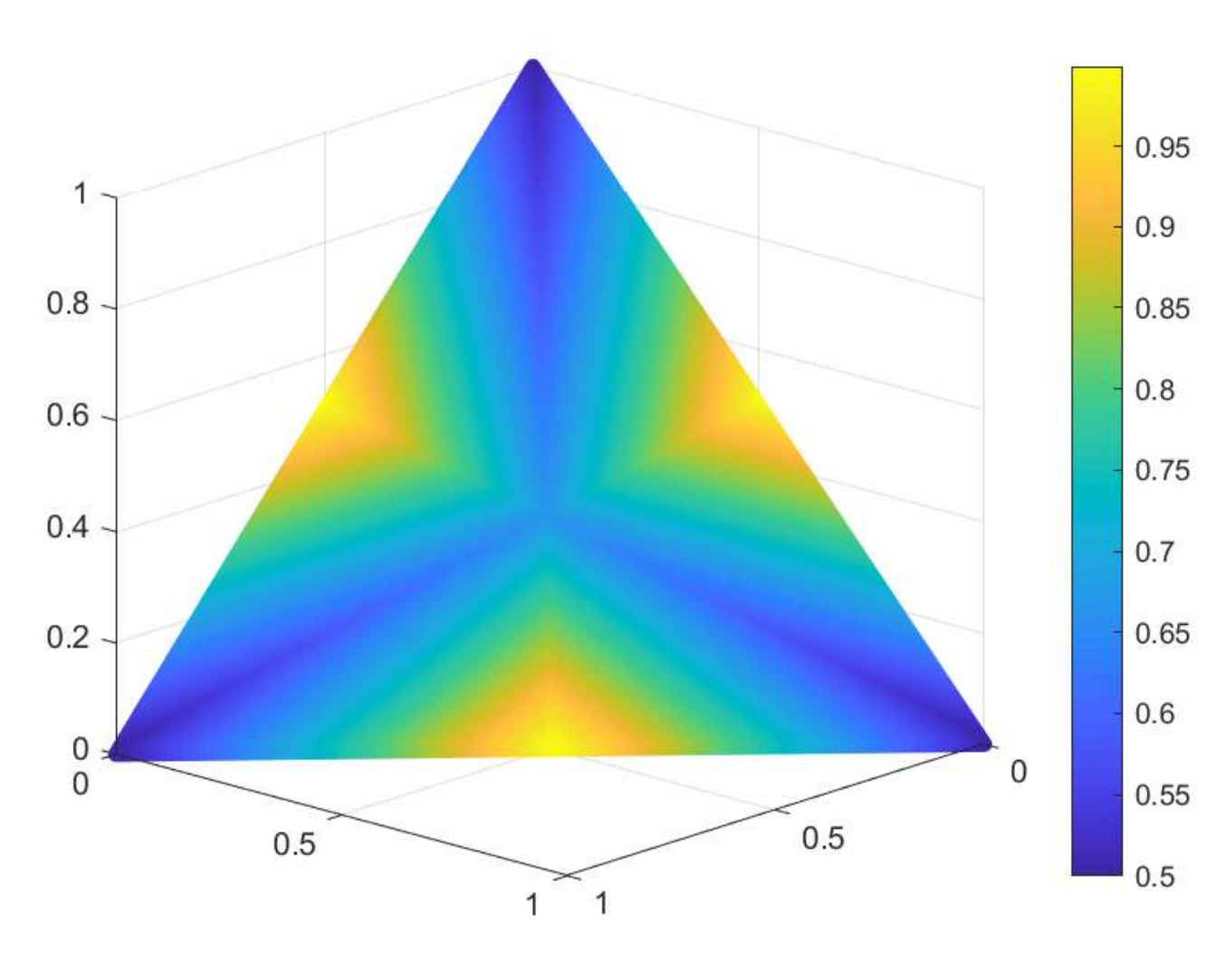}
	\caption{An illustration of the value distribution of $r$ on the linear hyperplane.}
	\label{fig:DC}
\end{figure}

On the other hand, to inherit the uniformity of the NBI method, especially for the ability to deal with the steep frontiers. We need to ensure that the closer to the vertex region of the hyperplane the reference vector is, the smaller its rotation angle is. Here, we define $\beta_i$ as follows:
\begin{eqnarray}
\beta_i = (1 - max_{j=1}^m\{w_j^i\})\times 2. \label{eqn:beta}
\end{eqnarray}
Given a reference point $\textbf{w}^i$, its maximum term reflects the distance between the reference point and the vertex of the linear hyperplane, and the larger the value is, the closer it is. The values  $\beta_i$ of  each reference point are calculated by Equation \ref{eqn:beta}, and their distribution on the linear hyperplane is shown in Fig. \ref{fig:alphabeta}(b). For every reference point on the unit hyperplane, the calculation of formula \ref{eqn:DC} is performed to obtain the value distribution of the rotation coefficient as shown in Fig. \ref{fig:DC}. 
And Algorithm \ref{alg:rotation_factor} gives the construction procedure of the set $\textbf{r}$ of reference-point-based rotation factors associated with the boundary subproblems.

\begin{algorithm}
	\caption{generate\_rotation\_factor }
      \label{alg:rotation_factor}
	\KwIn{
		 $\textbf{w}^i$: the reference point of the $i$-th boundary subproblem;
		}
	\KwOut{$r$: the set of reference-point-based rotation factors associated with the boundary subproblems; }
	\ForEach{boundary reference point $\textbf{w}^i$\label{alg:code:DF_1}}
	{
		{$\alpha_i \leftarrow 1-min_{j=1}^m\{w_j^i\} \times m$\label{alg:code:DF_2}}\;
		{$\beta_i \leftarrow   (1 - max_{j=1}^m\{w_j^i\})\times 2$\label{alg:code:DF_3}}\;
		{$r_i \leftarrow \frac{\alpha_i+\beta_i}{2}$\label{alg:code:DF_4}}\;
	}
	{\Return{$r=\{r_i\}$}\label{alg:code:DF_6}}\;
\end{algorithm}

Previously, we set $k_m$ as the parameter only affected by the objective dimension $m$, it is mainly because the method of reference point generation is affected by the objective dimension $m$. When the size of objective dimensions is not more than 5, we generate single-layer reference points according to Das and Dennis's systematic approach  \cite{DasAndDenns}, otherwise, we generate two-layer reference points according to Deb and Jain’s method  \cite{NSGA-III}. When the objective dimension is greater than 5, we need to use different methods to deal with the boundary layer subproblems and the inner layer subproblems, respectively. It is because the generation of two-layer reference points destroys the design premise that the reference points are uniform. In this case, the value of $k_m$ does not follow the same premise like that in another case. Therefore, $k_m$ is defined as equation (\ref{eqn:CoD_3}):
\begin{eqnarray}
k_m = \frac{m}{1+\textit{e}^{-m  (m-5.5)}}.
\label{eqn:CoD_3}
\end{eqnarray}

\section{The proposed algorithm}
\label{sec:CoD-DEA}
In this section, the overall framework of CoDEA is first introduced. Afterwards, its CoD-based environmental selection mechanism is described in detail.

\subsection{General framework}

The basic framework of CoDEA is similar to $\theta$-DEA  \cite{7820439} and MP-DEA  \cite{8931629}. The pseudocode of CoDEA is described in Algorithms \ref{alg:Basic Framework} and Algorithms \ref{alg:ES}. First, in Algorithm \ref{alg:Basic Framework}, a set of reference points $W$
is generated using Das and Dennis's systematic approach if the number of objectives, $m$, is not more than five. Otherwise, the Deb and Jain’s method is adopted.
Then the rotation coefficient set $r$ of the boundary reference points is calculated by Algorithm \ref{alg:rotation_factor}. Thereafter, a population $P$ with size $N$ and the ideal point $z^*$ are initialized successively. 
While the current number of generations $t$ does not reach the allowed maximum number of generations, $G_{max}$, the evolutionary operations in lines 7-10 of Algorithm \ref{alg:Basic Framework} are performed iteratively. 
Specifically, the offspring population $Q$ is first created and it is merged together with the parent population $P$ into a union population $U$. Then a collaborative decomposition-based environment selection mechanism as shown in Algorithm \ref{alg:ES} is designed to extract the next generation population $P$ from $U$. This mechanism is depicted in detail in Section \ref{sec:CoDES}.

\begin{algorithm}
	\caption{CoDEA}
      \label{alg:Basic Framework}
	\KwIn{
		  $N$: the population size; $G_{max}$: the allowed maximum number of generations; 
		}
	\KwOut{ $P$: the population;}
	{ Initialize the reference point set W and the number of boundary layer reference points $max\_ex$\label{alg:code:CoD_1}}\;
	{$r \leftarrow generate\_rotation\_factor(W,max\_ex)$\label{alg:code:CoD_2}}\;
	{ Initialize population P\label{alg:code:CoD_3}}\;
	{ Initialize ideal point $z^*$ and nadir point $z^{nadir}$\label{alg:code:CoD_4}}\;
	{ $t \leftarrow 0$ \label{alg:code:CoD_5}}\;
	\While{$t < G_{max}$\label{alg:code:CoD_6}}	
	{
		{$Q \leftarrow create\_offspring\_population(P)$\label{alg:code:CoD_7}}\;
		{$U \leftarrow P \cup Q$\label{alg:code:CoD_8}}\;
		{
		$P,z^*\leftarrow CoD\_environmental\_selection(U,W,z^*, z^{nadir},r)$\label{alg:code:CoD_9}}\;
		{$t \leftarrow t + 1$\label{alg:code:CoD_10}}\;
	}
	{\Return{$P$}\label{alg:code:CoD_12}}\;
\end{algorithm}

%
%
%
\begin{algorithm}
	\caption{CoD-based environmental selection}
      \label{alg:ES}
	\KwIn{
		   $U$: the union population; $W$: the reference points; $z^*$: the ideal points; $z^{nadir}$: the nadir points; $r$: the rotation factor;
		}
	\KwOut{ Population $P$, $z^*$;}
	 {$S \leftarrow nondominated\_sorting(U)$\label{alg:code:ES_1}}\;
	 {Update ideal point $z^*$ and nadir point $z^{nadir}$\label{alg:code:ES_2}}\;
	 {Normalize the population $S$\label{alg:code:ES_3}}\;
	 {Associate the individual in $S$ with the corresponding subproblem in $W$\label{alg:code:ES_4}}\;
	 {$S' \leftarrow CoD\_ranking(S,W)$\label{alg:code:ES_5}}\;
	 {$P \leftarrow \emptyset$\label{alg:code:ES_6}}\;
	 {$i \leftarrow 0$\label{alg:code:ES_7}}\;
	\While{$|P|+|R_i| < N$\label{alg:code:ES_8}}
	{
		{ $P \leftarrow P\cup R_i$\label{alg:code:ES_9}}\;
		{$i \leftarrow i + 1$\label{alg:code:ES_10}}\;
	}
	{
	The remaining $N-|P|$ individuals are randomly selected from the last accepted rank $R_i$ to form a set $R'_i$
	\label{alg:code:ES_12}}\;
	{$P \leftarrow P\cup R'_i$\label{alg:code:ES_13}}\;
	{\Return{$P,z^*$}\label{alg:code:ES_14}}\;
\end{algorithm}

\subsection{CoD-based environmental selection}
\label{sec:CoDES}


The procedure of the CoD-based environmental selection mechanism is shown in Algorithm \ref{alg:ES}.
This selection mechanism
first conducts a non-dominated sorting on the population $R$.
Afterwards, the individuals in the first left non-dominated level are moved into a set, $S$, until the size of $S$ is not less than $N$.
Then the ideal point $z^*$ and nadir point is updated.
All the individuals in $S$ are normalized by the same normalization method adopted in NSGA-III \cite{NSGA-III}.  After normalization, every individual in the set $S$ is associated with the closest reference line in the set $W$ through an association mechanism.

CoDEA adopts a global association strategy based on vertical distance, different from the local association strategy in the classic MOEA/D. In the normalized objective space, the origin is the mapping of the ideal point $z^*$. Then each individual $\textbf{x}$ is associated with the reference vector $\textbf{w}^j$ with the shorest vertical distances to $\textbf{f}(\textbf{x})$ in the normalized objective space where
\begin{eqnarray}
j= \arg\min_{i\in\{1,2,\cdots,N\}} d_2(\textbf{f}(\textbf{x}),\textbf{w}^i). \label{eqn:norm_distance} 
\end{eqnarray}




After the global association procedure, a CoD-based ranking strategy and an angle-based ranking strategy are designed to sort all the individuals associated with every boundary reference vector and those with every inner one, respectively.
Thus these two ranking strategies divide the population $S$ into $n_R$ different CoD-based ranks $R_i$, $i=1, 2, \cdots,n_R$. Essentially, the $i$-th CoD-based rank $R_i$ of the population includes the individual in the $i$-th place associated with every reference vector.
Then the next population $P$ is filled with the top $l-1$ CoD-based ranks as described in lines \ref{alg:code:ES_8}–\ref{alg:code:ES_10} of Algorithm \ref{alg:ES} where $\sum^{l-1}_{i = 1}{|R_i|}<N$ and $\sum^{l}_{i = 1}{|R_i|}\ge N$. Finally, the remaining individuals of $P$ are randomly selected  from the last accepted rank $R_l$ to form a complete population $P$.

\subsection{CoD-based ranking and angle-based ranking}






In the CoD-based environmental selection mechanism, the individuals associated with the boundary subproblems and those with the inner ones are, respectively, sorted by two different strategies.
A CoD-based ranking strategy is designed to sort all the individuals associated with every boundary subproblem in terms of the CoD aggregation function of the collaborative decomposition method.
To be specific,
given two solutions $\textbf{x}$ and $\textbf{y}$ associated with a boundary reference vector $\textbf{w}^i$, $\textbf{x}$ is said to take precedence over $\textbf{y}$ in the CoD-based ranking strategy if and only if
\begin{eqnarray}
g^{CoD}(\textbf{x}|\textbf{w}^i) < g^{CoD}(\textbf{y}|\textbf{w}^i). \label{eqn:CoDR}
\end{eqnarray}
Since the CoD-based ranking strategy integrates both advantages of the PBI and NBI methods, it helps all the boundary subproblems to obtain a relatively uniform and wide distribution of solutions near the boundaries of frontiers. 

In addition, an angle-based ranking strategy is
designed to sort all the individuals associated with every inner subproblem.
Given any two solutions $\textbf{x}$ and $\textbf{y}$ associated with an inner reference vector $\textbf{w}^i$, $\textbf{x}$ is said to take precedence over $\textbf{y}$ in the angle-based ranking strategy if and only if
\begin{eqnarray}
angle(\textbf{f}(\textbf{x}),\lambda) <
angle(\textbf{f}(\textbf{y}),\lambda). \label{eqn:CoD_4}
\end{eqnarray}
where $\lambda = (\frac{1}{m},\frac{1}{m},\cdots,\frac{1}{m}) $ is the vector pointing to the center of the linear hyperplane from $z^*$,  and $angle(\textbf{f}(\textbf{x}),\lambda)=arccos(|\frac{\textbf{f}(\textbf{x})\cdot \lambda}{||\textbf{f}(\textbf{x})||\cdot ||\lambda||}|)$ denotes the acute angle between objective vector $\textbf{f}(\textbf{x})$ and $\lambda$ in the normalized objective space.
Essentially, this strategy aims to push the individuals associated with each inner subproblem as far away from the center of frontier as possible by applying the angle measure.
Therefore, this strategy relieves the phenomenon of population gathering and maintain the uniformity of population distribution in the inner regions of frontiers.

\subsection{Analysis of computational complexity } 

Since the basic framework of CoDEA is similar to $\theta$-DEA  \cite{7820439} and MP-DEA  \cite{8931629}, the computational complexity of CoDEA can be analyzed in a similar way. The main difference between CoDEA, $\theta$-DEA, and MP-DEA at every evolutionary generation is only the niche strategy of ranking the solutions associated with each subproblem. The CoD-based ranking strategy in CoDEA does not bring additional computational costs. Therefore, similar to $\theta$-DEA and MP-DEA, CoDEA has a computational complexity of  $O(mN^2)$ at every evolutionary generation.

\section{Experimental study}
\label{sec:EXPERIMENTAL}

This section is devoted to the experimental design for investigating the performance of the proposed CoDEA. First, the test problems and the performance measure used in our experiments are given. Then, we briefly introduce five state-of-the-art algorithms that are employed for comparison. After that, the experimental settings adopted in our paper are described. Finally, we discuss the obtained results are presented and discussed.

\subsection{Test problems and performance measure}

\begin{table}[htbp]
  \centering
  \caption{Settings of $G_{max}$ for different test instances.}
    \begin{tabular}{cccccc}
    \toprule
    \multicolumn{1}{l}{\multirow{2}[2]{*}{     Problem}} & \multirow{2}[2]{*}{m = 3} & \multirow{2}[2]{*}{m = 5} & \multirow{2}[2]{*}{m = 8} & \multirow{2}[2]{*}{m = 10} & \multirow{2}[2]{*}{m = 15} \\
          &       &       &       &       &  \\
    \midrule
    DTLZ1 & \multirow{2}[2]{*}{36800} & \multirow{2}[2]{*}{127200} & \multirow{2}[2]{*}{117000} & \multirow{2}[2]{*}{276000} & \multirow{2}[2]{*}{204000} \\
    CDTLZ1 &       &       &       &       &  \\
    \midrule
    DTLZ2 & \multirow{2}[2]{*}{23000} & \multirow{2}[2]{*}{74200} & \multirow{2}[2]{*}{78000} & \multirow{2}[2]{*}{207000} & \multirow{2}[2]{*}{136000} \\
    CDTLZ2 &       &       &       &       &  \\
    \midrule
    DTLZ3 & \multirow{2}[2]{*}{92000} & \multirow{2}[2]{*}{212000} & \multirow{2}[2]{*}{156000} & \multirow{2}[2]{*}{414000} & \multirow{2}[2]{*}{272000} \\
    CDTLZ3 &       &       &       &       &  \\
    \midrule
    DTLZ4 & \multirow{2}[2]{*}{55200} & \multirow{2}[2]{*}{212000} & \multirow{2}[2]{*}{195000} & \multirow{2}[2]{*}{552000} & \multirow{2}[2]{*}{408000} \\
    CDTLZ4 &       &       &       &       &  \\
    \midrule
    WFG1-9 & 92000 & 265000 & 234000 & 552000 & 405000 \\
    \bottomrule
    \end{tabular}%
  \label{tab:GMAX}%
\end{table}%

For the purposes of comparison, two well-known benchmark test suites for MaOPs, DTLZ  \cite{2002Scalable} and Walking Fish Group (WFG) \cite{WFG}, are involved in our experiments. To compute the quality indicators reliably, we only consider DTLZ1–4 problems for the DTLZ test suite in our experiments. Moreover, to increase the universality of the test problems, we construct four test problems with convex PFs with reference to   \cite{NSGA-III}, referred to as Convex\_DTLZ1-4 or CDTLZ1-4 problems. Five different numbers of objectives, i.e., $m =3$, 5, 8, 10, and 15, are considered in our experiments. 

Since the hypervolume (HV) indicator \cite{1999Multiobjective,2003Performance} can fairly reflect the convergence and distribution of a set of solutions at the same time, it is widely applied to evaluate the solution sets obtained by MaOEAs \cite{2009Theory}. Hence, the HV indicator is also adopted as the performance measure in our experiments. As recommended in \cite{YING202097}, the reference point for calculating HV is set to $(1.1, 1.1, \cdots, 1.1)$ in the normalized objective space and all the HV values in our experiments are finally standardized by dividing $1.1^m$ . 



The Wilcoxon rank-sum test \cite{Wilcoxon} is usually used to calculate the significance of the performances of two random algorithms. In our experiments, the Wilcoxon rank-sum test is applied in pairs at the significance level of 0.05. We run each pair (algorithm and problem) for 21 times, and the results of Wilcoxon rank-sum test are presented in the form of ($-$: significantly worse), ($+$: significantly better) and ($\approx$: statistically similar).


\subsection{ Algorithms under comparison}

In this paper, CoDEA is compared with five popular MaOEAs: VaEA \cite{2017A}, RVEA  \cite{7386636}, NSGA-III   \cite{NSGA-III}, $\theta$-DEA   \cite{7820439}, and MP-DEA   \cite{8931629}. NSGA-III is an improved version of NSGA-II. These considered algorithms are summarized in the following. It introduces a set of uniformly distributed reference points on the basic framework of NSGA-II. And it uses these reference points to guide the distribution of the population.
RVEA adopts the APD method to adaptively control the population convergence and population distribution in the process of population evolution. This method is committed to accelerating the population convergence in the early stage of population evolution and focusing on the control of population distribution in the later stage of population evolution.
VaEA is an EA based on vector angle. After locating the solutions in the direction of the target axis, the 	diversity is maintained by the principle of the maximum angle between solutions, and the convergence is maintained by excluding the 			worse one between the two solutions with the minimum angle. $\theta$-DEA uses a $\theta$-dominated method that sorts the  solutions associated with the same subproblem according to the PBI function after the association operation. Because the elite individuals are  picked in every generation, both the diversity and convergence of population are well maintained in $\theta$-DEA. In our experiments, the parameter $\theta$ is set to $\theta = 10^6 $ for the axis reference vectors and $\theta = 5$ for the other reference vectors.
MP-DEA is an improved version of  $\theta$-DEA. Based on the framework of  $\theta$-DEA, MP-DEA adopts the MPR method instead of the PBI method to perform the internal sorting operation for the solutions associated with each subproblem.
In our experiments, all the six MaOEAs are implemented and performed on a MATLAB-based evolutionary multi-objective optimization platform PlatEMO \cite{Tian2017}. 



\subsection{  Experimental settings}





The settings of several common parameters for all algorithms are listed as follows.
The simulated binary crossover (SBX) with distribution index $\eta_c$=30 and the polynomial mutation with distribution index $\eta_m$=30 are used in all the considered algorithms. The crossover probability $p_c$ and mutation probability $p_m$ are set to 1.0 and $1/n$, respectively. Suggested by Deb et al. \cite{NSGA-III}, the number of divisions for reference vectors is set as $H=12$ and $H=6$ for $m=3$ and $m=5$, respectively. Meanwhile, the numbers of divisions for the boundary and inner layers of reference vectors are $(H_1, H_2)=(3, 2)$, $(3, 2)$, and $(2, 1)$, respectively, for $m=8$, 10, and 15. As a result, the population sizes for all the MaOEAs except for NSGA-III are $N=91$, 210, 156, 275, and 135, respectively, for $m=3$, 5, 8, 10, and 15. The population sizes for NSGA-III are set as the least multiple of four not less than the corresponding sizes of the other MaOEAs.
The termination condition for each algorithm is specified in the form of the maximum number $G_{max}$ of generations. The predefined maximum numbers of generations of all algorithms for different test instances with different numbers of objectives are listed in Table \ref{tab:GMAX}.
For the sake of fairness, each algorithm is performed 21 times independently for each test instance.



\subsection{Comparisons on the DTLZ test suite with three and five objectives}
\begin{table*}[htbp]
  \centering
  \caption{Median HV values and IQRs (in brackets) obtained by six algorithms on the DTLZ1-4 and CDTLZ1-4 test instances with three and five objectives.}
	\footnotesize
	\renewcommand\arraystretch{0.6}
    \begin{tabu} to 1 \linewidth{|X[2.5,c]|X[0.4,c]|X[3.5,c]|X[3.5,c]|X[3.5,c]|X[3.5,c]|X[3.5,c]|X[3.5,c]|}
    \toprule
    Pro. & m     & RVEA  & NSGA-III & VaEA  & $\theta -$DEA  & MP-DEA & CoDEA \\
    \midrule
    \multirow{2}[4]{*}{DTLZ1} & 3     & 8.4079e-1 (2.37e-3) $\approx$ & \textbf{8.4087e-1 (1.13e-3) $\approx$} & 8.1679e-1 (4.68e-2) - & 8.4052e-1 (1.82e-3) $\approx$ & 8.1308e-1 (9.28e-2) - & 8.4054e-1 (1.08e-3) \\
\cmidrule{2-8}          & 5     & 9.7981e-1 (1.66e-4) $\approx$ & \textbf{9.7984e-1 (2.96e-4) $\approx$} & 9.2145e-1 (5.55e-2) - & 9.7983e-1 (2.36e-4) $\approx$ & 9.6513e-1 (7.54e-2) - & 9.7979e-1 (1.84e-4) \\
    \midrule
    \multirow{2}[4]{*}{DTLZ2} & 3     & 5.5905e-1 (2.61e-4) - & 5.5923e-1 (1.51e-4) - & 5.5424e-1 (9.93e-4) - & 5.5930e-1 (1.21e-4) - & 5.5679e-1 (6.31e-4) - & \textbf{5.6132e-1 (8.29e-4)} \\
\cmidrule{2-8}          & 5     & 8.1216e-1 (6.13e-4) $\approx$ & 8.1197e-1 (5.52e-4) $\approx$ & 7.9210e-1 (2.71e-3) - & 8.1212e-1 (4.23e-4) $\approx$ & 8.0648e-1 (1.11e-3) - & \textbf{8.1217e-1 (5.92e-4)} \\
    \midrule
    \multirow{2}[4]{*}{DTLZ3} & 3     & 5.5721e-1 (1.91e-3) - & 5.5618e-1 (5.50e-3) - & 5.5223e-1 (3.68e-3) - & 5.5385e-1 (4.22e-3) - & 5.5216e-1 (6.43e-3) - & \textbf{5.5976e-1 (5.00e-3)} \\
\cmidrule{2-8}          & 5     & \textbf{8.1224e-1 (6.99e-4) +} & 8.1031e-1 (2.25e-3) - & 6.0096e-1 (8.46e-2) - & 8.1156e-1 (1.37e-3) $\approx$ & 5.4416e-1 (3.89e-1) - & 8.1179e-1 (1.44e-3) \\
    \midrule
    \multirow{2}[4]{*}{DTLZ4} & 3     & 5.5958e-1 (5.35e-5) $\approx$ & 5.5955e-1 (1.10e-4) - & 5.5412e-1 (2.06e-3) - & 5.5951e-1 (2.34e-1) - & 5.5665e-1 (2.28e-1) - & \textbf{5.6186e-1 (2.29e-1)} \\
\cmidrule{2-8}          & 5     & 8.1250e-1 (3.93e-4) - & 8.1243e-1 (5.43e-4) $\approx$ & 7.8859e-1 (5.49e-3) - & 8.1258e-1 (4.98e-4) $\approx$ & 8.0636e-1 (8.91e-4) - & \textbf{8.1269e-1 (4.17e-4)} \\
    \midrule
    \multirow{2}[4]{*}{CDTLZ1} & 3     & 9.8373e-1 (8.28e-4) - & 9.8330e-1 (4.43e-4) - & 9.4431e-1 (3.69e-2) - & 9.7721e-1 (1.21e-3) - & 9.7240e-1 (4.11e-3) - & \textbf{9.8442e-1 (2.58e-4)} \\
\cmidrule{2-8}          & 5     & 9.9931e-1 (5.80e-4) - & 9.9982e-1 (1.93e-5) $\approx$ & 9.8723e-1 (2.33e-2) - & 9.9507e-1 (1.09e-3) - & 9.7775e-1 (5.52e-3) - & \textbf{9.9983e-1 (2.25e-5)} \\
    \midrule
    \multirow{2}[4]{*}{CDTLZ2} & 3     & 9.5858e-1 (9.05e-4) - & 9.5829e-1 (3.03e-4) - & 9.5314e-1 (1.74e-3) - & 9.5758e-1 (6.40e-4) - & 9.5232e-1 (3.12e-3) - & \textbf{9.6110e-1 (1.20e-4)} \\
\cmidrule{2-8}          & 5     & 9.9725e-1 (9.72e-4) - & \textbf{9.9942e-1 (3.18e-5) $\approx$} & 9.9892e-1 (1.30e-4) - & 9.9351e-1 (1.16e-3) - & 9.7556e-1 (5.53e-3) - & 9.9941e-1 (4.40e-5) \\
    \midrule
    \multirow{2}[4]{*}{CDTLZ3} & 3     & 9.5914e-1 (1.60e-3) $\approx$ & 9.5734e-1 (9.91e-4) - & 9.2321e-1 (4.87e-2) - & 9.5644e-1 (8.93e-4) - & 9.5176e-1 (3.89e-3) - & \textbf{9.5950e-1 (2.49e-3)} \\
\cmidrule{2-8}          & 5     & 9.9892e-1 (3.67e-4) - & 9.9933e-1 (8.57e-5) - & 9.9165e-1 (5.38e-3) - & 9.9378e-1 (1.23e-3) - & 9.7608e-1 (7.71e-3) - & \textbf{9.9935e-1 (6.25e-5)} \\
    \midrule
    \multirow{2}[4]{*}{CDTLZ4} & 3     & 9.5916e-1 (1.07e-3) - & 9.5817e-1 (1.71e-4) - & 9.5435e-1 (1.50e-3) - & 9.5752e-1 (1.07e-1) - & 9.5734e-1 (3.32e-2) - & \textbf{9.6123e-1 (1.22e-1)} \\
\cmidrule{2-8}          & 5     & \textbf{9.9934e-1 (1.11e-4) $\approx$} & 9.9932e-1 (3.32e-5) - & 9.9901e-1 (1.26e-4) - & 9.9516e-1 (2.02e-3) - & 9.9689e-1 (2.68e-3) - & 9.9933e-1 (3.72e-5) \\
	
    \midrule
    \multicolumn{2}{|c|}{+/-/$\approx$} & 1/9/6 & 0/10/6 & 0/16/0 & 0/11/5 & 0/16/0 &  \\

    \bottomrule
    \end{tabu}%
  \label{tab:1}%
\end{table*}%

\begin{figure}[htbp]
	\centering
	\subfloat[]{
		\includegraphics[width = 0.29\linewidth]{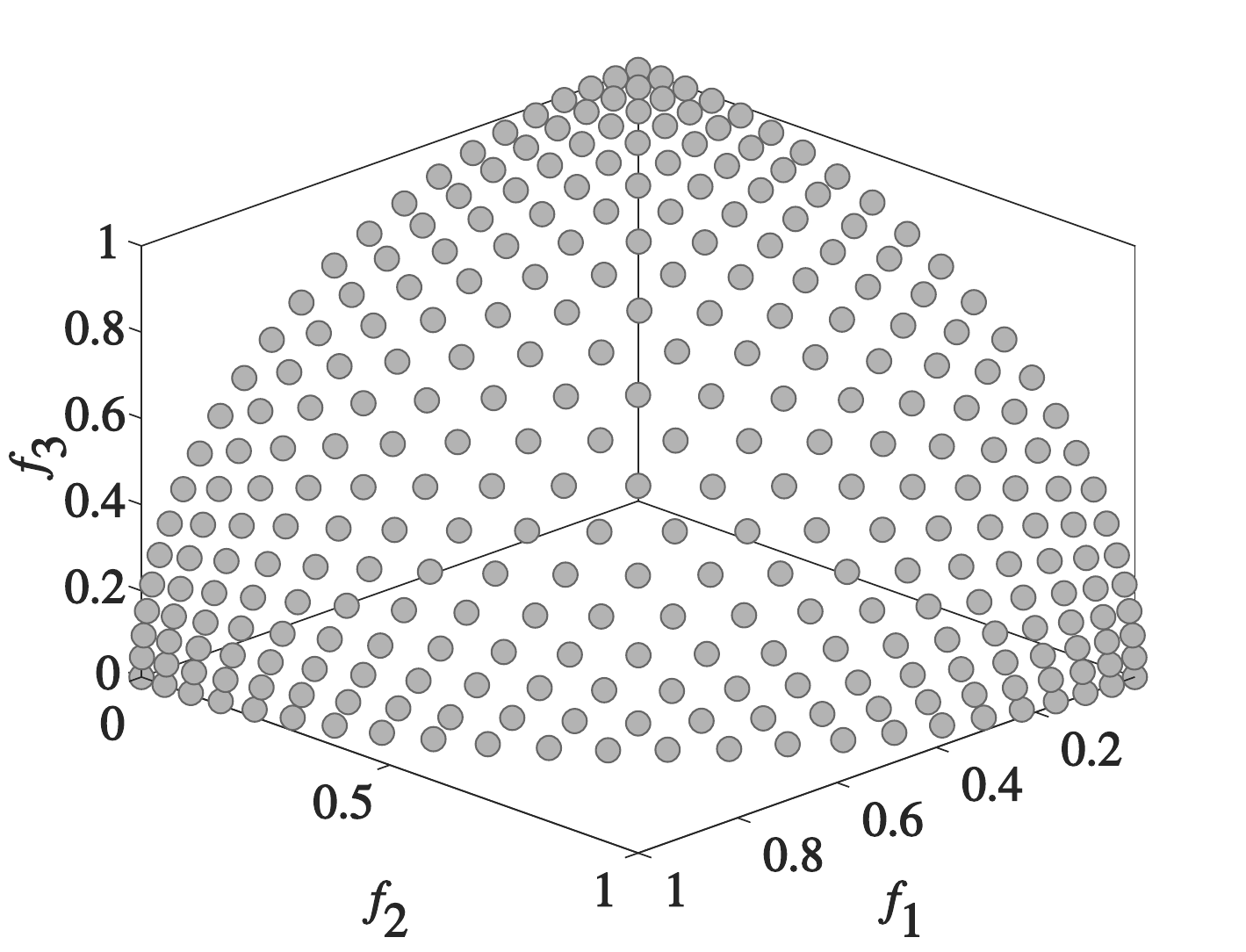}
		\label{fig:RVEA_DTLZ2_3}
	}
	\subfloat[]{
		\includegraphics[width = 0.29\linewidth]{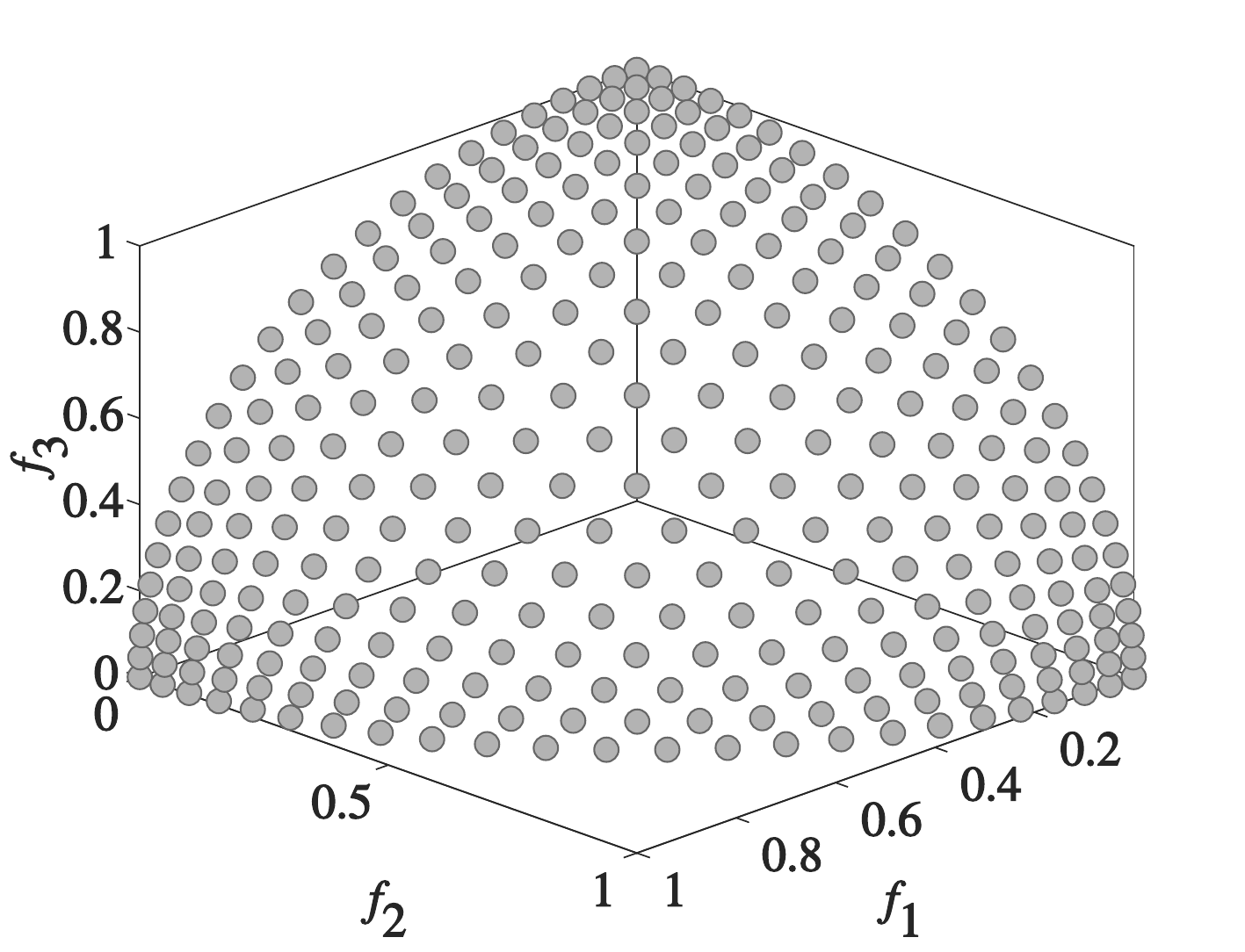}
		\label{fig:NSGA3_DTLZ2_3}
	}
	\subfloat[]{
		\includegraphics[width = 0.29\linewidth]{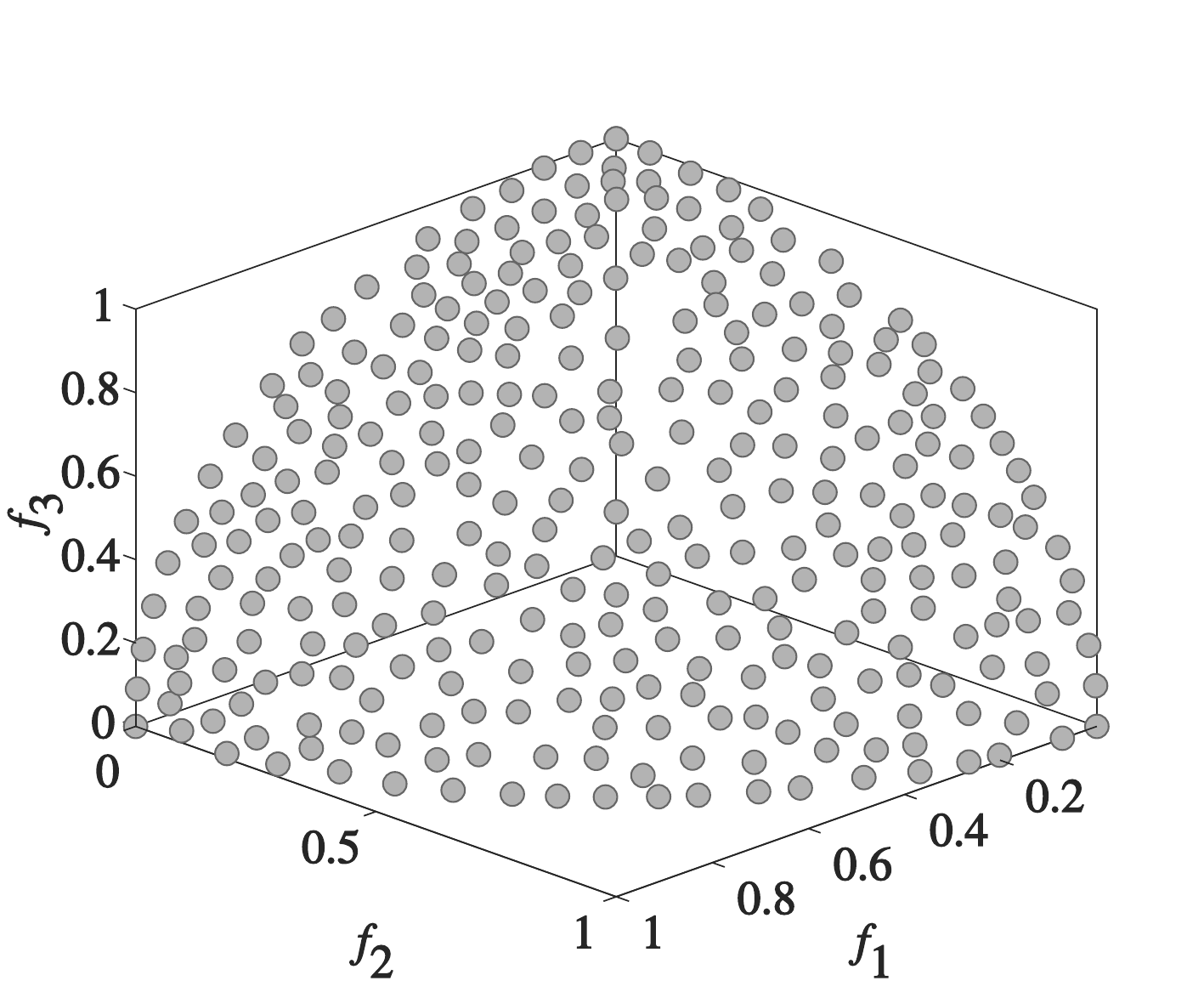}
		\label{fig:VaEA_DTLZ2_3}
	}\\
	\subfloat[]{
		\includegraphics[width = 0.29\linewidth]{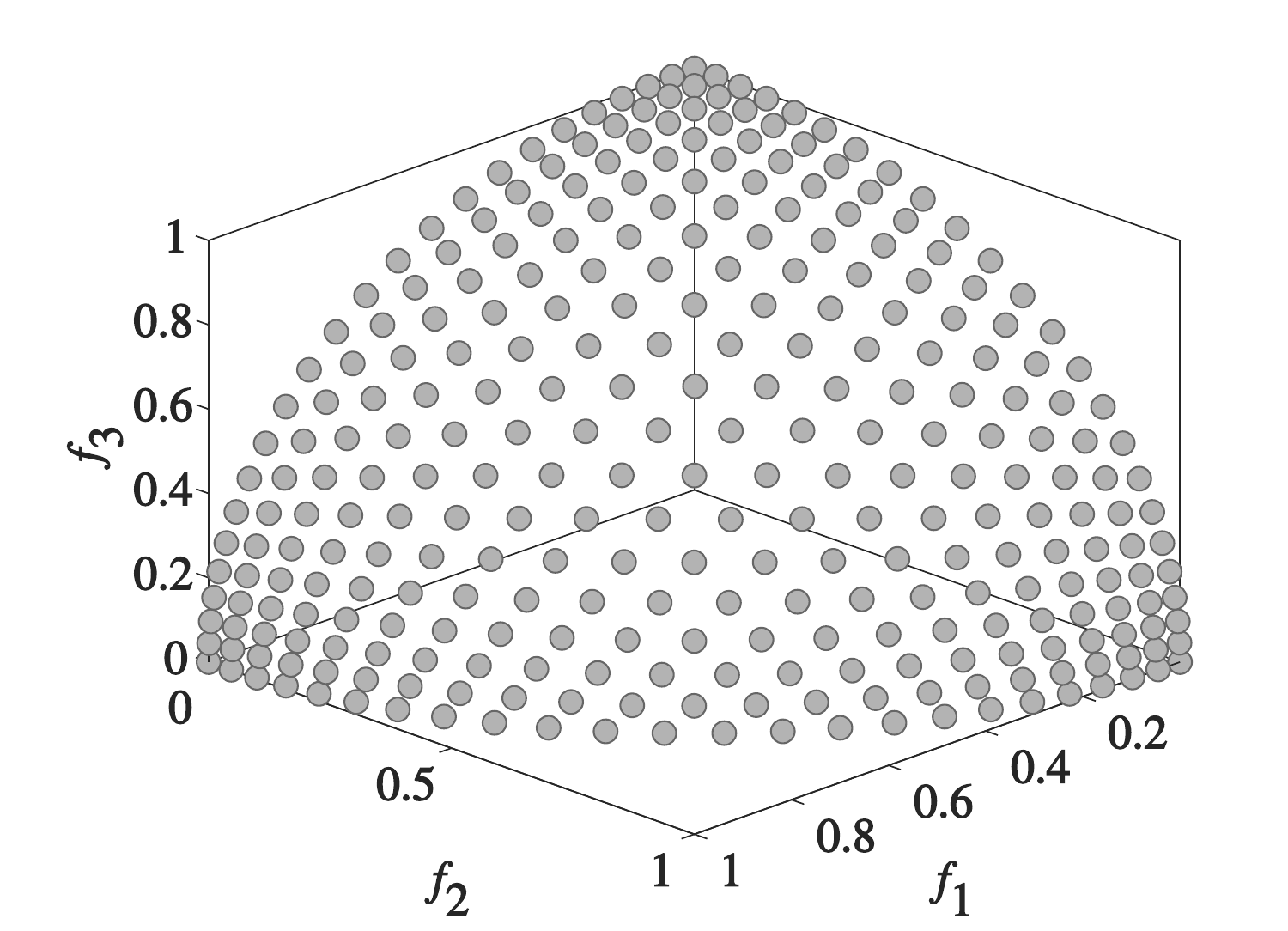}
		\label{fig:tDEA_DTLZ2_3}
	}
	\subfloat[]{
		\includegraphics[width = 0.29\linewidth]{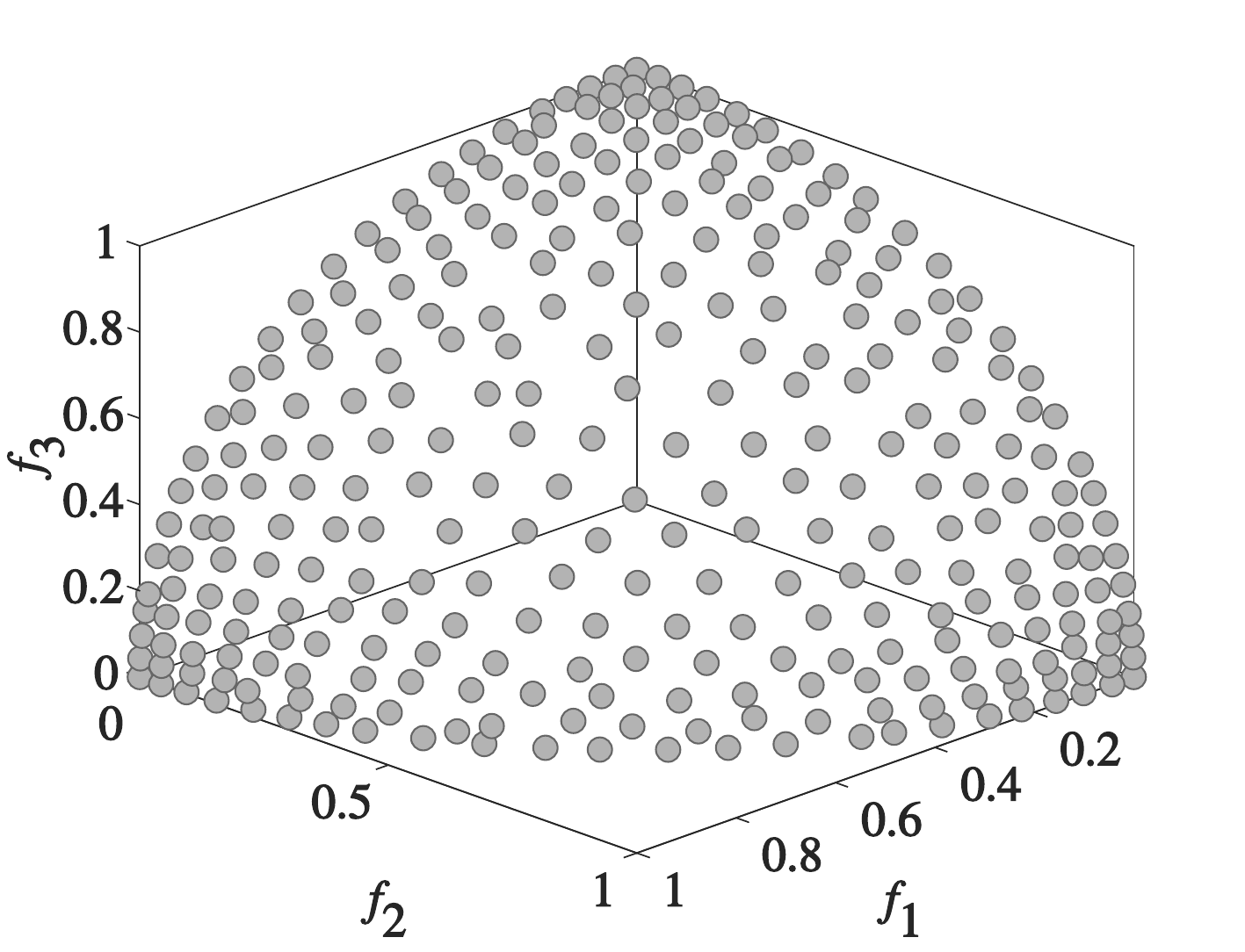}
		\label{fig:MP_DEA_DTLZ2_3}
	}
	\subfloat[]{
		\includegraphics[width = 0.29\linewidth]{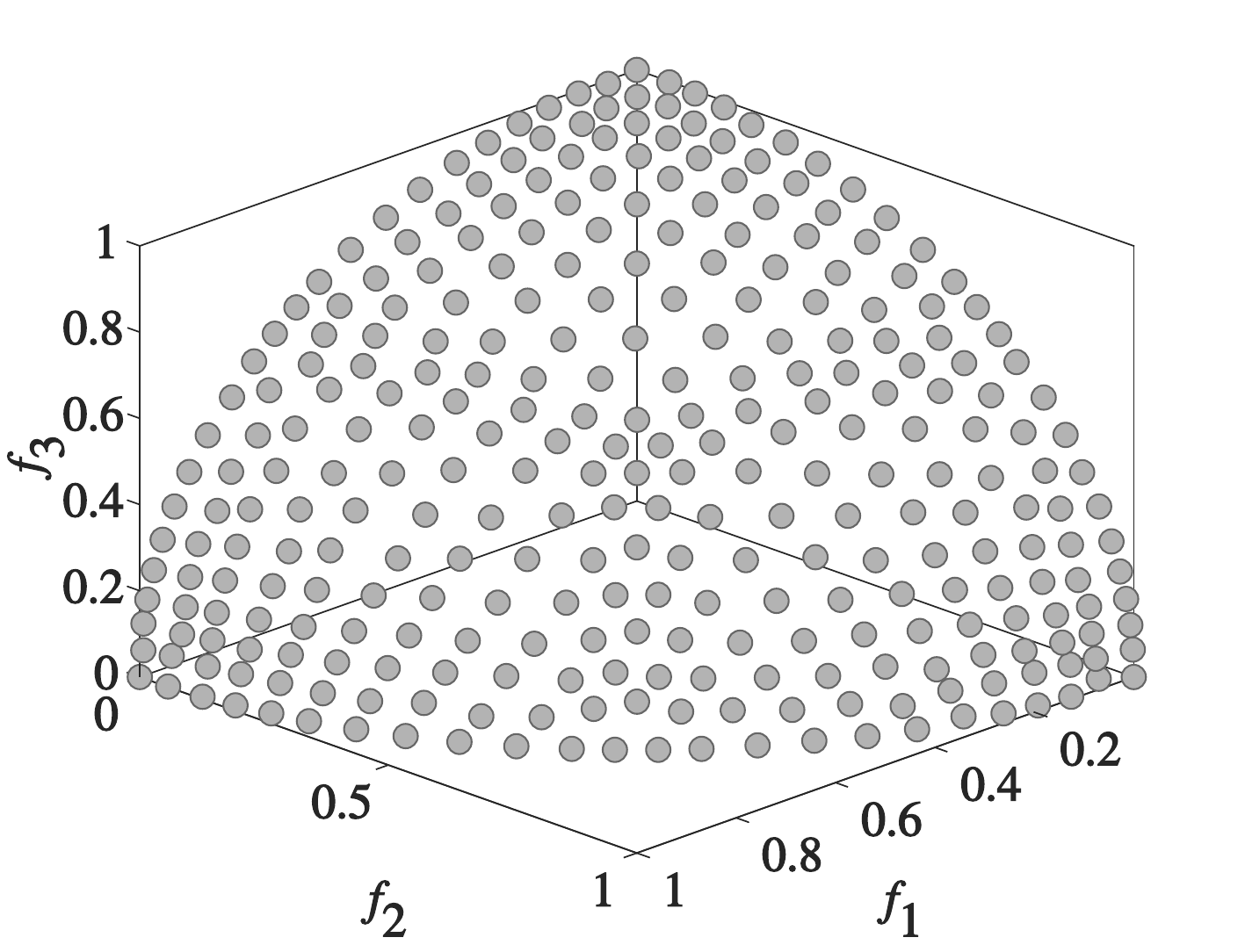}
		\label{fig:CoD_DTLZ2_3}
	}
	\caption{The distributions of frontiers achieved by the six algorithms on the tri-objective DTLZ2 test instance. (a) RVEA, (b) NSGA-III, (c) VaEA, (d) $\theta$-DEA, (e) MP-DEA, and (f) CoDEA.}
	\label{fig:PF_3D_DTLZ2}
\end{figure}


In this section, the experimental results are presented to validate the performance and effectiveness of CoDEA. According to different test suites and different numbers of objectives, these experiments are divided into four groups: (1) comparisons on the DTLZ test suite with three and five objectives, (2) comparisons on the WFG test suite with three and five objectives, (3) comparisons on the DTLZ test suite with 8, 10 and 15 objectives, and (4) comparisons on the WFG test suite with 8, 10 and 15 objectives. Table \ref{tab:1} lists the median HV values and interquartile ranges (IQRs) obtained by each of the six algorithms over 21 runs on each of the DTLZ1-4 and CDTLZ1-4 test instances with three and five objectives. Here, a higher median HV value signifies a better performance. The best results are marked in bold and all the IQRs are in brackets. In Table \ref{tab:1},
`$+$', `$-$' and `$\approx$' indicate that the result is significantly better,
significantly worse and statistically similar
to that obtained by CoDEA, respectively, similarly hereinafter.



The DTLZ1-4 and CDTLZ1-4 test instances with three and five objectives are not very complex and most MaOEAs have good convergence on them. However, since their PFs are very representative, this group of experiments mainly aims to verify the distributions obtained by the algorithms on these test instances. It is clear from Table \ref{tab:1} that CoDEA achieves the highest HV scores in 11 out of 16 groups of comparative tests. Moreover, CoDEA has significant advantages on the tri-objective DTLZ2, DTLZ3, CDTLZ1, CDTLZ2, and CDTLZ4. Meanwhile, CoDEA does not perform significantly worse than the algorithms winning the highest HV scores on the tri-objective DTLZ1, and 5-objective DTLZ1, CDTLZ2, and CDTLZ4. Overall, CoDEA has a significant advantage over the other five algorithms for the DTLZ1-4 and CDTLZ1-4 test instances with three and five objectives.
 

\begin{figure}[htbp]
	\centering
	\subfloat[]{
		\includegraphics[width = 0.29\linewidth]{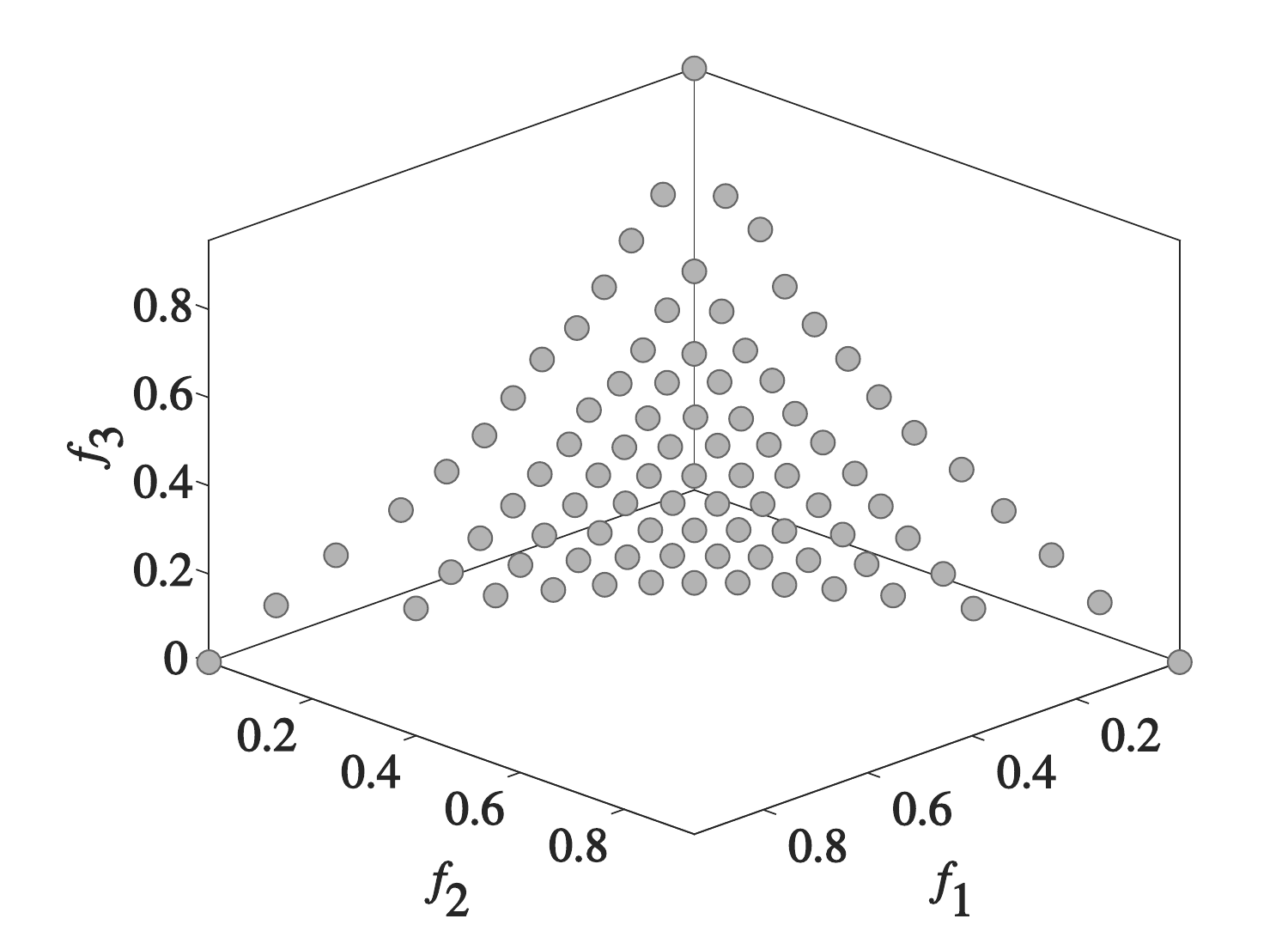}
		\label{fig:RVEA_CDTLZ2_3}
	}
	\subfloat[]{
		\includegraphics[width = 0.29\linewidth]{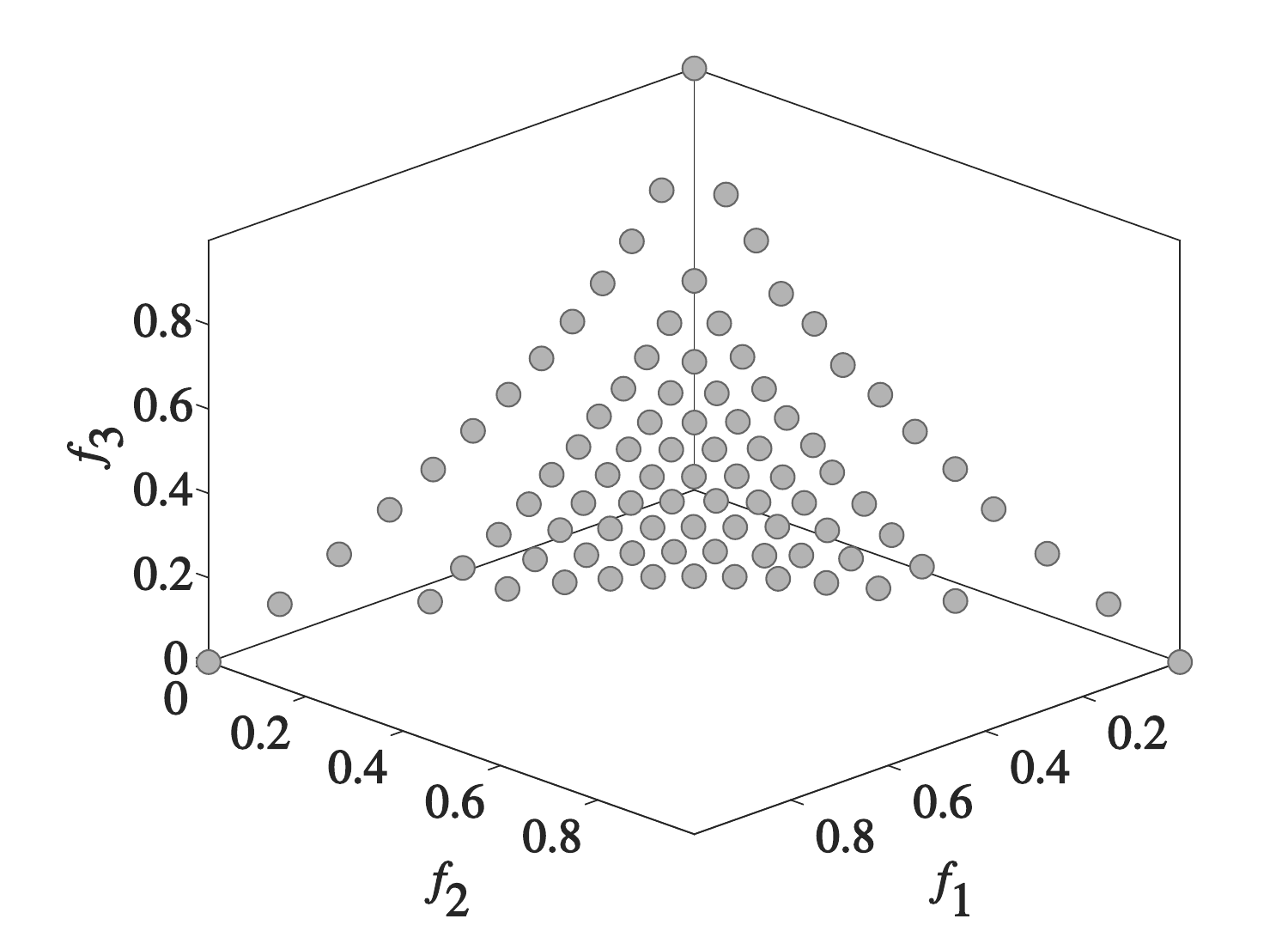}
		\label{fig:NSGA3_CDTLZ2_3}
	}
	\subfloat[]{
		\includegraphics[width = 0.29\linewidth]{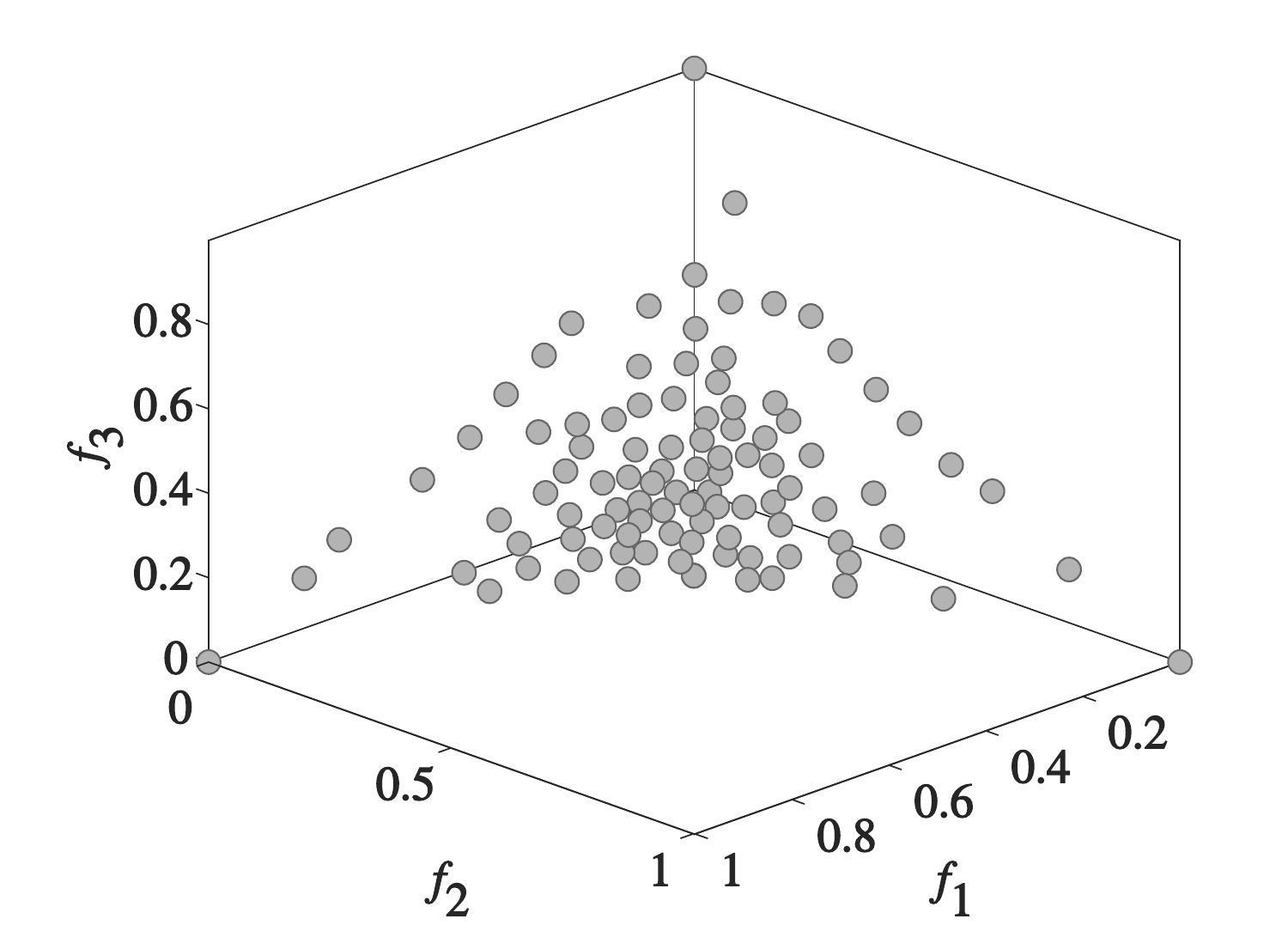}
		\label{fig:VaEA_CDTLZ2_3}
	}\\
	\subfloat[]{
		\includegraphics[width = 0.29\linewidth]{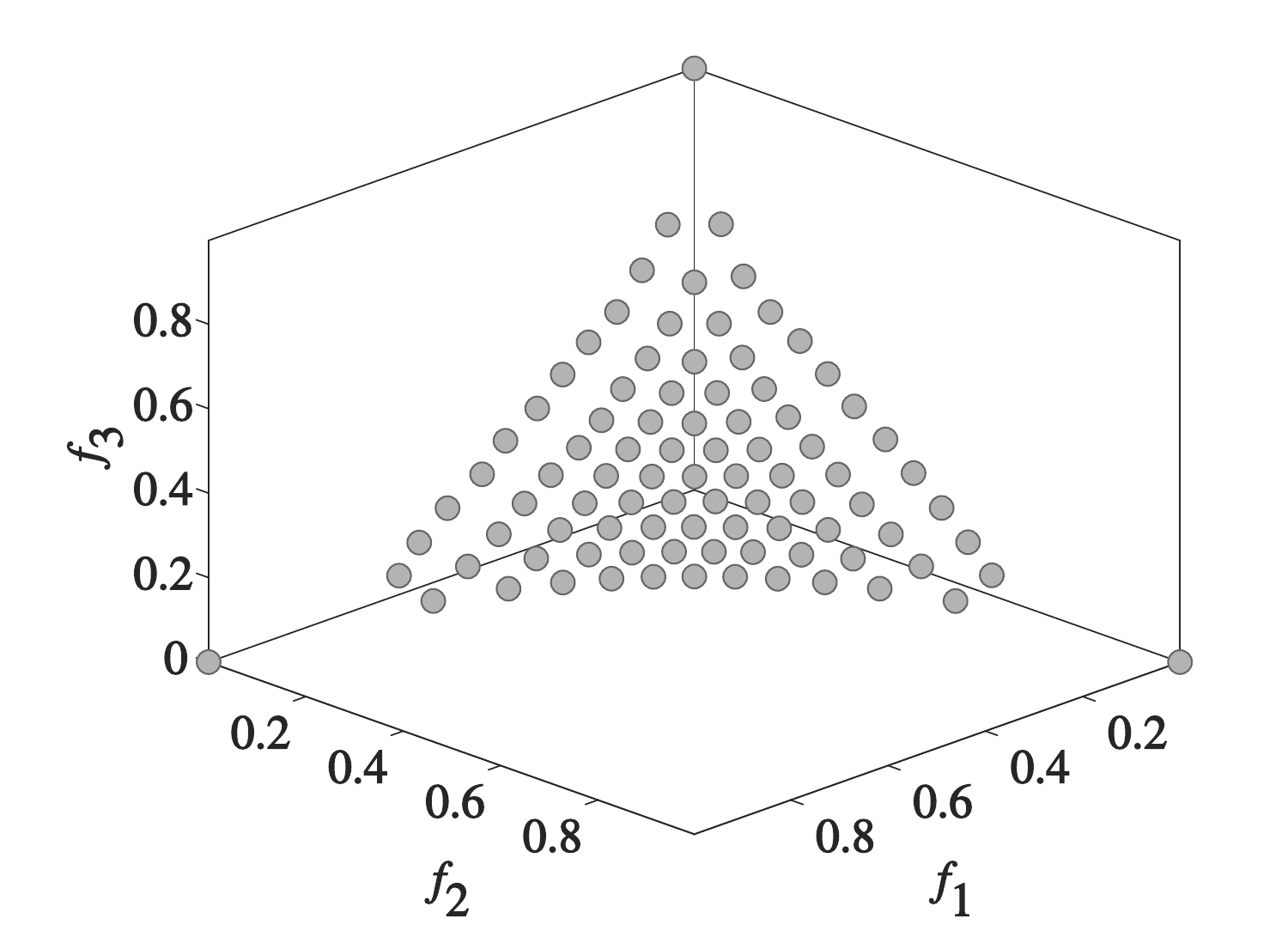}
		\label{fig:tDEA_CDTLZ2_3}
	}
	\subfloat[]{
		\includegraphics[width = 0.29\linewidth]{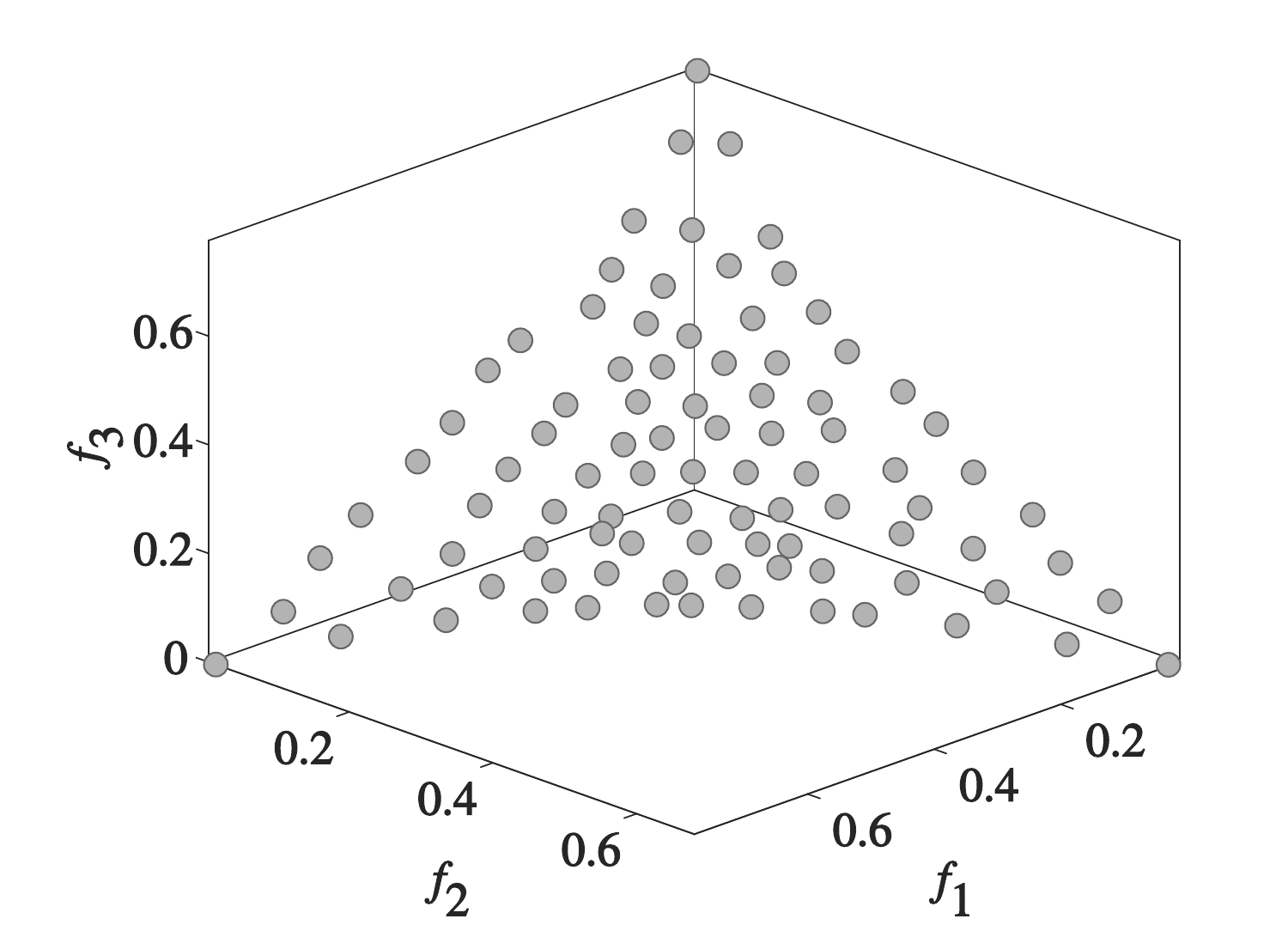}
		\label{fig:MP_DEA_CDTLZ2_3}
	}
	\subfloat[]{
		\includegraphics[width = 0.29\linewidth]{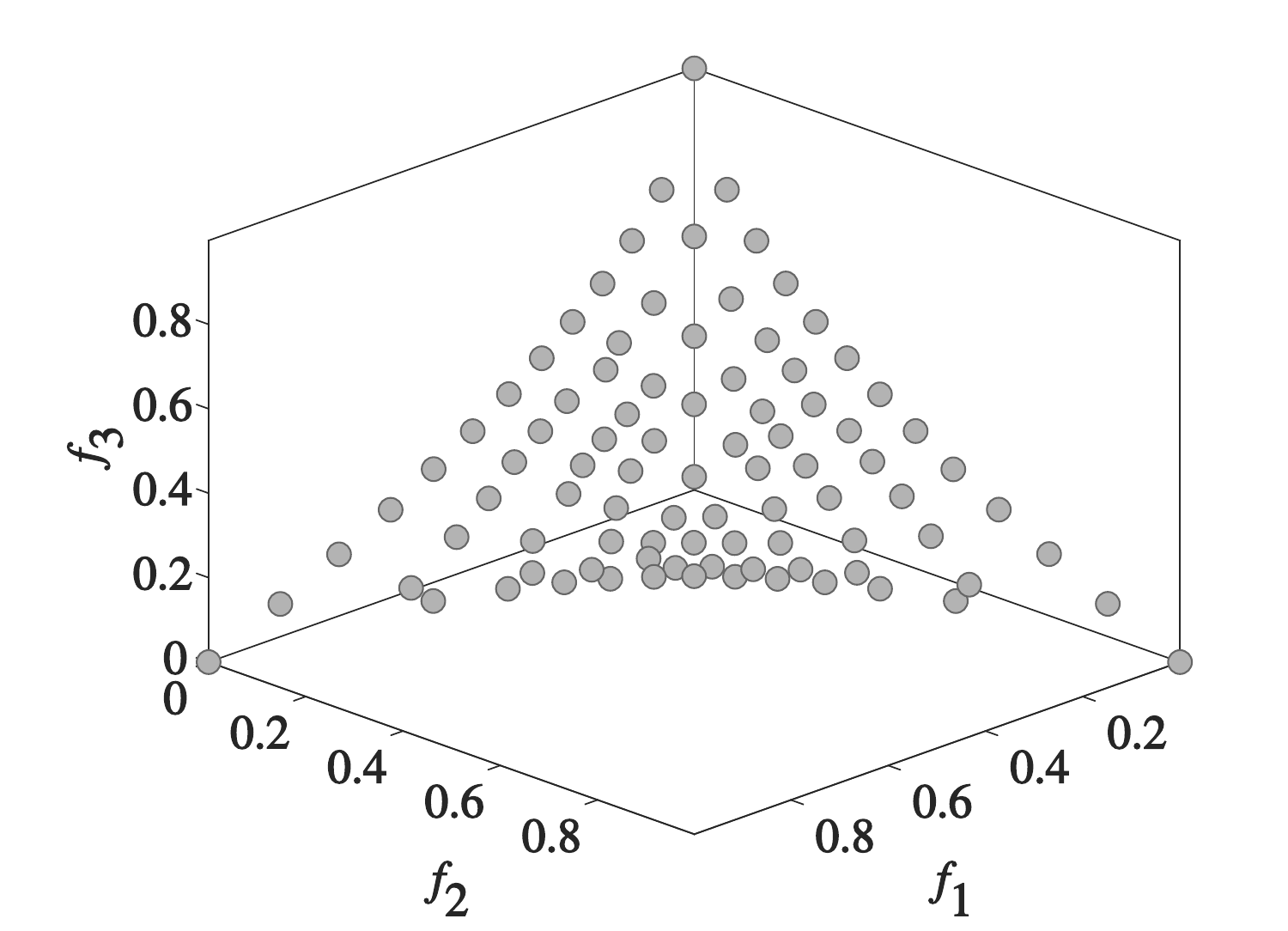}
		\label{fig:CoD_CDTLZ2_3}
	}
	\caption{The distributions of frontiers achieved by six algorithms on the tri-objective CDTLZ2 test instance. (a) RVEA, (b) NSGA-III, (c) VaEA, (d) $\theta$-DEA, (e) MP-DEA, and (f) CoDEA.}
	\label{fig:PF_3D_CDTLZ2}
\end{figure}

Fig. \ref{fig:PF_3D_DTLZ2} shows the distributions of frontiers achieved by the six algorithms on the tri-objective DTLZ2 test instance. For more intuitive obvious visual contrast, the population size is set to 300 for all algorithms in Fig. \ref{fig:PF_3D_DTLZ2}. Compared with four algorithms applying the cone decomposition idea, i.e., RVEA, NSGA-III, $\theta$-DEA, and MP-DEA, CoDEA alleviates the phenomenon that the population gathers at the vertex and disperses at the center. MP-DEA aggravates this phenomenon because the parameter of penalty term in MP-DEA is smaller. Although VaEA does not have the problem of aggregation or dispersion in a specific place, the uniformity of distribution of its final population is relatively poor due to the lack of guidance of reference vectors.

Fig. \ref{fig:PF_3D_CDTLZ2} shows the distributions of frontiers achieved by the six algorithms on the tri-objective CDTLZ2 test instance with a strongly convex PF. RVEA, NSGA-III, VaEA, and CoDEA obtain relatively complete approximate frontiers. Different from RVEA and NSGA-III, under the action of NBI method components, CoDEA greatly alleviates the trend of population aggregation to the center.  Similar to the phenomenon in Fig. \ref{fig:PF_3D_DTLZ2}, although VaEA obtains a relatively complete approximate PF, the population uniformity is not satisfactory. 
On the whole, it can be inferred from Figs. \ref{fig:PF_3D_DTLZ2} and \ref{fig:PF_3D_CDTLZ2} and Table \ref{tab:1} that the CoD method has a strong ability to preserve the uniformity of population in most cases.

\subsection{Comparisons on the WFG test suite with three and five objectives}

\begin{table*}[htbp]
  \centering
  \caption{Median HV values and IQRs (in brackets) obtained by six algorithms on the WFG1-9 test instances with three and five objectives.}
	\footnotesize
	\renewcommand\arraystretch{0.6}
    \begin{tabu} to 1 \linewidth{|X[2,c]|X[0.4,c]|X[3.5,c]|X[3.5,c]|X[3.5,c]|X[3.5,c]|X[3.5,c]|X[3.5,c]|}
    \toprule
    Pro. & m     & RVEA  & NSGA-III & VaEA  & $\theta -$DEA  & MP-DEA & CoDEA \\
    \midrule
    \multirow{2}[4]{*}{WFG1} & 3     & 9.3845e-1 (2.81e-3) - & 9.4390e-1 (7.49e-4) - & 9.3634e-1 (1.02e-3) - & 9.4240e-1 (1.28e-3) - & 9.3768e-1 (4.40e-3) - & \textbf{9.4479e-1 (9.29e-4)} \\
\cmidrule{2-8}          & 5     & 9.9841e-1 (1.15e-4) - & 9.9865e-1 (1.35e-4) $\approx$ & 9.9595e-1 (3.76e-3) - & 9.9647e-1 (6.29e-4) - & 9.8991e-1 (3.08e-3) - & \textbf{9.9865e-1 (9.30e-5)} \\
    \midrule
    \multirow{2}[4]{*}{WFG2} & 3     & 9.2773e-1 (1.09e-3) - & 9.3078e-1 (1.33e-3) - & 9.2472e-1 (1.67e-3) - & \textbf{9.3544e-1 (5.54e-4) +} & 9.2880e-1 (4.27e-3) - & 9.3199e-1 (9.16e-4) \\
\cmidrule{2-8}          & 5     & 9.9643e-1 (8.89e-4) - & 9.9737e-1 (4.66e-4) $\approx$ & 9.9148e-1 (1.89e-3) - & 9.9659e-1 (4.07e-4) - & 9.7309e-1 (2.66e-3) - & \textbf{9.9740e-1 (3.44e-4)} \\
    \midrule
    \multirow{2}[4]{*}{WFG3} & 3     & 3.4329e-1 (4.54e-3) - & 3.8746e-1 (5.03e-3) $\approx$ & 3.7179e-1 (7.72e-3) - & 3.8579e-1 (5.09e-3) - & \textbf{3.8938e-1 (3.37e-3) $\approx$} & 3.8906e-1 (3.60e-3) \\
\cmidrule{2-8}          & 5     & 1.8077e-1 (8.68e-3) $\approx$ & 1.6867e-1 (1.51e-2) - & 1.2438e-1 (2.81e-2) - & 2.1355e-1 (2.78e-3) + & \textbf{2.1506e-1 (1.10e-2) +} & 1.8005e-1 (9.01e-3) \\
    \midrule
    \multirow{2}[4]{*}{WFG4} & 3     & 5.5484e-1 (1.07e-3) - & 5.5879e-1 (9.87e-5) - & 5.4843e-1 (3.79e-3) - & 5.5901e-1 (2.56e-4) - & 5.5613e-1 (6.69e-4) - & \textbf{5.6086e-1 (9.03e-4)} \\
\cmidrule{2-8}          & 5     & 8.0997e-1 (8.46e-4) - & 8.1030e-1 (1.14e-3) $\approx$ & 7.7447e-1 (6.59e-3) - & \textbf{8.1075e-1 (3.10e-4) $\approx$} & 8.0556e-1 (7.92e-4) - & 8.1068e-1 (8.02e-4) \\
    \midrule
    \multirow{2}[4]{*}{WFG5} & 3     & 5.1735e-1 (5.62e-4) $\approx$ & 5.1842e-1 (3.35e-5) + & 5.1390e-1 (2.15e-3) - & \textbf{5.1843e-1 (3.78e-5) +} & 5.1558e-1 (6.83e-4) - & 5.1753e-1 (1.43e-3) \\
\cmidrule{2-8}          & 5     & 7.6154e-1 (7.19e-4) $\approx$ & 7.6164e-1 (5.65e-4) $\approx$ & 7.4260e-1 (4.69e-3) - & 7.6150e-1 (2.75e-4) $\approx$ & 7.5584e-1 (1.14e-3) - & \textbf{7.6169e-1 (4.92e-4)} \\
    \midrule
    \multirow{2}[4]{*}{WFG6} & 3     & 5.0123e-1 (1.69e-2) $\approx$ & \textbf{5.1541e-1 (1.93e-2) $\approx$} & 4.9704e-1 (1.73e-2) - & 5.0492e-1 (1.85e-2) $\approx$ & 5.0747e-1 (1.59e-2) $\approx$ & 5.0964e-1 (1.73e-2) \\
\cmidrule{2-8}          & 5     & \textbf{7.4903e-1 (1.45e-2) $\approx$} & 7.4051e-1 (1.76e-2) $\approx$ & 7.2758e-1 (1.73e-2) - & 7.4187e-1 (1.24e-2) $\approx$ & 7.4001e-1 (1.38e-2) - & 7.4309e-1 (2.04e-2) \\
    \midrule
    \multirow{2}[4]{*}{WFG7} & 3     & 5.5493e-1 (6.39e-4) - & 5.5835e-1 (3.28e-4) - & 5.4609e-1 (3.98e-3) - & 5.5865e-1 (1.97e-4) - & 5.5713e-1 (5.40e-4) - & \textbf{5.6076e-1 (1.00e-3)} \\
\cmidrule{2-8}          & 5     & 8.0895e-1 (5.02e-4) - & 8.1049e-1 (7.55e-4) - & 7.8901e-1 (3.93e-3) - & \textbf{8.1141e-1 (7.08e-4) $\approx$} & 8.0698e-1 (6.41e-4) - & 8.1140e-1 (4.01e-4) \\
    \midrule
    \multirow{2}[4]{*}{WFG8} & 3     & 4.7107e-1 (2.67e-3) $\approx$ & 4.7250e-1 (3.22e-3) $\approx$ & 4.6203e-1 (4.11e-3) - & 4.7325e-1 (2.53e-3) + & \textbf{4.7699e-1 (2.55e-3) +} & 4.7218e-1 (3.88e-3) \\
\cmidrule{2-8}          & 5     & \textbf{7.0387e-1 (2.30e-3) +} & 6.9940e-1 (5.23e-3) $\approx$ & 6.5223e-1 (9.62e-3) - & 7.0072e-1 (2.85e-3) $\approx$ & 6.9946e-1 (2.58e-3) $\approx$ & 6.9973e-1 (3.61e-3) \\
    \midrule
    \multirow{2}[4]{*}{WFG9} & 3     & 5.3805e-1 (3.27e-3) $\approx$ & 5.3666e-1 (4.37e-3) $\approx$ & 5.2892e-1 (4.05e-3) - & 5.3865e-1 (3.42e-3) $\approx$ & \textbf{5.4188e-1 (2.12e-3) +} & 5.3851e-1 (2.55e-3) \\
\cmidrule{2-8}          & 5     & 7.7637e-1 (3.58e-3) $\approx$ & 7.6907e-1 (5.69e-3) - & 7.4409e-1 (9.06e-3) - & 7.7735e-1 (2.01e-3) $\approx$ & \textbf{7.7780e-1 (2.89e-3) +} & 7.7566e-1 (3.31e-3) \\

    \midrule
    \multicolumn{2}{|c|}{+/-/$\approx$} & 1/9/8 & 1/7/10 & 0/18/0 & 4/6/8 & 4/11/3 &  \\

    \bottomrule

    \end{tabu}%
  \label{tab:2}%
\end{table*}%

The WFG test suite is one of the most popular benchmark test suites for many-objective optimization. The WFG1-9 test problems can evaluate their ability of algorithms in obtaining a well-converged and well-distributed set of solutions since they are designed to present several complexities.
Table \ref{tab:2} records the median HV values and IQRs obtained by each of the six algorithms over 21 runs on each of the WFG1-9 test instances with three and five objectives.

The WFG1 test instances are characterized by scaling, strong biases, and mixed frontier geometries. As shown in Table \ref{tab:2}, CoDEA achieves the best performances of HV  on the tri-objective and 5-objective WFG1 test instances and it performs statistically significantly better than the other five algorithms. The PF of the WFG2 test problem is composed of several disconnected convex segments. And $\theta$-DEA and CoDEA achieve the best HV values on the tri-objective and 5-objective WFG2 test instances, respectively. Unlike WFG1 and WFG2, WFG3 is a degenerate problem. MP-DEA is superior to the other algorithms on the tri-objective and 5-objective WFG3. In the 12 groups of comparative tests on WFG4-9, each of CoDEA, $\theta$-DEA and MP-DEA using similar algorithmic frameworks wins the best HV values three times while RVEA, NSGA-III, and VaEA do so twice, one time, and none, respectively.

Overall, CoDEA, $\theta$-DEA and MP-DEA with the similar frameworks perform significantly better than the other three algorithms, i.e., RVEA, NSGA-III, and VaEA, on these 18 WFG1-9 test instances with three and five objectives. On the whole, CoDEA, $\theta$-DEA and MP-DEA achieve the best HV performances in 15 out of 18 comparative tests. Among them, CoDEA behaves slightly better than MP-DEA and $\theta$-DEA. Specifically, CoDEA wins the best HV values in six comparative tests while MP-DEA and $\theta$-DEA do so in five ones and four ones, respectively.

\begin{figure}[htbp]
	\centering
	\subfloat[]{
		\includegraphics[width = 0.29\linewidth]{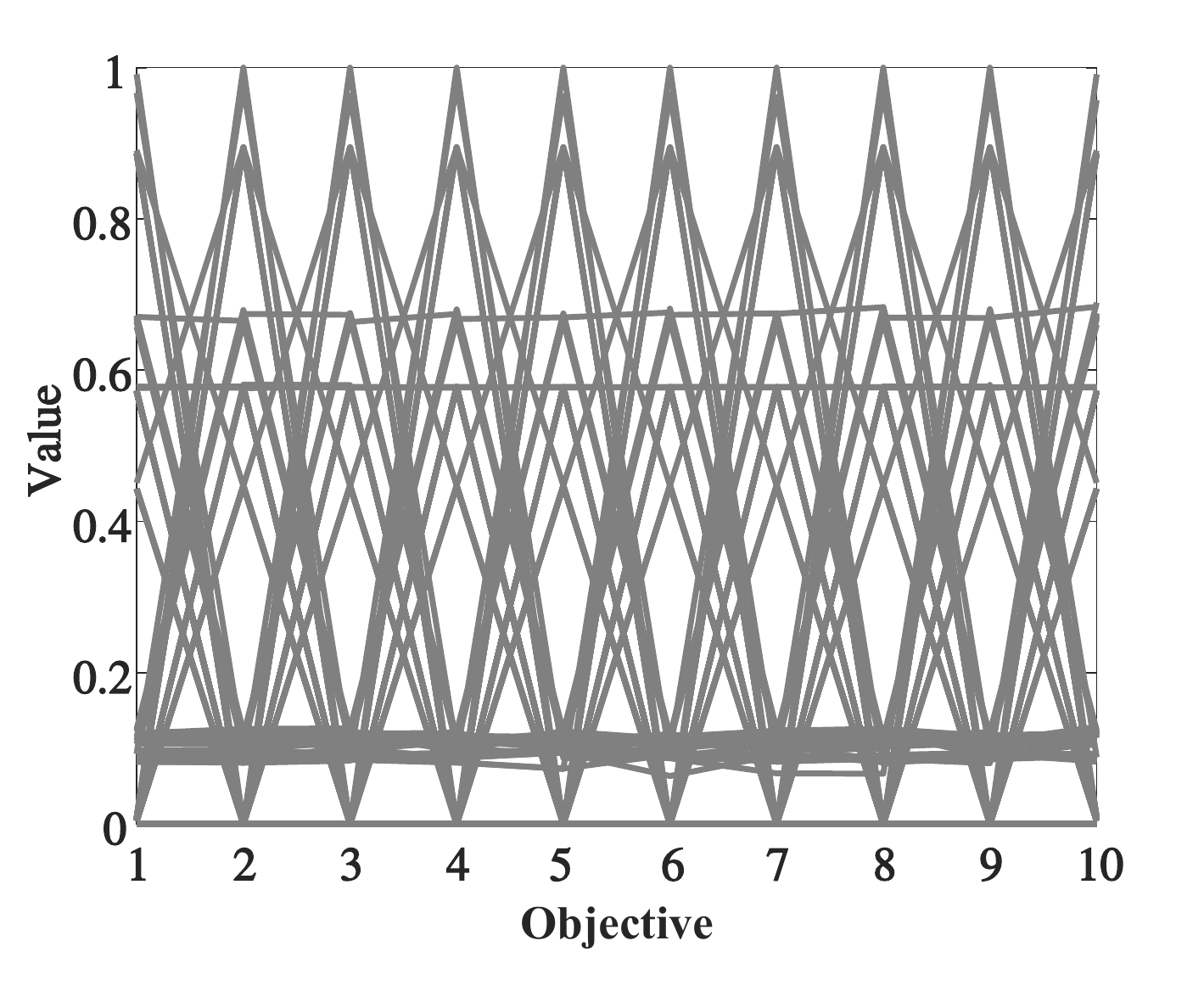}
		\label{fig:RVEA_CDTLZ2_10}
	}
	\subfloat[]{
		\includegraphics[width = 0.29\linewidth]{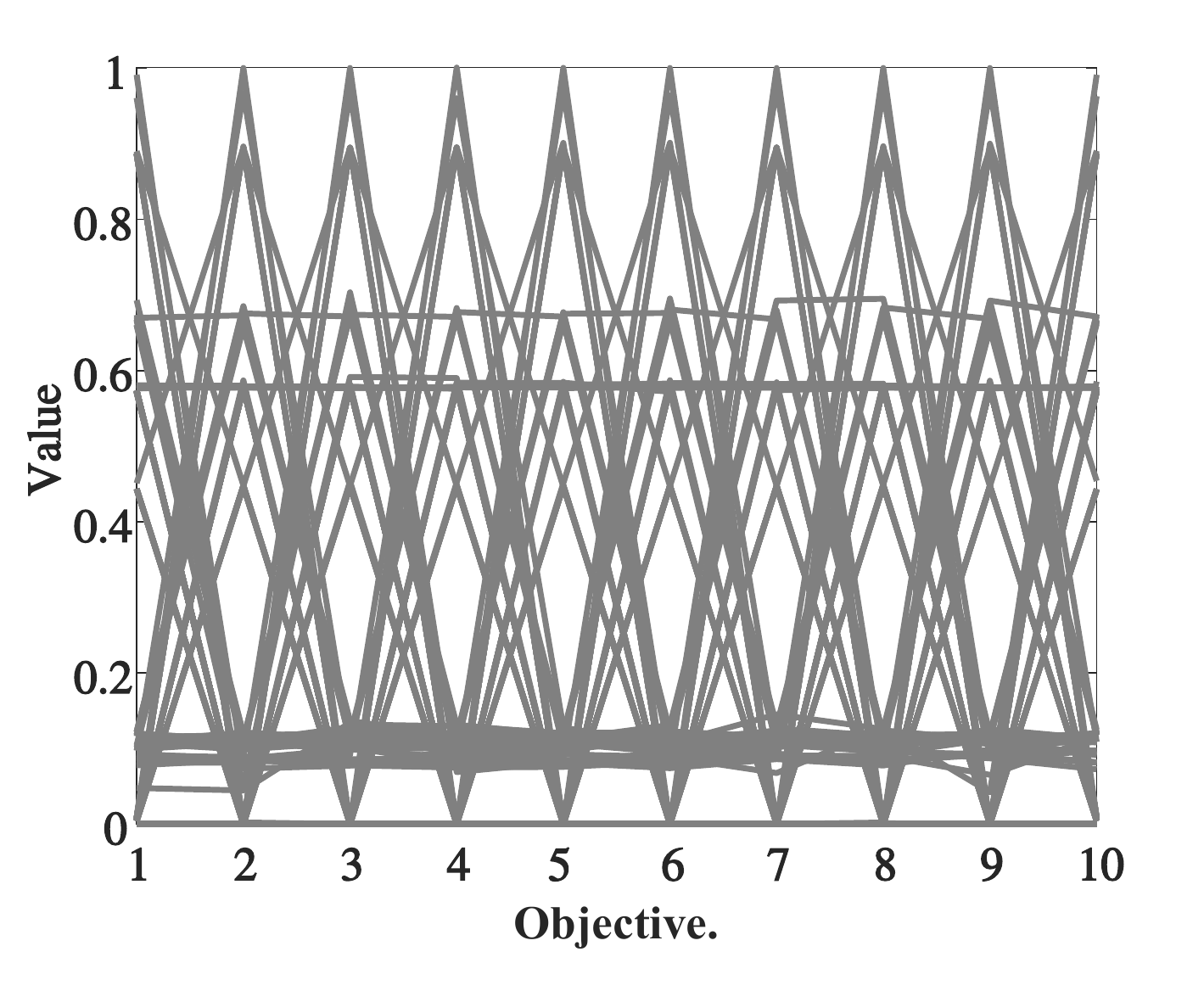}
		\label{fig:NSGA3_DTLZ2_10}
	}
	\subfloat[]{
		\includegraphics[width = 0.29\linewidth]{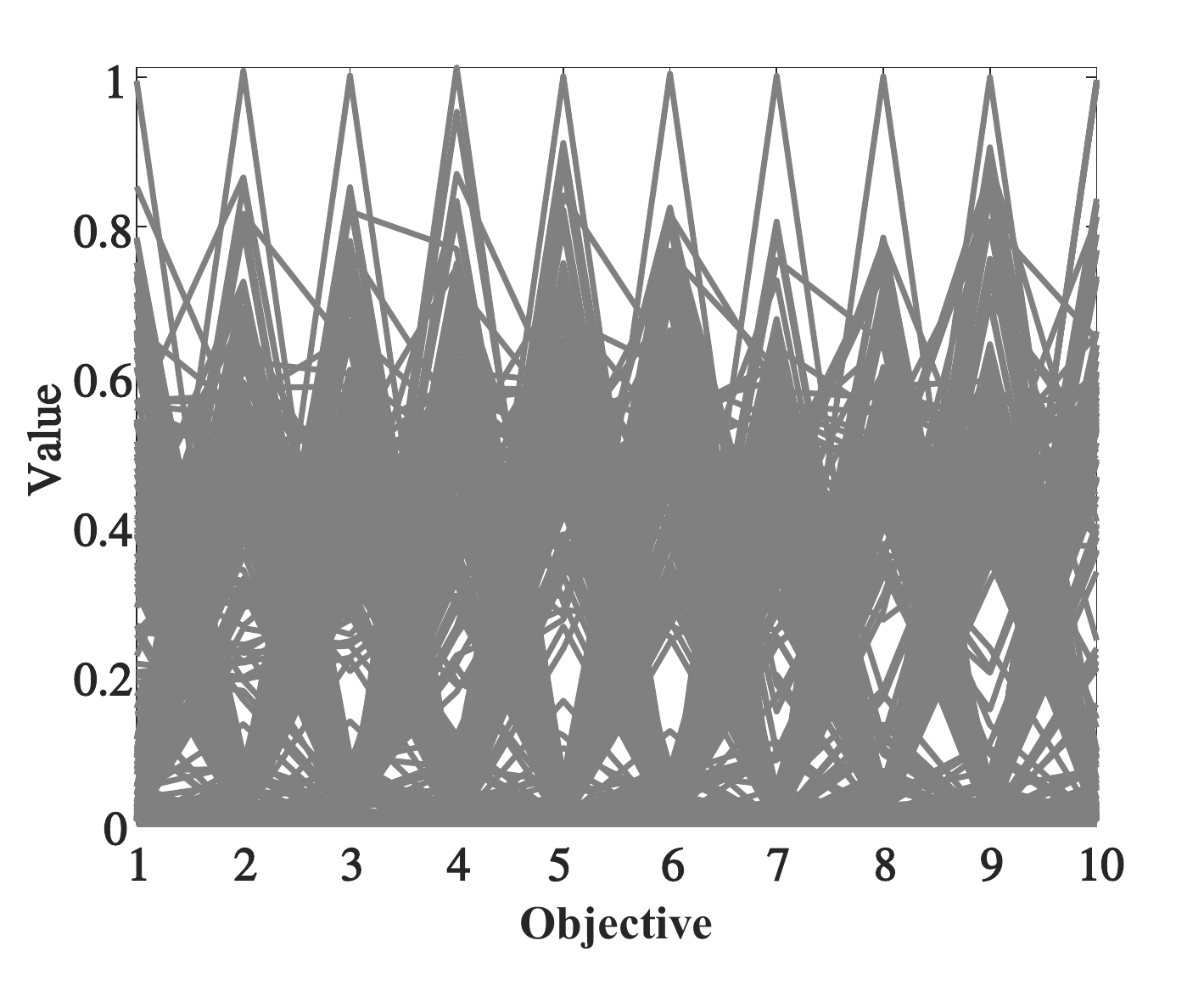}
		\label{fig:VaEA_DTLZ2_10}
	}\\
	\subfloat[]{
		\includegraphics[width = 0.29\linewidth]{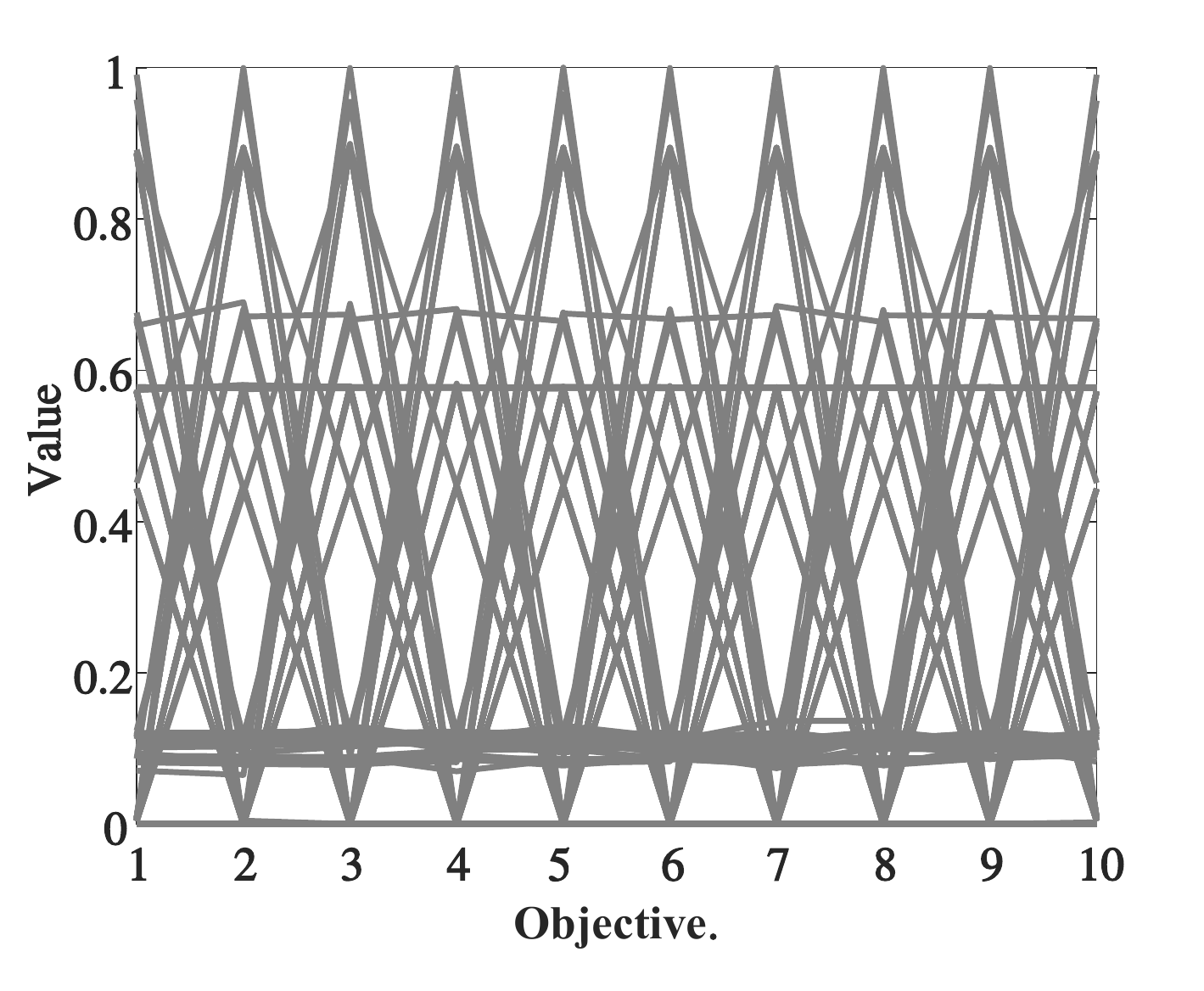}
		\label{fig:tDEA_DTLZ2_10}
	}
	\subfloat[]{
		\includegraphics[width = 0.29\linewidth]{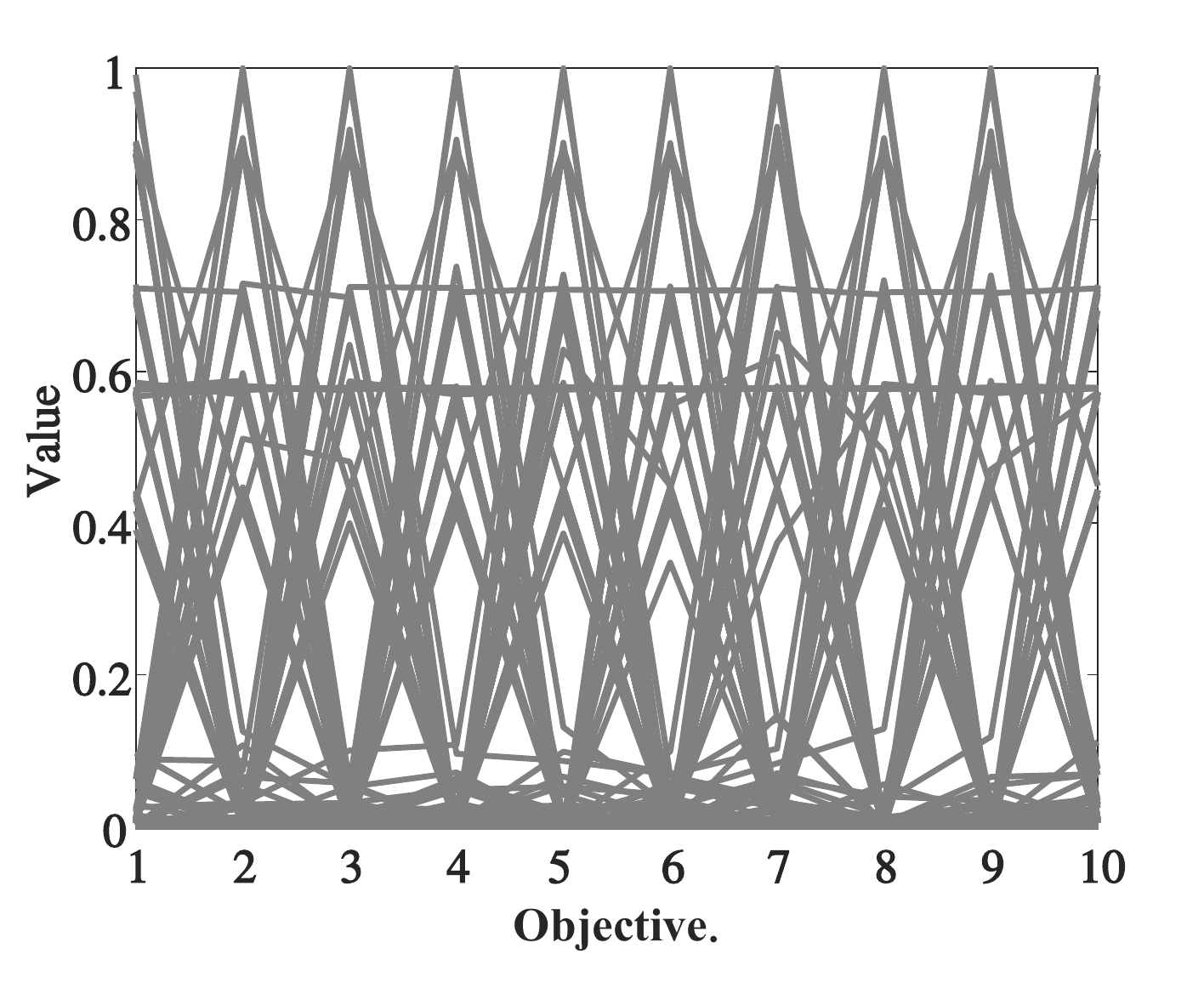}
		\label{fig:MP_DEA_DTLZ2_10}
	}
	\subfloat[]{
		\includegraphics[width = 0.29\linewidth]{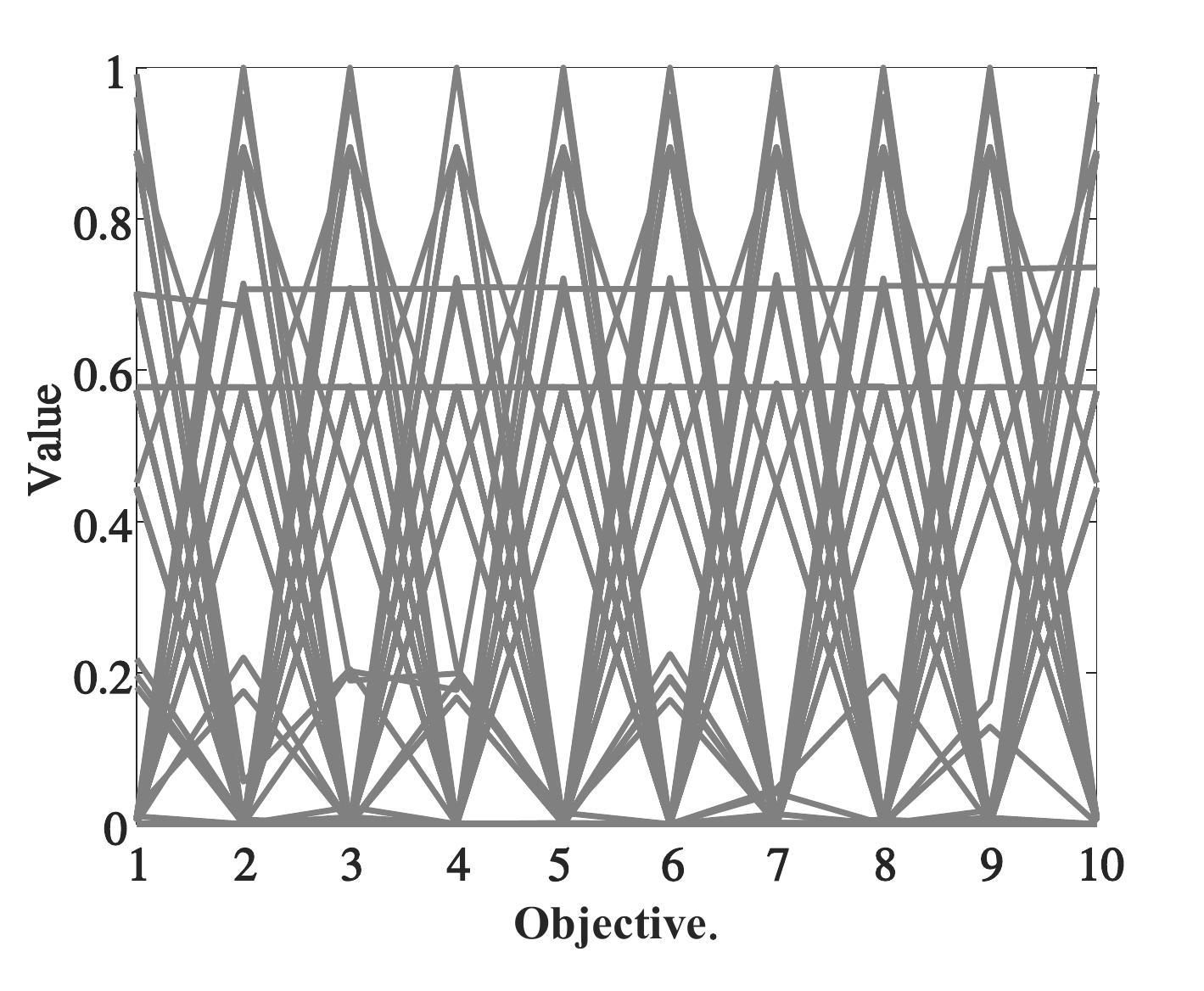}
		\label{fig:CoD_DTLZ2_10}
	}
	\caption{The parallel coordinates of frontiers achieved by six algorithms on 10-objective DTLZ2. (a) RVEA, (b) NSGA-III, (c) VaEA, (d) $\theta$-DEA, (e) MP-DEA, and (f) CoDEA.}
	\label{fig:PF_10D_DTLZ2}
\end{figure}

 \begin{figure}[htbp]
	\centering
	\subfloat[]{
		\includegraphics[width = 0.29\linewidth]{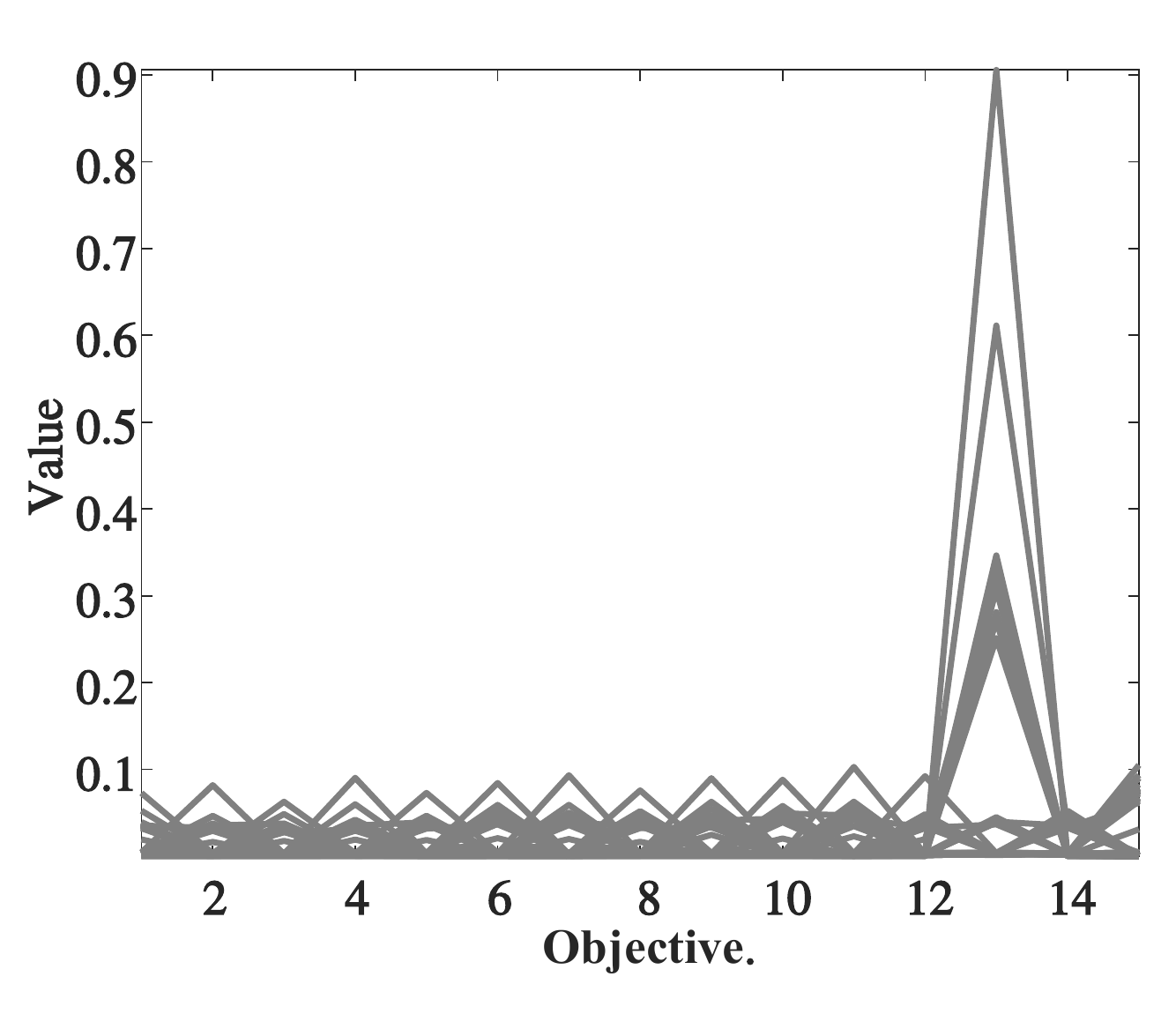}
		\label{fig:RVEA_CDTLZ2_15}
	}
	\subfloat[]{
		\includegraphics[width = 0.29\linewidth]{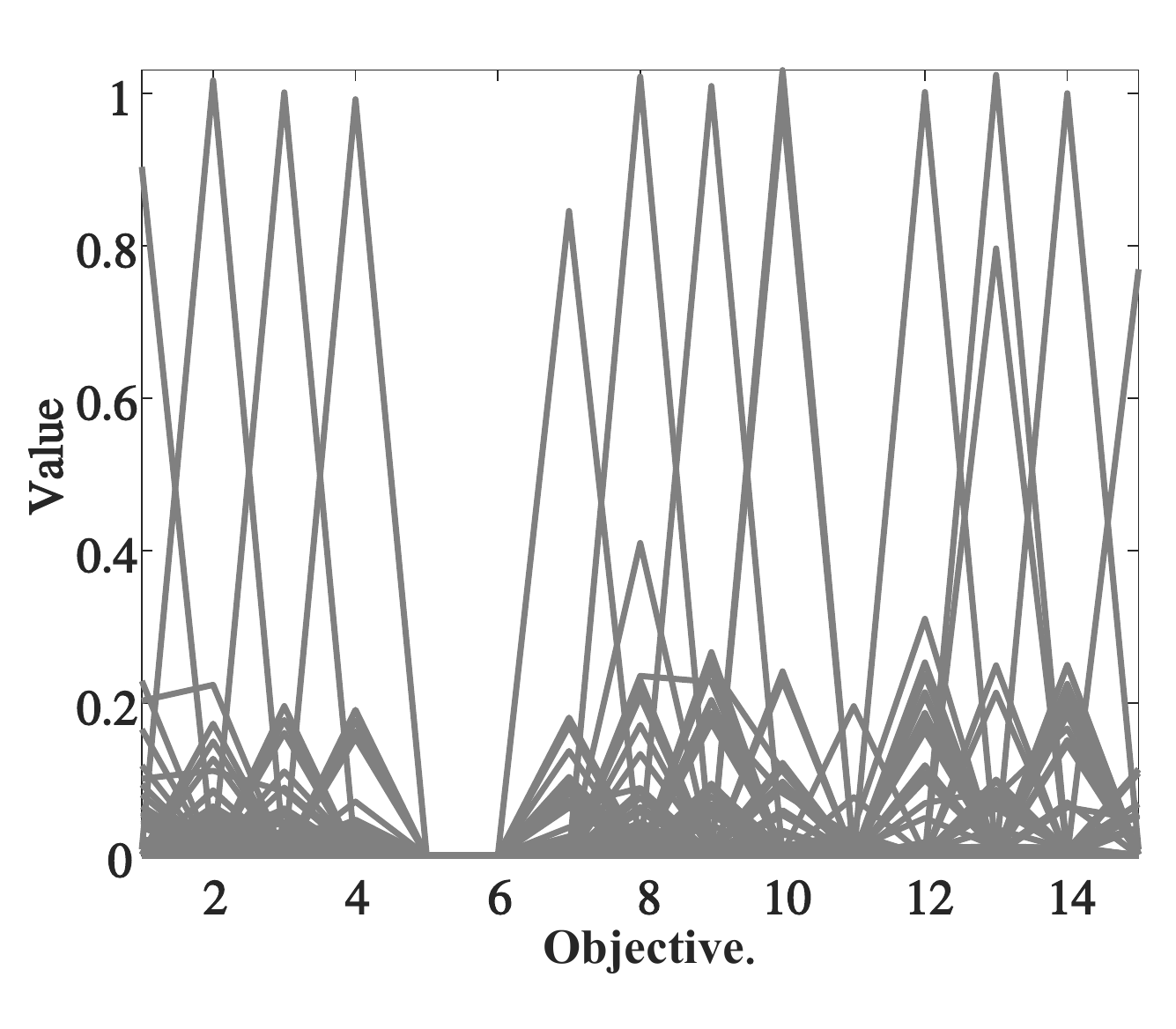}
		\label{fig:NSGA3_DTLZ2_15}
	}
	\subfloat[]{
		\includegraphics[width = 0.29\linewidth]{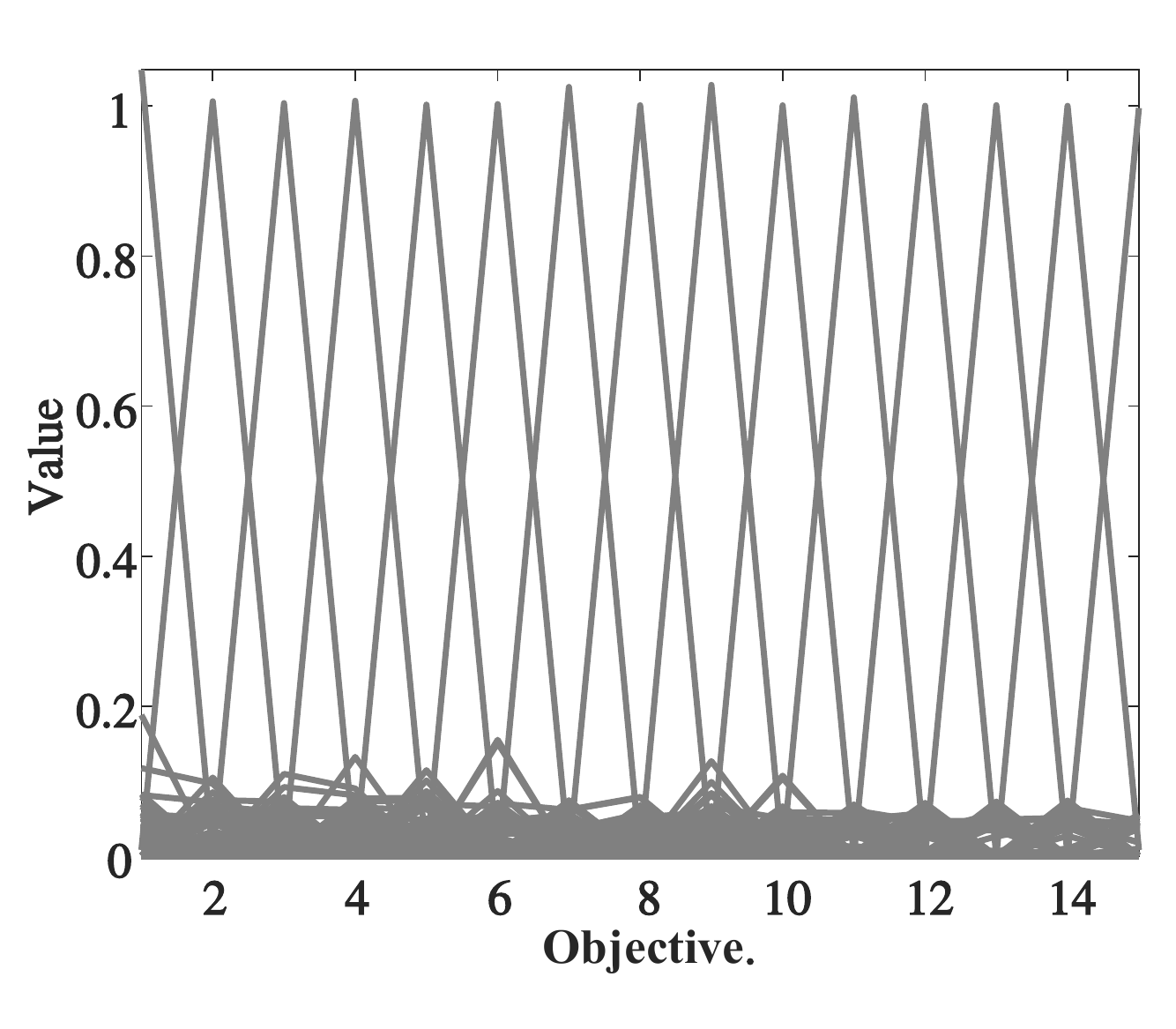}
		\label{fig:VaEA_CDTLZ2_15}
	}\\
	\subfloat[]{
		\includegraphics[width = 0.29\linewidth]{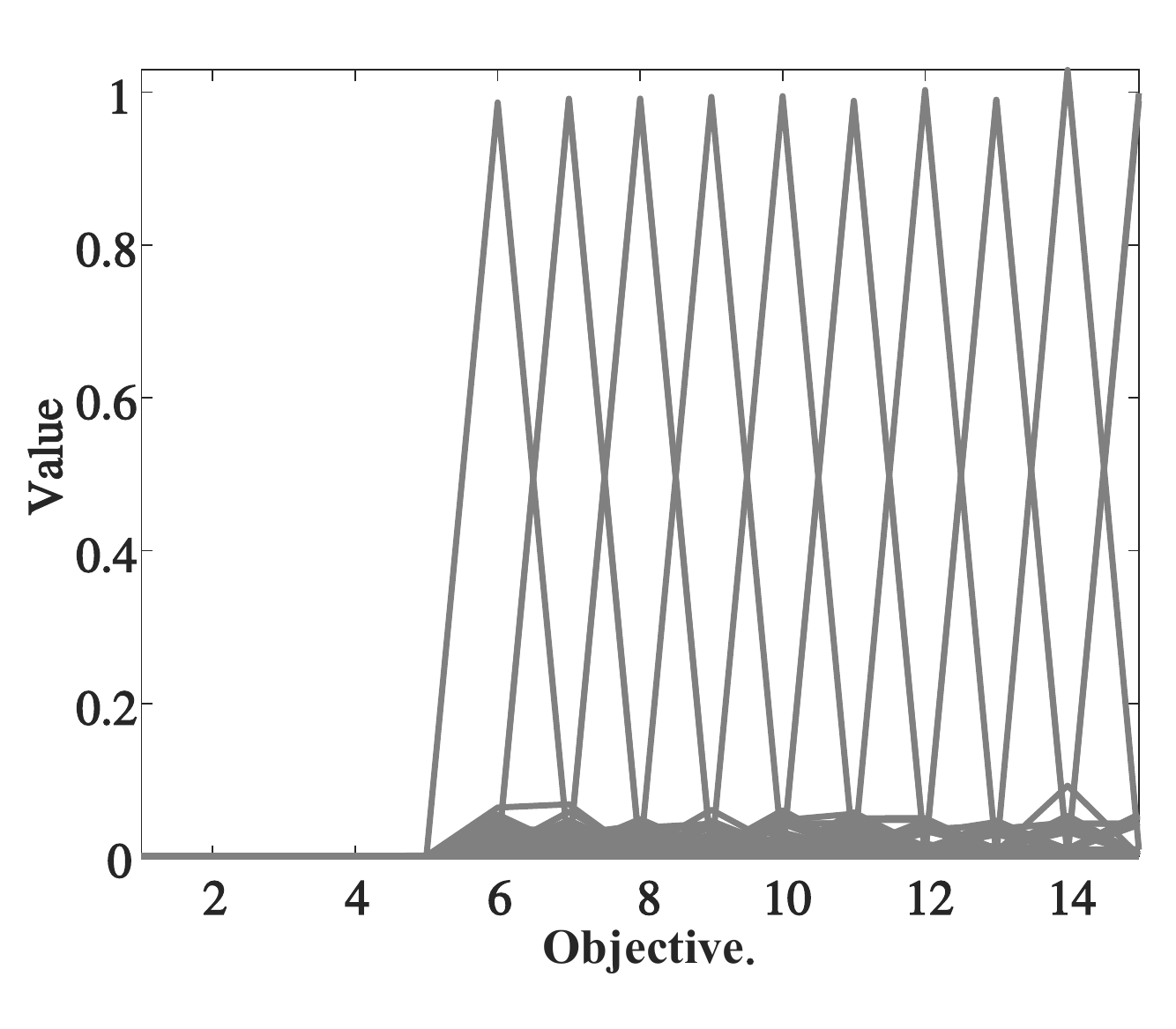}
		\label{fig:tDEA_CDTLZ2_15}
	}
	\subfloat[]{
		\includegraphics[width = 0.29\linewidth]{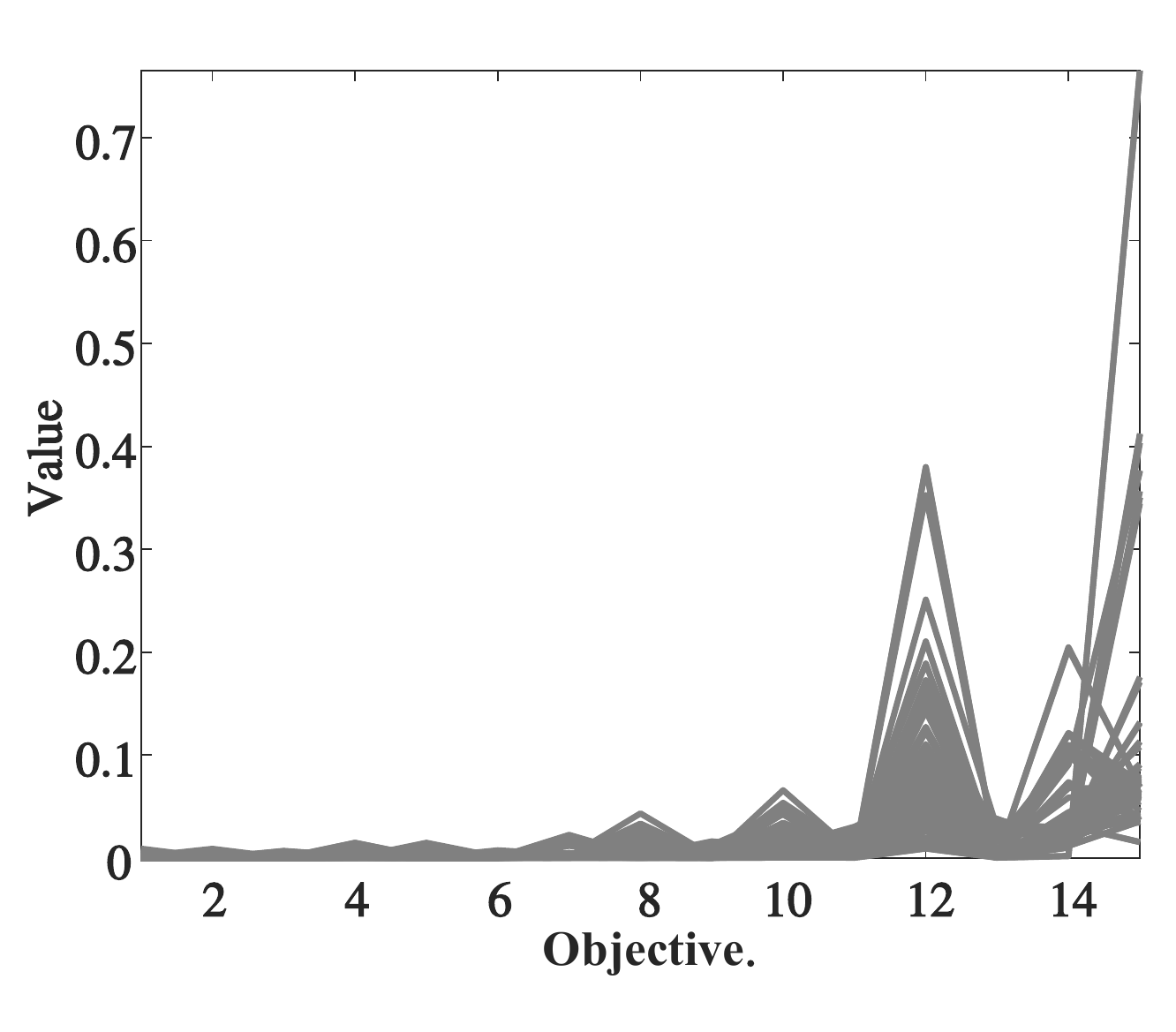}
		\label{fig:MP_DEA_CDTLZ2_15}
	}
	\subfloat[]{
		\includegraphics[width = 0.29\linewidth]{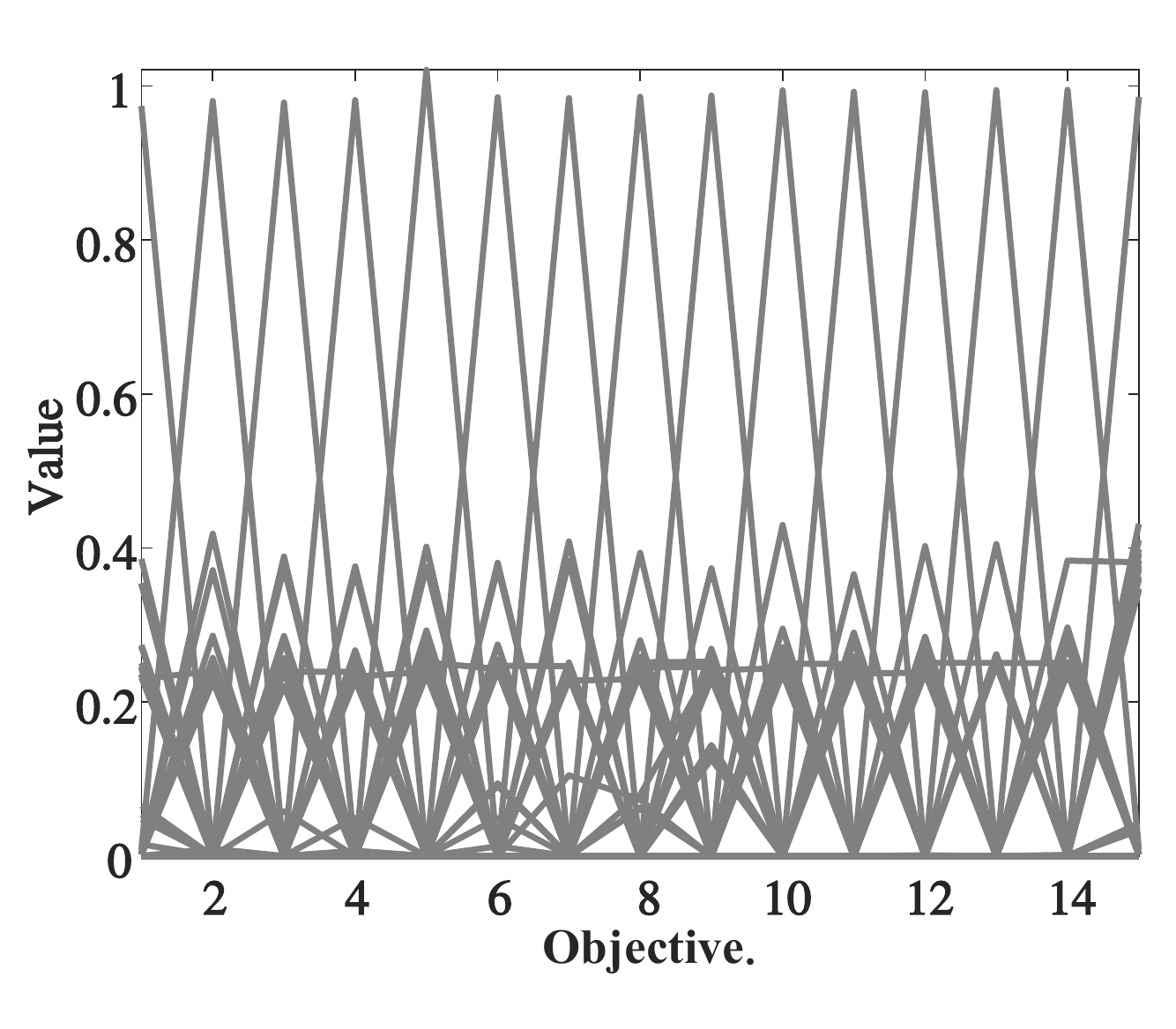}
		\label{fig:CoD_CDTLZ2_15}
	}
	\caption{The parallel coordinates of frontiers achieved by six algorithms on 15-objective CDTLZ2. (a) RVEA, (b) NSGA-III, (c) VaEA, (d) $\theta$-DEA, (e) MP-DEA, and (f) CoDEA.}
	\label{fig:PF_15D_CDTLZ2}
\end{figure}

\subsection{Comparisons on the DTLZ test suite with 8, 10 and 15 objectives}

Table \ref{tab:3} lists the median HV values and IQRs obtained by each of the six algorithms over 21 runs on each of the DTLZ1-4 and CDTLZ1-4 test instances with 8, 10 and 15 objectives.
It is evident from Table \ref{tab:3} that CoDEA achieves the best HV performances on 18 out of 24 test instances in this group of comparative experiments. 
The remaining six test instances include the 8-, 10- and 15-objective DTLZ1, 10-objective DTLZ4, 15-objective CDTLZ3, and 8-objective CDTLZ4.

On the three DTLZ1 test instances, RVEA wins the best HV performances and CoDEA is slightly inferior to RVEA. In addition,  VaEA and MP-DEA obtain fairly high IQRs on them. It signifies that they have poor robustness on the three DTLZ1 test instances. This phenomenon can be also observed on the DTLZ3, CDTLZ1, and CDTLZ3 test instances, which have the multimodal characteristics similar to that of DTLZ1. 
CoDEA is slightly inferior to MP-DEA on 10-objective DTLZ4 and to NSGA-III on 8-objective CDTLZ4, respectively. For 15-objective DTLZ3, RVEA has obvious advantages over the other five algorithms. The cause is probably that the effectiveness of the various operations of objective normalization adopted in NSGA-III, $\theta$-DEA, MP-DEA, and CoDEA depends on the accuracy of the nadir point. However, the multimodal characteristics of DTLZ3 and the growth of the number of objectives increase the difficulty in obtaining an accurate nadir point for 15-objective DTLZ3.  


%
\begin{table*}[htbp]
  \centering
  \caption{Median HV values and IQRs (in brackets) obtained by six algorithms on the DTLZ1-4 and CDTLZ1-4 test instances with 8, 10 and 15 objectives.}
	\footnotesize
	\renewcommand\arraystretch{0.6}
    \begin{tabu} to 1 \linewidth{|X[1.6,c]|X[0.4,c]|X[2.8,c]|X[2.8,c]|X[2.8,c]|X[2.8,c]|X[2.8,c]|X[2.8,c]|}
    \toprule
    Pro. & m     & RVEA  & NSGA-III & VaEA  & $\theta -$DEA  & MP-DEA & CoDEA \\
    \midrule
    \multirow{3}[6]{*}{DTLZ1} & 8     & \textbf{9.9762e-1 (9.18e-5) +} & 9.9749e-1 (3.59e-3) $\approx$ & 9.0364e-1 (8.21e-2) - & 9.9757e-1 (8.70e-5) + & 7.4891e-1 (3.78e-1) - & 9.9742e-1 (1.05e-4) \\
\cmidrule{2-8}          & 10    & \textbf{9.9969e-1 (2.55e-5) +} & 9.9968e-1 (4.00e-4) $\approx$ & 9.5354e-1 (2.73e-2) - & 9.9969e-1 (2.57e-5) + & 7.1184e-1 (2.91e-1) - & 9.9965e-1 (1.05e-5) \\
\cmidrule{2-8}          & 15    & \textbf{9.9992e-1 (1.05e-5) +} & 9.9992e-1 (1.88e-5) + & 9.6864e-1 (1.63e-2) - & 9.9991e-1 (1.90e-2) $\approx$ & 9.5612e-1 (3.46e-2) - & 9.9991e-1 (2.42e-5) \\
    \midrule
    \multirow{3}[6]{*}{DTLZ2} & 8     & 9.2382e-1 (3.40e-4) - & 9.2302e-1 (5.22e-4) - & 9.0275e-1 (2.58e-3) - & 9.2375e-1 (4.12e-4) - & 9.3160e-1 (6.34e-4) - & \textbf{9.3289e-1 (1.48e-3)} \\
\cmidrule{2-8}          & 10    & 9.6972e-1 (3.06e-4) - & 9.6927e-1 (4.49e-4) - & 9.4455e-1 (4.84e-3) - & 9.6974e-1 (2.27e-4) - & 9.7403e-1 (2.52e-4) - & \textbf{9.7468e-1 (2.76e-4)} \\
\cmidrule{2-8}          & 15    & 9.9059e-1 (1.28e-4) - & 9.7671e-1 (2.10e-2) - & 9.4914e-1 (9.99e-3) - & 9.9063e-1 (1.21e-4) - & 9.9081e-1 (9.65e-5) - & \textbf{9.9114e-1 (2.82e-4)} \\
    \midrule
    \multirow{3}[6]{*}{DTLZ3} & 8     & 9.2243e-1 (1.06e-3) $\approx$ & 9.1646e-1 (8.23e-3) $\approx$ & 0.0000e+0 (2.21e-1) - & 9.2219e-1 (2.58e-3) $\approx$ & 2.4550e-1 (1.86e-1) - & \textbf{9.2351e-1 (2.44e-2)} \\
\cmidrule{2-8}          & 10    & 9.6974e-1 (4.03e-4) - & 9.6931e-1 (2.03e-2) - & 2.3840e-1 (6.85e-1) - & 9.6962e-1 (6.74e-4) - & 8.1308e-1 (4.21e-1) - & \textbf{9.7312e-1 (8.42e-4)} \\
\cmidrule{2-8}          & 15    & 9.9051e-1 (1.58e-4) $\approx$ & 9.1824e-1 (9.48e-2) - & 0.0000e+0 (0.00e+0) - & 9.9053e-1 (6.56e-4) $\approx$ & 8.5746e-1 (8.86e-1) - & \textbf{9.9053e-1 (4.89e-4)} \\
    \midrule
    \multirow{3}[6]{*}{DTLZ4} & 8     & 9.2406e-1 (5.49e-4) - & 9.2366e-1 (4.64e-4) - & 9.0370e-1 (5.53e-3) - & 9.2401e-1 (4.47e-4) - & 9.3173e-1 (5.76e-4) - & \textbf{9.3227e-1 (1.42e-3)} \\
\cmidrule{2-8}          & 10    & 9.6982e-1 (2.03e-4) - & 9.6963e-1 (2.29e-4) - & 9.4319e-1 (3.39e-3) - & 9.6976e-1 (1.90e-4) - & \textbf{9.7398e-1 (2.83e-4) $\approx$} & 9.7382e-1 (8.19e-4) \\
\cmidrule{2-8}          & 15    & 9.9059e-1 (1.67e-4) - & 9.9041e-1 (1.35e-2) - & 9.6368e-1 (1.63e-3) - & 9.9068e-1 (9.47e-5) - & 9.9082e-1 (1.96e-4) - & \textbf{9.9121e-1 (1.90e-4)} \\
    \midrule
    \multirow{3}[6]{*}{CDTLZ1} & 8     & 9.9734e-1 (1.88e-3) - & 9.8754e-1 (9.26e-2) - & 9.8559e-1 (1.41e-2) - & 9.9190e-1 (6.09e-3) - & 7.2353e-1 (1.24e-1) - & \textbf{9.9999e-1 (1.15e-4)} \\
\cmidrule{2-8}          & 10    & 9.9900e-1 (6.91e-4) - & 9.8295e-1 (3.02e-2) - & 9.9924e-1 (1.59e-3) - & 9.9742e-1 (3.49e-3) - & 9.4140e-1 (9.83e-2) - & \textbf{1.0000e+0 (3.92e-4)} \\
\cmidrule{2-8}          & 15    & 9.9462e-1 (2.35e-3) - & 9.9585e-1 (6.35e-3) - & 9.9876e-1 (2.91e-3) - & 9.8354e-1 (2.55e-2) - & 9.9365e-1 (2.77e-2) - & \textbf{9.9983e-1 (1.15e-3)} \\
    \midrule
    \multirow{3}[6]{*}{CDTLZ2} & 8     & 9.9383e-1 (3.90e-3) - & 9.9999e-1 (2.47e-4) - & 9.9999e-1 (8.50e-6) - & 9.9249e-1 (2.65e-3) - & 9.8745e-1 (6.01e-3) - & \textbf{1.0000e+0 (1.00e-6)} \\
\cmidrule{2-8}          & 10    & 9.9633e-1 (2.33e-3) - & 1.0000e+0 (2.50e-4) - & 1.0000e+0 (2.00e-6) - & 9.9637e-1 (1.79e-3) - & 9.9292e-1 (3.77e-3) - & \textbf{1.0000e+0 (0.00e+0)} \\
\cmidrule{2-8}          & 15    & 9.9428e-1 (2.08e-3) - & 9.9999e-1 (3.28e-5) - & 1.0000e+0 (0.00e+0) - & 9.9630e-1 (4.40e-3) - & 9.6508e-1 (1.99e-2) - & \textbf{1.0000e+0 (0.00e+0)} \\
    \midrule
    \multirow{3}[6]{*}{CDTLZ3} & 8     & 9.9889e-1 (1.50e-3) - & 9.9996e-1 (7.72e-2) - & 0.0000e+0 (0.00e+0) - & 9.8575e-1 (3.44e-3) - & 3.8191e-1 (7.28e-1) - & \textbf{9.9999e-1 (5.00e-6)} \\
\cmidrule{2-8}          & 10    & 9.9946e-1 (3.82e-4) $\approx$ & 9.9425e-1 (8.49e-2) $\approx$ & 8.9788e-1 (9.69e-1) - & 7.8362e-1 (3.82e-1) - & 9.8385e-1 (5.29e-2) $\approx$ & \textbf{1.0000e+0 (3.42e-1)} \\
\cmidrule{2-8}          & 15    & \textbf{9.9847e-1 (5.94e-4) +} & 9.8495e-1 (2.44e-2) + & 0.0000e+0 (0.00e+0) - & 8.8000e-1 (8.30e-2) - & 7.6134e-1 (9.72e-1) - & 9.3880e-1 (8.50e-2) \\
    \midrule
    \multirow{3}[6]{*}{CDTLZ4} & 8     & 9.9834e-1 (1.21e-3) - & \textbf{9.9999e-1 (3.00e-6) $\approx$} & 9.9999e-1 (4.25e-6) $\approx$ & 9.8961e-1 (2.32e-3) - & 9.9910e-1 (1.87e-3) - & 9.9999e-1 (3.50e-6) \\
\cmidrule{2-8}          & 10    & 9.9923e-1 (5.31e-4) - & 1.0000e+0 (0.00e+0) $\approx$ & 1.0000e+0 (0.00e+0) $\approx$ & 9.9022e-1 (2.20e-3) - & 9.9955e-1 (3.04e-4) - & \textbf{1.0000e+0 (0.00e+0)} \\
\cmidrule{2-8}          & 15    & 9.9606e-1 (4.75e-3) - & 1.0000e+0 (0.00e+0) $\approx$ & 1.0000e+0 (0.00e+0) $\approx$ & 9.9197e-1 (4.36e-4) - & 9.9879e-1 (1.21e-3) - & \textbf{1.0000e+0 (0.00e+0)} \\

    \midrule
    \multicolumn{2}{|c|}{+/-/$\approx$} & 3/18/3 & 2/15/7 & 0/21/3 & 2/19/3 & 0/22/2 &  \\

    \bottomrule
    \end{tabu}%
  \label{tab:3}%
\end{table*}%

\begin{table*}[htbp]
  \centering
  \caption{Median HV values and IQRs (in brackets) obtained by six algorithms on the WFG1-9 test instances with 8, 10 and 15 objectives.}
	\footnotesize
	\renewcommand\arraystretch{0.6}
    \begin{tabu} to 1 \linewidth{|X[1.4,c]|X[0.4,c]|X[2.8,c]|X[2.8,c]|X[2.8,c]|X[2.8,c]|X[2.8,c]|X[2.8,c]|}
    \toprule
	 Pro. & m     & RVEA  & NSGA-III & VaEA  & $\theta-$DEA  & MP-DEA & CoDEA \\
	\midrule
    \multirow{3}[6]{*}{WFG1} & 8     & 9.9799e-1 (6.98e-4) - & 9.9985e-1 (6.98e-5) - & 9.9957e-1 (4.38e-4) - & 9.9635e-1 (1.63e-3) - & 9.8895e-1 (4.34e-3) - & \textbf{9.9993e-1 (7.91e-5)} \\
\cmidrule{2-8}          & 10    & 9.9829e-1 (3.30e-4) - & 9.9991e-1 (5.85e-5) - & 9.9992e-1 (5.94e-5) - & 9.9640e-1 (8.02e-4) - & 9.9135e-1 (2.11e-3) - & \textbf{9.9998e-1 (1.82e-5)} \\
\cmidrule{2-8}          & 15    & 9.9937e-1 (3.34e-4) - & \textbf{1.0000e+0 (4.26e-6) $\approx$} & 1.0000e+0 (3.25e-6) + & 9.9538e-1 (1.62e-3) - & 9.8303e-1 (9.52e-3) - & 1.0000e+0 (2.21e-5) \\
    \midrule
    \multirow{3}[6]{*}{WFG2} & 8     & 9.9090e-1 (5.70e-3) - & 9.9885e-1 (7.85e-4) - & 9.9655e-1 (1.34e-3) - & 9.9068e-1 (2.93e-3) - & 9.8265e-1 (4.95e-3) - & \textbf{9.9934e-1 (2.76e-4)} \\
\cmidrule{2-8}          & 10    & 9.9300e-1 (2.97e-3) - & 9.9934e-1 (3.92e-4) - & 9.9687e-1 (1.20e-3) - & 9.9282e-1 (3.06e-3) - & 9.9147e-1 (4.18e-3) - & \textbf{9.9983e-1 (1.97e-4)} \\
\cmidrule{2-8}          & 15    & 9.8551e-1 (5.44e-3) - & 9.9914e-1 (4.53e-4) - & 9.9785e-1 (1.16e-3) - & 8.0918e-1 (4.88e-2) - & 9.6920e-1 (1.18e-2) - & \textbf{9.9953e-1 (3.88e-4)} \\
    \midrule
    \multirow{3}[6]{*}{WFG3} & 8     & 0.0000e+0 (0.00e+0) - & 7.5246e-2 (8.24e-3) - & 8.2418e-2 (7.40e-3) - & 8.4733e-2 (9.99e-3) $\approx$ & \textbf{9.0665e-2 (1.98e-2) $\approx$} & 8.5175e-2 (6.35e-3) \\
\cmidrule{2-8}          & 10    & 0.0000e+0 (0.00e+0) - & 5.3023e-2 (2.69e-2) - & 6.3012e-2 (1.21e-2) $\approx$ & 7.0864e-2 (1.97e-2) + & \textbf{8.7490e-2 (2.09e-2) +} & 6.0617e-2 (1.57e-2) \\
\cmidrule{2-8}          & 15    & \textbf{0.0000e+0 (0.00e+0) $\approx$} & 0.0000e+0 (0.00e+0) $\approx$ & 0.0000e+0 (0.00e+0) $\approx$ & 0.0000e+0 (0.00e+0) $\approx$ & 0.0000e+0 (0.00e+0) $\approx$ & 0.0000e+0 (0.00e+0) \\
    \midrule
    \multirow{3}[6]{*}{WFG4} & 8     & 9.2098e-1 (1.15e-3) - & 9.2182e-1 (9.58e-4) - & 8.9580e-1 (6.30e-3) - & 9.2216e-1 (6.52e-4) - & 9.2838e-1 (1.02e-3) - & \textbf{9.3282e-1 (7.84e-4)} \\
\cmidrule{2-8}          & 10    & 9.6758e-1 (6.02e-4) - & 9.6733e-1 (6.76e-4) - & 9.3128e-1 (5.40e-3) - & 9.6807e-1 (7.22e-4) - & 9.7216e-1 (4.43e-4) - & \textbf{9.7402e-1 (5.75e-4)} \\
\cmidrule{2-8}          & 15    & 9.9037e-1 (2.24e-4) - & 9.9030e-1 (3.96e-4) - & 9.5286e-1 (4.66e-3) - & 9.9061e-1 (1.83e-4) - & 9.9065e-1 (1.59e-4) - & \textbf{9.9119e-1 (1.39e-4)} \\
    \midrule
    \multirow{3}[6]{*}{WFG5} & 8     & 8.6284e-1 (5.23e-4) - & 8.6352e-1 (5.29e-4) - & 8.4192e-1 (4.69e-3) - & 8.6362e-1 (3.18e-4) - & 8.7008e-1 (4.53e-4) - & \textbf{8.7437e-1 (6.09e-4)} \\
\cmidrule{2-8}          & 10    & 9.0445e-1 (2.37e-4) - & 9.0475e-1 (1.83e-4) - & 8.7538e-1 (3.41e-3) - & 9.0482e-1 (2.12e-4) - & 9.0852e-1 (2.21e-4) - & \textbf{9.1052e-1 (2.09e-4)} \\
\cmidrule{2-8}          & 15    & 9.1760e-1 (1.07e-4) - & 9.1750e-1 (1.37e-4) - & 8.7793e-1 (4.70e-3) - & 9.1760e-1 (1.49e-4) - & 9.1768e-1 (1.31e-4) - & \textbf{9.1829e-1 (1.43e-4)} \\
    \midrule
    \multirow{3}[6]{*}{WFG6} & 8     & 8.2679e-1 (4.17e-2) - & 8.4506e-1 (2.77e-2) $\approx$ & 8.4438e-1 (1.85e-2) $\approx$ & 8.4682e-1 (1.79e-2) $\approx$ & \textbf{8.5312e-1 (2.10e-2) $\approx$} & 8.4323e-1 (3.42e-2) \\
\cmidrule{2-8}          & 10    & 8.7094e-1 (2.63e-2) - & 8.8086e-1 (2.55e-2) $\approx$ & 8.7070e-1 (2.02e-2) - & 8.7919e-1 (2.10e-2) $\approx$ & 8.8464e-1 (2.60e-2) $\approx$ & \textbf{8.8799e-1 (1.31e-2)} \\
\cmidrule{2-8}          & 15    & 8.6501e-1 (2.86e-2) - & 8.9094e-1 (2.25e-2) $\approx$ & 8.7607e-1 (4.19e-2) - & 8.8926e-1 (2.83e-2) $\approx$ & 8.9192e-1 (2.99e-2) $\approx$ & \textbf{8.9712e-1 (2.17e-2)} \\
    \midrule
    \multirow{3}[6]{*}{WFG7} & 8     & 9.1280e-1 (2.86e-3) - & 9.2141e-1 (5.43e-4) - & 9.0726e-1 (4.12e-3) - & 9.2236e-1 (4.04e-4) - & 9.3032e-1 (6.70e-4) - & \textbf{9.3365e-1 (5.84e-4)} \\
\cmidrule{2-8}          & 10    & 9.6455e-1 (1.31e-3) - & 9.6833e-1 (4.72e-4) - & 9.4970e-1 (1.90e-3) - & 9.6870e-1 (2.60e-4) - & 9.7351e-1 (3.20e-4) - & \textbf{9.7487e-1 (3.40e-4)} \\
\cmidrule{2-8}          & 15    & 9.9015e-1 (5.71e-4) - & 9.8304e-1 (6.58e-3) - & 9.6314e-1 (2.85e-3) - & 9.9056e-1 (1.39e-4) - & 9.9073e-1 (1.18e-4) - & \textbf{9.9127e-1 (3.53e-3)} \\
    \midrule
    \multirow{3}[6]{*}{WFG8} & 8     & 7.8730e-1 (1.71e-2) - & 7.9661e-1 (2.98e-2) - & 7.2624e-1 (2.20e-2) - & 8.0538e-1 (1.22e-2) - & 8.0689e-1 (5.39e-3) - & \textbf{8.1889e-1 (4.94e-3)} \\
\cmidrule{2-8}          & 10    & 8.9019e-1 (9.37e-2) $\approx$ & 8.7359e-1 (2.39e-2) - & 7.9650e-1 (2.00e-2) - & 8.8897e-1 (4.05e-2) $\approx$ & 8.8310e-1 (1.35e-2) - & \textbf{8.9145e-1 (9.30e-3)} \\
\cmidrule{2-8}          & 15    & 7.2478e-1 (2.22e-1) - & 9.1857e-1 (1.91e-2) - & 8.3348e-1 (1.26e-2) - & 9.2206e-1 (1.17e-2) - & 9.3075e-1 (9.81e-3) $\approx$ & \textbf{9.3443e-1 (2.21e-2)} \\
    \midrule
    \multirow{3}[6]{*}{WFG9} & 8     & 8.6854e-1 (1.29e-2) - & 8.6466e-1 (1.94e-2) - & 8.2087e-1 (1.58e-2) - & 8.7089e-1 (6.24e-3) - & 8.8140e-1 (4.79e-3) $\approx$ & \textbf{8.8246e-1 (5.28e-3)} \\
\cmidrule{2-8}          & 10    & 9.1790e-1 (1.10e-2) - & 9.1451e-1 (8.47e-3) - & 8.6307e-1 (1.36e-2) - & 9.1884e-1 (4.86e-3) - & 9.2431e-1 (3.86e-3) $\approx$ & \textbf{9.2524e-1 (2.99e-3)} \\
\cmidrule{2-8}          & 15    & 8.8999e-1 (6.53e-2) - & 9.3357e-1 (2.22e-2) $\approx$ & 8.3137e-1 (1.59e-1) - & 9.2794e-1 (1.23e-2) - & \textbf{9.3436e-1 (5.10e-3) $\approx$} & 9.3432e-1 (5.52e-3) \\

    \midrule
    \multicolumn{2}{|c|}{+/-/$\approx$} & 0/25/2 & 0/21/6 & 1/23/3 & 1/20/6 & 1/17/9 &  \\

    \bottomrule
    \end{tabu}%
  \label{tab:4}%
\end{table*}%

In CoDEA, the subproblems in the boundary layer and inner layer  adopt two different aggregation functions, respectively. The purpose is to alleviate the risk of poor distribution that the solutions for the subproblems in the inner layer gather to the central regions of frontiers excessively. Fig. \ref{fig:PF_10D_DTLZ2} plots the parallel coordinates of frontiers obtained by the six algorithms on 10-objective DTLZ2. It can be observed from Fig. \ref{fig:PF_10D_DTLZ2} that the frontiers obtained by RVEA, NSGA-III, and $\theta$-DEA all have some certain degrees of crowding phenomenons at the value of 0.1, while MP-DEA alleviates this phenomenon due to the use of mirror points. Although there is no obvious crowding phenomenon in the frontier acquired by VaEA, the very irregular distribution indicates that the frontier obtained by VaEA is with  poor uniformity of distribution. CoDEA alleviates this phenomenon to a great extent by the special treatment for the inner subproblems.

Fig. \ref{fig:PF_15D_CDTLZ2} depicts the parallel coordinates of frontiers obtained by the six algorithms on 15-objective CDTLZ2. It is clear from Fig. \ref{fig:PF_15D_CDTLZ2} that CoDEA among the six algorithms achieves obvious advantages in terms of the diversity and uniformity of frontier distribution. CoDEA not only captures the steep regions of the frontier, but also achieves a good distribution of population on 15-objective CDTLZ2. Although VaEA approximately obtains the maximum value of each objective, the diversity of frontier captured by VaEA is slightly worse than that by CoDEA due to the maximum angle selection and minimum fitness replacement strategy of VaEA. 
The objective values of frontiers achieved by the other four algorithms are below 0.1 for some objectives. It signifies that these algorithms do not capture some boundary regions of the frontier on 15-objective CDTLZ2.


\subsection{Comparisons on the WFG test suite with 8, 10 and 15 objectives}

Table \ref{tab:4} records the median HV values and IQRs obtained by each of the six algorithms over 21 runs on each of the WFG1-9 test instances with 8, 10 and 15 objectives.
It is evident that from Table \ref{tab:4} that CoDEA achieves the best HV scores on 21 out of 27 test instances in this group of comparative experiments.
It is worth noting that all the six algorithms do not perform well on the three WFG3 test instances with degradation frontiers. Specifically, MP-DEA and RVEA achieve, respectively, the best and worst HV performances on them. Further, it can be observed from Tables \ref{tab:2} and \ref{tab:4} that CoDEA achieves the best HV performances only on three out of the 12 tri-objective and 5-objective WFG4-9 test instances while it does so on 16 out of the 18 WFG4-9 test instances.
It is mainly due to the effectiveness of different strategies for dealing with the boundary and inner subproblems.



In a word, the results in Tables \ref{tab:1} and \ref{tab:2} validate the efficiency and stability of the CoD method for MOPs with three and five objectives. And the results in Tables \ref{tab:3} and \ref{tab:4} fully illustrate the effectiveness of adopting different optimization strategies for the boundary and inner subproblems for MaOPs with more than five objectives. It can be concluded from the Wilcoxon rank-sum test results in these four tables that CoDEA has significant advantages over the other five comparison algorithms on the DTLZ and WFG test instances.

\subsection{Comparison of ranking strategies for the inner subproblems}
\begin{table*}[htbp]
  \centering
  \caption{Median HV values and IQRs (in brackets) obtained by CoDEA and its four variants on the DTLZ2 and CDTLZ2 test instances with 8, 10 and 15 objectives.}
	\footnotesize
	\renewcommand\arraystretch{0.6}
    \begin{tabu} to1 \linewidth{|X[2.0,c]|X[0.4,c]|X[3.0,c]|X[3.0,c]|X[3.0,c]|X[3.0,c]|X[3.0,c]|}
    \toprule
    Pro. & m     & CoDEA+MPR  & CoDEA+NBI  & CoDEA+PBI  & CoDEA$^\star$  &CoDEA \\
    \midrule
    \multirow{6}[6]{*}{DTLZ2} & \multirow{2}[2]{*}{8} & 9.3008e-01 & 9.2143e-01 & 9.2340e-01 & 9.2360e-01 & \textbf{9.3290e-01} \\
          &       & (1.93e-3) - & (5.51e-4) - & (2.40e-4) - & (3.27e-4) - & \textbf{(1.32e-3)} \\
\cmidrule{2-7}          & \multirow{2}[2]{*}{10} & 9.7402e-01 & 9.6869e-01 & 9.6962e-01 & 9.6979e-01 & \textbf{9.7482e-01} \\
          &       &  (6.83e-4) - & (3.86e-4) - & (3.32e-4) - &  (2.33e-4) - & \textbf{(5.14e-4)} \\
\cmidrule{2-7}          & \multirow{2}[2]{*}{15} & \textbf{9.9123e-01} & 9.9074e-01 & 9.9067e-01 & 9.9066e-01 & 9.9109e-01 \\
          &       & \textbf{(1.40e-4) +} & (1.83e-4) - & (1.11e-4) - & (7.95e-5) - & (2.02e-4) \\
    \midrule
    \multirow{6}[6]{*}{CDTLZ2} & \multirow{2}[2]{*}{8} & 9.9999e-01 & 9.9999e-01 & 9.9997e-01 & 9.9994e-01 & \textbf{1.0000e+00} \\
          &       &  (7.75e-6) - & (2.15e-5) - &  (8.95e-5) - & (2.83e-4) - & \textbf{(3.00e-6)} \\
\cmidrule{2-7}          & \multirow{2}[2]{*}{10} & 1.0000e+00 & 1.0000e+00 & 9.9997e-01 & 9.9970e-01 & \textbf{1.0000e+00} \\
          &       & (3.25e-6) - & (4.52e-5) - & (1.89e-4) - & (3.33e-4) - & \textbf{(7.24e-13)} \\
\cmidrule{2-7}          & \multirow{2}[2]{*}{15} & 1.0000e+00 & 1.0000e+00 & 1.0000e+00 & 1.0000e+00 & \textbf{1.0000e+00} \\
          &       & (1.28e-5) - & (2.26e-6) - & (1.46e-5) - &  (6.34e-6) - & \textbf{(0.00e+0)} \\
	
    \midrule
    \multicolumn{2}{|c|}{+/-/$\approx$} & 1/5/0 & 0/6/0 & 0/6/0 & 0/6/0 &  \\

    \bottomrule
    \end{tabu}%
  \label{tab:5}%
\end{table*}%


In CoDEA, an angle-based ranking strategy is adopted to optimize the inner subproblems. The effectiveness of different optimization strategies for the inner subproblems are further compared in our experiments. 
CoDEA and its four variants, referred to as CoDEA+PBI, CoDEA+NBI, CoDEA+MPR, and CoDEA$^\star$, are considered in our experiments.
The four variants of CoDEA replace the
angle-based ranking strategy in the original CoDEA for the inner subproblems with 
the PBI-based, NBI-based, MPR-based ranking, and CoD-based ranking strategies, respectively.
The representative DTLZ2 and CDTLZ2 problems are selected and six groups of comparative tests on 8-, 10-, and 15-objective DTLZ2 and CDTLZ2 test instances are conducted in our experiments. 
The median HV values and IQRs obtained by each of CoDEA and its four variants over 21 runs on each of the DTLZ2 and CDTLZ2 test instances with 8, 10 and 15 objectives are listed in Table \ref{tab:5}.

It is clear from Table \ref{tab:5} that CoDEA achieves the best HV performances on all six test instances except for 15-objective DTLZ2. Although CoDEA+MPR achieves the best performance on 15-objective DTLZ2, the overall performance of CoDEA+MPR is slightly worse than that of CoDEA.  
The results of Wilcoxon rank-sum tests at the bottom of Table \ref{tab:5}
indicate that the original CoDEA has significant advantages over its  five variants owing to its effective angle-based ranking strategy for the inner subproblems.

\section{ Conclusion}
\label{sec:CONCLUSION}

In order to effectively balance the convergence, uniformity, and diversity of distribution, a collaborative
decomposition-based evolutionary algorithm, CoDEA, is proposed for many-objective optimization in this paper.
The core idea of CoDEA is the collaborative decomposition method, CoD.
This method inherits the NBI-style Tchebycheff function as a convergence measure to heighten the convergence and uniformity of distribution of the PBI method. Moreover, this method adopts an adaptive rotation strategy to enhance the distribution diversity of the NBI method. This strategy flexibly adjusts the degree of rotation of the NBI reference line to the PBI reference line for each subproblem. 
The CoD method skillfully recombines the components of the NBI method and PBI method, so as to inherit their respective advantages and make up for their respective defects. 
In order to alleviate the phenomenon of population aggregation, CoDEA adopts a maximum angle strategy for the inner subproblems different from that for the boundary subproblems.
The experimental results on the DTLZ and WFG test instances show that, compared with five popular MaOEAs, CoDEA achieves the best HV performances in 56 out of 85 groups of empirical tests.  CoDEA exhibits the best overall performances, benefitting from the collaborative decomposition method maintaining a
good balance among the convergence, uniformity, and diversity of distribution. 

In the future, we plan to further apply the collaborative decomposition method to heighten the performances of the other MaOEAs. It is also interesting to extend CoDEA to constrained MaOPs and MaOPs with complex frontiers such as discontinuity and degradation. In addition, it is quite promising to apply CoDEA to some real world MaOPs such as multi-objective neural architecture search of deep neural networks for medical image segmentation.


\section*{Declaration of interests}
The authors declare that they have no known competing financial interests or personal relationships that could have appeared to influence the work reported in this paper.

\section*{Acknowledgments}
    This work was supported in part by the Natural Science Foundation of Guangdong Province, China, under Grant 2020A1515011491 and Grant 2019A1515011792,
	in part by the Science Research Project of Guangzhou University under Grant YG2020008,
	in part by the Project of Innovation and Developing Universities of Education Department of Guangdong Province under Grant 2019KTSCX130,
	in part by the Guangzhou Science and Technology Project under Grant 202102080161,
	in part by the Guangdong Science and Technology Department, Grant 2019B010154004, 
	and in part by the Fundamental Research Funds for the Central Universities, SCUT, under Grant 2017MS043.


\bibliographystyle{elsarticle-harv}
\bibliography{CoDEA}

\end{document}